%% file: main.tex
\documentclass[twoside,11pt]{article}

\usepackage{blindtext}
\usepackage{style/jmlr2e}

\input{math_commands}

\usepackage{bm}
\usepackage{booktabs}
\usepackage{caption}
\usepackage{bbm}
\usepackage{subcaption}
\usepackage{sidecap}
\usepackage{wrapfig}

\allowdisplaybreaks

\newcommand{\vspacecaption}{\vspace{0pt}}
\newcommand{\vspacecaptionlow}{\vspace{0pt}}
\newcommand{\vspacesubcaption}{\vspace{0pt}}
\newcommand{\vspaceequation}{\vspace{0pt}}

\usepackage{lastpage}
\input{revision_macros}

\jmlrheading{24}{2023}{1-\pageref{LastPage}}{11/22; Revised
12/23}{12/23}{22-1254}{Jonas Rothfuss, Martin Josifoski, Vincent Fortuin, Andreas Krause}

\ShortHeadings{Scalable PAC-Bayesian Meta-Learning via the PAC-Optimal Hyper-Posterior}{Rothfuss, Josifoski, Fortuin, Krause}
\firstpageno{1}

\begin{document}

\title{\added{Scalable PAC-Bayesian Meta-Learning via the PAC-Optimal Hyper-Posterior: From Theory to Practice}}

\author{\name Jonas Rothfuss \email jonas.rothfuss@inf.ethz.ch \\
\addr ETH Zurich, Switzerland \\
\AND
\name Martin Josifoski \email martin.josifoski@epfl.ch \\
\addr EPFL Lausanne, Switzerland \\
\AND
\name Vincent Fortuin \email vincent.fortuin@helmholtz-munich.de \\
\addr Helmholtz AI and TU Munich, Germany \\
\AND
\name Andreas Krause \email krausea@ethz.ch \\
\addr ETH Zurich, Switzerland \\
}

\editor{Daniel Roy}

\maketitle

\begin{abstract}
\noindent
\edit{\emph{Meta-Learning} aims to accelerate the learning on new tasks by acquiring useful inductive biases from related data sources.
In practice, the number of tasks available for meta-learning is often small. Yet, most of the existing approaches rely on an abundance of meta-training tasks, making them prone to overfitting. \emph{How to regularize the meta-learner to ensure generalization to unseen tasks}, is a central question in the literature. 
We provide a theoretical analysis using the PAC-Bayesian framework and derive the first bound for meta-learners with \emph{unbounded loss functions}. Crucially, our bounds allow us to derive the \emph{PAC-optimal hyper-posterior} (PACOH)---\emph{the closed-form-solution of the PAC-Bayesian meta-learning problem}, thereby avoiding the reliance on nested optimization, giving rise to an optimization problem amenable to standard variational methods that \emph{scale well}}
{ Meta-Learning aims to speed up the learning process on new tasks by acquiring useful inductive biases from datasets of related learning tasks. While, in practice, the number of related tasks available is often small, most of the existing approaches assume an abundance of tasks; making them unrealistic and prone to overfitting. A central question in the meta-learning literature is how to regularize to ensure generalization to unseen tasks. In this work, we provide a theoretical analysis using the PAC-Bayesian theory and present a generalization bound for meta-learning, which was first derived by \citet{rothfuss2020pacoh}.
Crucially, the bound allows us to derive the closed form of the optimal hyper-posterior, referred to as PACOH, which leads to the best performance guarantees. We provide a theoretical analysis and empirical case study under which conditions and to what extent these guarantees for meta-learning improve upon PAC-Bayesian per-task learning bounds.
The closed-form PACOH inspires a practical meta-learning approach that avoids the reliance on bi-level optimization, giving rise to a stochastic optimization problem that is amenable to standard variational methods that scale well.}Our experiments show that, when instantiating the PACOH with Gaussian processes and Bayesian Neural Networks models, the resulting methods are more scalable, and yield state-of-the-art performance, both in terms of predictive accuracy and the quality of uncertainty estimates.
%

\end{abstract}

\begin{keywords}
  Meta-Learning, Transfer-Learning, PAC-Bayes, Learning Theory, Bayesian Neural Networks
\end{keywords}


\input{content/introduction}

\input{content/related_work}

\input{content/background}

\input{content/method1}

\input{content/method2}

\input{content/lr_case_study}

\input{content/method_algorithm}

\input{content/experiments}

\input{content/discussion}

\input{content/acknowledgements}

\bibliography{library}

\setlength\parindent{0pt}
\input{content/supplement_theory}

\input{content/supplement_blr}

\input{content/supplement_method}

\input{content/supplement_exps}

\end{document}

%% file: math_commands.tex
\usepackage[utf8]{inputenc} 
\usepackage{amsmath}
\usepackage{amssymb}
\usepackage{nicefrac}       
\usepackage{microtype}      
\usepackage{color}
\usepackage{graphicx}
\usepackage{tabularx}
\usepackage{caption}
\usepackage{subcaption}
\usepackage[english]{babel}
\usepackage{placeins}
\usepackage{color}
\usepackage{algorithm}
\usepackage{algorithmic}
\usepackage{lipsum}
\usepackage{xspace}
\usepackage{mathtools}

\everypar{\looseness=-1}

\usepackage{tikz}
\usetikzlibrary{bayesnet}

\newcommand{\rom}[1]
    {\text{\MakeUppercase{\romannumeral #1}}}

\newcommand{\bX}{\mathbf{X}}
\newcommand{\bx}{\mathbf{x}}

\newcommand{\bz}{\mathbf{z}}

\newcommand{\bm}{\mathbf{m}}
\newcommand{\bw}{\mathbf{w}}

\newcommand{\by}{\mathbf{y}}

\newcommand{\bI}{\mathbf{I}}

\newcommand{\given}{\, \vert \,}

\newcommand{\calA}{\mathcal{A}}

\newcommand{\calD}{\mathcal{D}}

\newcommand{\calF}{\mathcal{F}}

\newcommand{\calH}{\mathcal{H}}

\newcommand{\calL}{\mathcal{L}}
\newcommand{\calM}{\mathcal{M}}
\newcommand{\calN}{\mathcal{N}}
\newcommand{\calO}{\mathcal{O}}
\newcommand{\calP}{\mathcal{P}}
\newcommand{\calQ}{\mathcal{Q}}

\newcommand{\calS}{\mathcal{S}}
\newcommand{\calT}{\mathcal{T}}
\newcommand{\calU}{\mathcal{U}}

\newcommand{\calX}{\mathcal{X}}
\newcommand{\calY}{\mathcal{Y}}
\newcommand{\calZ}{\mathcal{Z}}

\newcommand{\E}{\mathbb{E}}
\newcommand{\R}{\mathbb{R}}

\newcommand{\expect}[2]{\mathbb{E}_{#1} \; #2}

\renewcommand{\overline}{\tilde}

\DeclareMathOperator*{\argmin}{arg\,min}
\DeclareMathOperator*{\argmax}{arg\,max}

\graphicspath{{./figures/}}

\newcommand{\beginsupplement}{%
        \setcounter{table}{0}
        \renewcommand{\thetable}{S\arabic{table}}%
        \setcounter{figure}{0}
        \renewcommand{\thefigure}{S\arabic{figure}}%
     }

%% file: revision_macros.tex
\usepackage{ifthen}
\usepackage[normalem]{ulem}
\usepackage{xcolor}
\usepackage{etoolbox}

\definecolor{lightgrey}{RGB}{200,200,200}

\newboolean{showRevision}
\setboolean{showRevision}{false} 

\newcommand{\added}[1]{
    \ifthenelse{\boolean{showRevision}}{\textcolor{blue}{#1}}{#1}
}

\newcommand{\edit}[2]{
    \ifthenelse{\boolean{showRevision}}{\textcolor{blue}{#2}}{#2}
}

\newcommand{\deleted}[1]{
    \ifthenelse{\boolean{showRevision}}{
        \textcolor{blue}{\sout{#1}}
    }{
    }
}

\newboolean{showRevisionII}
\setboolean{showRevisionII}{false} 

\newcommand{\addedII}[1]{
    \ifthenelse{\boolean{showRevisionII}}{\textcolor{blue}{#1}}{#1}
}

\newcommand{\deletedII}[1]{
    \ifthenelse{\boolean{showRevisionII}}{\textcolor{blue}{\sout{#1}}}{}
}

%% file: content/introduction.tex
\vspacecaption
\section{Introduction}
\vspacecaptionlow

\added{Learning new concepts and skills from a small number of examples as well as adapting them quickly in the face of changing circumstances is a key aspect of human intelligence. 
While modern machine learning systems are remarkably successful in learning complex patterns from vast quantities of data, they lack such adaptive capabilities under limited data. As a result, one often has to train our machine-learning models from scratch even though we have previously solved similar learning problems/tasks.}

\added{{\em Meta-Learning} \citep{thrun1998, Schmidhuber1987b} has emerged as a promising
avenue towards alleviating this issue by facilitating transfer across learning tasks, allowing us to harness related data sources or previous experience. In particular, meta-learning aims to do so by extracting prior knowledge (i.e., inductive bias) from a set of learning tasks, so that inference on a new learning task of interest is accelerated.
For example, meta-learning can be instrumental in making medical diagnoses based on imaging data such as MRI or X-ray scans where obtaining large-scale, disease-specific datasets is challenging. Here, meta-learning can be used to learn the statistical properties of the medical imaging domain from a variety of datasets. The gained knowledge about the problem domain is typically represented as some form of prior. Such a domain-specific prior would then enable us to train reliable prediction models for new diagnoses with much less data.}

The majority of existing meta-learning approaches rely on settings\added{with an abundance of related tasks/datasets that are available for meta-learning}\citep[e.g.,][]{finn2017model, garnelo2018neural}. However, in most practical settings, the number of tasks that are available for\added{meta-learning}is small. 
In such settings, we face the issue of {\em overfitting\added{on the meta-level,}}\added{i.e., overfitting to the tasks used during the meta-learning stage}\citep[cf.][]{qin2018rethink, yin2020meta}.\added{This}would impair our learning performance on yet unseen target tasks. Thus, a key question is{\em, how to regularize\added{meta-learning so that it does not overfit and}generalizes well to unseen tasks.}

PAC-Bayesian learning theory gives us a rigorous framework for reasoning about the generalization of learners \citep{mcallester1999some}. However, initial PAC-Bayesian analyses of\added{meta-learning}\citep{pentina2014pac, amit2017meta} only consider {\em bounded} loss functions, which\added{excludes}important applications such as regression or probabilistic inference, where losses are typically unbounded. More\added{importantly, their generalization bounds involve PAC-Bayesian posterior distributions for each task as well as a hyper-posterior---a distribution over meta-learning hypotheses. Obtaining each posterior in itself is a challenging stochastic optimization problem whose solution, in turn, depends on the hyper-posterior. Hence, the resulting meta-learning approaches that minimize the corresponding generalization bounds rely on solving a challenging bi-level optimization problem. This makes them}computationally much more expensive\added{and unstable}than standard meta-learning approaches.

This manuscript constitutes a significantly extended version of\addedII{\citet{rothfuss2020pacoh}} that aims to address the above-mentioned issues. First, we derive a {\em PAC-Bayesian bound} for\added{meta-learning that also holds for}\textit{unbounded loss} functions.\added{Hence, the corresponding learning guarantees apply to a much larger set of problems.}For Bayesian learners, we further tighten our PAC-Bayesian bounds, relating them directly to the generalized marginal log-likelihoods of the Bayesian model\added{instead of the posteriors. This allows us to avoid the difficult bi-level optimization. Going one step further, we present}the {\em PAC-optimal hyper-posterior (PACOH)}---the closed form of the PAC-Bayesian meta-learning problem.\added{In particular, the PACOH minimizes the upper bounds on the generalization error of meta-learning. Thus, it promises}strong performance guarantees and\added{comes with}principled meta-level regularization\added{which alleviates the aforementioned problem of overfitting.}Importantly, the {\em PACOH} can be approximated using standard variational methods \citep{Blei2016}.\added{This gives}rise to a range of {\em scalable} meta-learning algorithms\added{which we explain and compare in depth.}Furthermore, we analyze and discuss the improvement\added{of meta-learning over per-task learning within the PAC-Bayesian framework. From this, we gain useful insights about}factors that determine how much we can benefit from meta-learning. In a detailed case study on linear and logistic regression, we validate the tightness of our meta-learning bounds and empirically compare them to\added{per-task}learning bounds.

\looseness -1 In our experiments,\added{which are an extension of those in \citet{rothfuss2020pacoh},}we instantiate our framework with Gaussian Process (GP) and  Bayesian Neural Network (BNN)\added{models and empirically compare different}methods for approximating the hyper-posterior. Across several regression and classification environments, our proposed approach achieves {\em state-of-the-art} predictive {\em accuracy}, while also improving the {\em calibration} of the uncertainty estimates. Moreover, we demonstrate that,\added{through its principled regularization on the meta-level,}{\em PACOH} effectively {\em alleviates the\added{problem of overfitting on the meta-level.}}\added{This allows}us to successfully extract inductive bias from as little as five tasks while reliably reasoning about the learner's epistemic uncertainty. Thanks to these properties, {\em PACOH} can also be employed in a broad range of  {\em sequential decision problems}, which we showcase through a real-world Bayesian optimization task concerning the development of vaccines.\added{The}promising experimental results suggest that many other challenging real-world problems\added{such as molecular biology or medical imaging}may benefit from our approach as well.  
%

%% file: content/related_work.tex
\section{Related work}
\label{sec:related_work}

\looseness -1 In this section, we review the relevant literature and its connections to our work. First, we discuss the field of meta-learning followed by a second subsection on learning theory. Third, we draw connections to kernel and multi-task learning, as well as hierarchical Bayesian methods. Finally, we discuss how this paper relates to \citet{rothfuss2020pacoh} and follow-up work.

\subsection{Meta-Learning}
Meta-learning aims to extract inductive bias from a set of related tasks so that inference on a new learning task is accelerated \citep{Schmidhuber1987b, thrun1998}. For instance, a popular approach is to learn an embedding space shared across tasks\added{\citep{baxter2000model, vinyals2016matching, snell2017prototypical, goldblum2020unraveling, xu2020metafun}. Another class of meta-learning methods learns to update the model parameters \citep{bengio1991, Hochreiter2001, Andrychowicz2016, Ravi2017, Chen2017a}. Going one step further, \citet{santoro2016meta, Mishra2018, kim2019attentive} train a recurrent or attention-based model to learn the entire learning and inference process. An alternative popular approach is to learn the} initialization of a neural network so it can be quickly adapted to new tasks \citep{finn2017model, li2017meta, nichol2018firstorder, rothfuss2019promp}. 

\added{Recent}methods also use probabilistic modeling adaptation to enable uncertainty quantification \citep{yoon2018bayesian, finn2018probabilistic, garnelo2018neural, kim2019attentive}. 
Such approaches partially or fully amortize the training/inference on a target task\added{which is prone to failure cases and unpredictable behavior.}In contrast,\added{our approach learns} a prior and relies on standard methods for (PAC-)Bayesian inference\added{on the target task.}

\looseness -1 Although the above-mentioned approaches are all able to learn complex inference patterns, they rely on the abundance of meta-training tasks and fall short of providing performance guarantees. 
The issue of\added{over-fitting on the meta-level}has previously been noted \citep{qin2018rethink, fortuin2019deep, yin2020meta}. However, it still lacks a rigorous formal analysis under realistic assumptions (e.g., unbounded loss functions). Addressing this shortcoming, we study the generalization properties of meta-learners within the PAC-Bayesian framework and, based on that, contribute a novel meta-learning approach with principled meta-level regularization.

\subsection{Learning Theory}
Our work builds on PAC-Bayesian learning theory, a framework for deriving generalization bounds for randomized predictors \added{\citep{mcallester1999some, seeger2002pac, maurer2004note, catoni2007pac, alquier2008pac, germain2016pac, Alquier2016a}. Such bounds typically require}that a prior distribution over hypotheses is given exogenously.\added{Here, we refer to the PAC-Bayesian prior which differs from Bayesian priors insofar that it does not have to reflect the data-generating process. The prior has considerable influence on the tightness of PAC-Bayesian bounds. Hence, a range of works study distribution-dependent \citep{Lever2013, oneto2016pac, rivasplata2018pac} and data-dependent priors \citep{parrado12pac, dziugaite2018data, dziugaite2021role, perez2021tighter}.}

In this paper, we also study a setting where the prior is acquired in a data-driven manner. However, while data-dependent priors are\added{typcially}adjusted to the current learning task, we consider priors that are meta-learned from\added{a set of}related learning tasks. 
In that, we build on previous work that studies meta-learning in the PAC-Bayesian framework\added{\citep{pentina2014pac, amit2017meta, farid2021generalization, liu2021statistical}.}However, their PAC-Bayesian generalization bounds\added{for meta-learning}only consider bounded loss functions. More\added{importantly, they}are hard to optimize as they leave both the hyper-posterior and posterior unspecified, leading to difficult\added{bi-level}optimization problems. In contrast, our bounds also hold for unbounded losses and yield a tractable meta-learning objective without the reliance on\added{bi-level}optimization.

\added{\citet{ding2021bridging} tailor our bounds (originally introduced in \citet{rothfuss2020pacoh}) to the few-shot meta-learning setting where the number of samples per task during the meta-learning stage is much larger than for inference on the target task. Thereby, they are able to connect popular meta-learning approaches such as MAML \citep{finn2017model} and Reptile \citep{nichol2018firstorder} to the PAC-Bayesian meta-learning setting which is discussed in this paper. Further extension work provides tighter bounds based on the proof techniques of \citet{catoni2007pac}. However, these bounds no longer admit a closed-form hyper-posterior and do not translate into improved algorithms. Finally, \citet{guan2022fast} present a generic PAC-Bayesian meta-learning bound that unifies many of the mentioned meta-learning bounds, including ours.}

Other work that theoretically studies generalization in meta-learning uses covering number arguments to obtain uniform generalization bounds\added{for meta-learning}over families of hypothesis spaces \citep{baxter2000model, maurer2005algorithmic}. 
Since such approaches translate into meta-learning hypothesis spaces, it is hard to convey probabilities and uncertainties from the\added{meta-level}learner to the base learner. In contrast, our PAC-Bayesian approach\added{learns}prior distributions over learning hypotheses, thus, giving us a natural way to also improve the uncertainty estimates of downstream predictions. 

\looseness -1 A recent line of work presents information-theoretic\added{generalization bounds for meta-learning}\citep{chen2021generalization, jose2021information2, rezazadeh2021conditional, Jose2022InformationTheoreticAO}.
Unlike our PAC-Bayesian bounds which are high-probability worst-case guarantees over\added{meta-learning} tasks and data, such information-theoretic arguments depend on the task and per-task data distributions as well as the particularities of\added{the meta-learning}algorithm.


\subsection{Connections to kernel learning, multi-task learning and hierarchical Bayes} 
%
\looseness -1 Similar to the GP variant of our proposed method,\added{\citet{Ong2005, zien2007, gonen2011multiple, wilson2016deep, reeb2018learning}}learn kernels. However, while we meta-learn a kernel from multiple related tasks, such\added{works focus}on kernel learning on a single target task.\added{More recent works study kernel learning in the meta-learning setting with guarantees \citep{cella2021multi, kassraie2022meta, cella2022meta, schur2023lifelong}, but only considered linear combinations of known base kernels, restricting its generality.}

\looseness -1 Similar to our problem setting, multi-task learning aims to transfer knowledge across tasks\added{\citep[e.g.][]{micchelli2004kernels, yu2005learning, bonilla2008multi, parameswaran2010large, sener2018multi}.}Such methods perform transduction on the meta-level, typically requiring a form of task similarity. In contrast, our approach performs induction on the meta-level, trying to find a global meta-learning hypothesis, i.e., a prior that works well for all tasks from the\added{distribution over tasks.}Finally, our\added{setting}closely resembles hierarchical Bayesian models \citep[e.g.][]{salakhutdinov2012one, grant2018recasting, yoon2018bayesian, ravi2018amortized}. However, compared to such hierarchical Bayesian meta-learners which assume exact knowledge of the data-generating process, our PAC-Bayesian model makes much weaker assumptions about the loss function and hyper-prior. In addition, we provide generalization guarantees which the\added{above-mentioned works lack.}


\subsection{Relation to \citet{rothfuss2020pacoh} and follow up work.~} 
This paper is a substantially extended version of \citet{rothfuss2020pacoh}, now including a theoretical comparison to single-task learning (Section \ref{sec:method2}), in-depth case studies for linear and logistic regression (Section \ref{sec:blr_case_study}), algorithms with alternative variational approximations (Section \ref{sec:algorithm}), and additional baselines in the empirical benchmark study (Section \ref{sec:experiments}). Follow-up work of \citet{rothfuss2021meta} extends the PACOH approach to stochastic processes\added{as priors on the meta-level (i.e., hyper-priors)}and employs the resulting method towards facilitating lifelong Bayesian Optimization.\addedII{The work of} \citet{rothfuss2022meta} discusses how to ensure that the corresponding meta-learned GP priors satisfy the strict calibration requirements necessary for interactive learning under safety constraints.

%% file: content/background.tex
\section{Background: PAC-Bayesian Framework}
\label{sec:background}

\added{In this section, we explain the relevant background upon which the remainder of the paper builds. In particular, we first define basic notation and elementary concepts. Then, we briefly discuss cumulant-generating functions and provide an introduction to PAC-Bayesian learning theory. Finally, we discuss the connections and differences between the PAC-Bayesian framework and classical Bayesian inference. The expositions follow previous work such as \citet{baxter2000model, Alquier2016a, germain2016pac}}.

\subsection{Preliminaries and Notation}
A learning task is characterized by an unknown data distribution $\calD$ over a domain $\calZ$ from which we are given a set of $m$ observations $S = \left\{ z_i \right\}_{i=1}^m$, $z_i \sim \calD$. By $S \sim \calD^m$ we denote the\added{independent and identically distributed  (i.i.d.)}sampling of $m$ data points.
%
In supervised learning, we are typically concerned with pairs $z_i = (x_i, y_i)$, where $x_i \in \calX $ are observed input features and $y_i \in \calY$ are target labels.
Given a sample $S$,\added{the}goal is to find a hypothesis $h \in \calH$, typically a function $h: \calX \rightarrow \calY$ in some hypothesis space $\calH$, that enables us to make predictions for new inputs $x^* \sim \calD_x$.
The quality of the predictions is measured by a {\em loss function} $l: \calH \times \calZ \rightarrow \R$. Accordingly, we want to minimize the {\em expected error} under the data distribution, that is, $\calL(h,\calD) = \expect{z^* \sim \calD}{l(h, z^*)}$. Since $\calD$ is unknown, we typically use the {\em empirical error}  $\hat{\calL}(h, S) = \frac{1}{m} \sum_{i=1}^m l(h, z_i)$ instead.

%
In the PAC-Bayesian framework, we are concerned with {\em randomized predictors}, that is, probability measures on the hypothesis space $\calH$. This allows us to give performance guarantees for machine learning models that can also reason about the epistemic uncertainty associated with their predictions. We consider two such probability measures, the \emph{prior} $P \in \calM(\calH)$ and the \emph{posterior} $Q \in \calM(\calH)$. Here, $\calM(\calH)$ denotes the set of all probability measures on $\calH$. Note that in Bayesian inference, the prior and posterior are assumed to be tightly connected\added{through the likelihood factor, by Bayes’ theorem.}In contrast, the PAC-Bayesian framework makes fewer assumptions and only requires the prior to be independent of the observed data, while the posterior may depend on it.
For a detailed\added{discussion of the literature on PAC-Bayesian learning theory, we refer to \citet{alquier2021user}.}In the following, we overload the notation by also denoting their probability densities as $Q$ and $P$ and assume that the Kullback-Leibler (KL) divergence $D_{KL}\left( Q \| P \right)$ exists. Based on the error definitions above, we can define the so-called \emph{Gibbs error} for a randomized predictor $Q$ as $\calL(Q, \calD) = \expect{h \sim Q}{\calL(h, \calD)}$ and its empirical counterpart 
as $\hat{\calL}(Q, S) = \expect{h \sim Q}{\hat{\calL}(h, S)}$.

\subsection{Cumulant-generating functions and concentration inequalities}
Before we proceed to PAC-Bayesian theory, we\edit{introduce}{recall}the concept of the centered cumulant-generating function (CGF)\added{which is an essential part of many relevant concentration inequalities \citep[cf.][]{Boucheron2013} that are employed in learning theory.}

\begin{definition} \textbf{(Centered cumulant)}
For a random variable\added{$X \in A$}with distribution \added{$\nu\in \calM(A)$}and a real-valued function\added{$f: A \mapsto \R$,}the centered cumulant-generating function is defined as
\begin{align}
\Psi_{\nu, f(\cdot)}(t) = \log \E_{X \sim \nu} \left[ e^{t(f(X) - \E [f(X)])} \right] .
\end{align}
\end{definition}
Centered CGFs are defined as the logarithm of the centered moment generating function and quantify how much the random variable\added{$f(X)$}deviates from its mean. They are a central tool for obtaining concentration inequalities and are particularly useful when $f(X)$ is unbounded. To derive meaningful statistical concentration results for unbounded $f(X)$, we need to make additional assumptions to ensure that the tails of the probability distribution of $f(X)$ decay sufficiently fast. Such assumptions can be conveniently expressed in terms of bounds on the CGF. We will give specific examples of such assumptions in Section~\ref{sec:pac_bayes_bounds_background}.

\subsection{PAC-Bayesian Bounds} \label{sec:pac_bayes_bounds_background}

In practice, the generalization error $\calL(Q, \calD)$ is unknown. Thus, one typically resorts to empirical risk minimization (ERM), that is, optimizing $\hat{\calL}(Q, S)$ instead.
However,\added{pure ERM often results in overfitting on the training data set $S$ and poor generalization to the actual data distribution $\calD$ \citep{shalev2014understanding}. Naturally, we would like to understand factors that influence the severity of over-fitting and provide generalization guarantees. PAC-Bayesian learning theory helps us do so by bounding the unknown generalization error based on its empirical estimate:}

\begin{theorem} \added{\citep[][Theorem 3]{germain2016pac}} \label{theorem:alquier_pac_bound} 
Given a data distribution $\calD$, hypothesis space $\calH$, loss function $l(h,z)$, prior $P$, confidence level $\delta \in (0,1]$, and $\beta > 0$, with probability at least $1-\delta$ over samples $S \sim \calD^m$, we have for all $Q \in \calM(\calH)$:  \vspaceequation
\begin{equation}  \label{eq:alquier_pac_bound}
\vspaceequation
\calL(Q, \calD) \leq \hat{\calL}(Q, S) + \frac{1}{\beta} \left[ D_{KL}(Q || P) + \log \frac{1}{\delta} + \Psi(\beta, m) \right]  \;,
\end{equation}
with\added{$\Psi(\beta, m) = \log  \E_{P} \E_{\calD^m} \exp \left[ \beta \left( \calL(h, \calD) - \hat{\calL}(h, S) \right) \right] $.}
\end{theorem}
Note that the prior distribution $P$ must be independent of the data $S$ that is used to evaluate the empirical risk\addedII{$\hat{\calL}(Q, S)$.}$\Psi(\beta, m)$ is the centered cumulant-generating function of the\added{negative}empirical loss $\hat{\calL}(h,S)$ under the prior $P$ and data distribution $\calD$. Typically, $\beta$ is chosen as $\beta=\sqrt{m}$ or $\beta=m$, such that the influence of the KL-complexity term $D_{KL}(Q || P)$ decreases as the number of training points $m$ increases  \citep{germain2016pac}. However, note that, at the same time, increasing $\beta$ comes at the cost of a larger CGF $\Psi(\beta, m)$.  

Since $\Psi(\beta, n)$ contains $\calL(h, \calD)$ which is unknown in practice, Theorem~\ref{theorem:alquier_pac_bound}, as it is, does not yet provide a tractable bound. However, if we make additional assumptions about the loss function $l$, we can bound $\Psi(\beta, m)$ and thereby obtain useful PAC-Bayesian bounds.\added{Common assumptions that have been used in the PAC-Bayesian literature are: Bounded loss functions \citep{mcallester1999some, maurer2004note}, assumptions on tail behavior \citep{Alquier2016a, germain2016pac}, and
moment assumptions \citep{alquier2018simpler, holland2019pac}.
In the following, we briefly discuss how to  bound the CGF for bounded losses and under tail assumptions.}

\paragraph{Bounded loss.}
When the loss function is bounded, that is, $l:\calH \times \calZ \rightarrow [a, b]$, we can use Hoeffding's lemma to bound $\Psi(\beta, m)$. In particular, we define the random variable $l_j = \calL(h, \calD) - l(h,z_j)$ and write 
\begin{align} 
\begin{split} 
\Psi(\beta, m) = \sum_{j=1}^m \log \E \exp \left(\frac{\beta}{m} l_j \right)
\leq  \sum_{j=1}^m \log \E \exp \left(\frac{\beta^2 (b-a)^2 }{8 m^2}\right) = \frac{\beta^2 (b-a)^2 }{8 m} \;.
\end{split}
\end{align}

\paragraph{Sub-gamma loss.}
\looseness -1 A loss function $l$ is considered \emph{sub-gamma} with variance factor $s^2$ and scale parameter $c$, under a prior $P$ and data distribution $\calD$, if it can be described by a sub-gamma random variable $V := \calL(h, \calD) - l(h,z)$, i.e., its moment generating function is upper bounded by that of a Gamma distribution $\Gamma(s, c)$:
$
\log  \E_{h \sim P} \E_{z \sim \calD} \left[ e^{\lambda V} \right]
\leq \frac{\lambda^2 s^2}{2(1- c \lambda)} \quad \forall \lambda \in ( 0, 1 / c ).
$
For details, see \citet{germain2016pac} and \citet{Boucheron2013}.\added{Note that the sub-gamma assumption here is referred to as "sub-gamma on the right tail" in \citet[][chapter 2.4]{Boucheron2013}.}We can use the sub-gamma assumption with $V = l_j$ for $j=1, ..., m$ and $\lambda = \beta / m$ to bound $\Psi(\beta, m)$ as follows
\begin{equation}
\Psi(\beta, m) = \sum_{j=1}^m \log \E \exp \left(\frac{\beta}{m} l_j \right) \leq   \frac{\beta^2 s^2}{2 m (1- \frac{c \beta}{m})} \;.
\end{equation}

\paragraph{Sub-gaussian loss.} A \emph{sub-gaussian} loss function with variance $s^2$ can be considered as a limit case of the previously discussed sub-gamma assumption when $c \rightarrow 0^+$. As direct consequence, $\Psi(\beta, m)$ can be bounded by $ \Psi(\beta, m) \leq \frac{\beta^2 s^2}{2 m}$.
%
%

\subsection{Connection between the PAC-Bayesian framework and Bayesian Inference}
\looseness -1 Typically, we are interested in a posterior distribution $Q$ that promises us the\added{lowest generalization error $\calL(\calQ, \calD)$.}Thus, it is natural to use the $Q \in \calM(\calH)$ that minimizes the\added{upper}bound in (\ref{eq:alquier_pac_bound}). The following lemma gives us the  closed-form solution to such a minimization problem over $\calM(\calH)$:
\begin{lemma}\citep{catoni2007pac} \label{lemma:optimal_gibbs_posterior}
Let $\calH$ be a set, $g: \calH \rightarrow \R$ a (loss) function, $Q \in \calM(\calH)$ and $P \in \calM(\calH)$ probability densities over $\calH$. Then, for any $\beta > 0$ and $h \in \calH$, 
\vspaceequation
\begin{equation} \label{eq:gibbs_dist}
Q^*(h) :=  \frac{P(h) e^{- \beta g(h)} }{Z} = \frac{P(h) e^{- \beta g(h)}}{\E_{h \sim P} \left[ e^{- \beta g(h)} \right]} \vspaceequation
\end{equation}
is the\edit{minimizing probability density}{solution}of
$
\argmin_{Q \in \calM(\calH)} ~ \beta \E_{h \sim Q} \left[ g(h) \right] + D_{KL}(Q || P)
$.
\end{lemma}
\looseness -1 This distribution is known as the \emph{Gibbs posterior} $Q^*$ \citep{catoni2007pac, Lever2013}.
As a direct consequence of Lemma~\ref{lemma:optimal_gibbs_posterior}, for fixed $P, S, m, \delta$, we can write the minimizer of\addedII{the upper bound given in}(\ref{eq:alquier_pac_bound}) as
\begin{equation}
Q^*(h)  := ~ \argmin_{Q \in \calM(\calH)} \beta \hat{\calL}(Q,S) + D_{KL}(Q||P) 
= ~ \frac{P(h)e^{- \beta \hat{\calL}(h,S)}}{Z_\beta(S,P)}
\end{equation}
where $Z_\beta(S,P)=  \int_{\calH} P(h) e^{- \beta  \hat{\calL}(h,S)} dh$ is a normalization constant.
%
%
In a probabilistic setting, \added{the loss function is the negative log-likelihood of the data,}that is, $l(h, z_i) := - \log p(z_i | h)$.\added{As has been pointed out by \citet{germain2016pac},}in this case, the optimal Gibbs posterior coincides with the \emph{generalized Bayesian posterior} 
$
Q^*(h ; P, S) = \frac{P(h) \, p(S \given h)^{\beta / m}}{Z_\beta(S,P)}
$
where $Z_\beta(S,P) = \int_{\calH} P(h) \left( \prod_{j=1}^m p(z_j | h) \right)^{\beta / m} \, dh$ is called the \emph{generalized marginal likelihood} of the sample $S$ \citep{guedj2019primer}. For $\beta = m$ we recover the standard Bayesian posterior.

%% file: content/method1.tex
\section{PAC-Bayesian Bounds for Meta-Learning}
\label{sec:method1}

\begin{figure}
	\centering
	\includegraphics[width=0.85\linewidth]{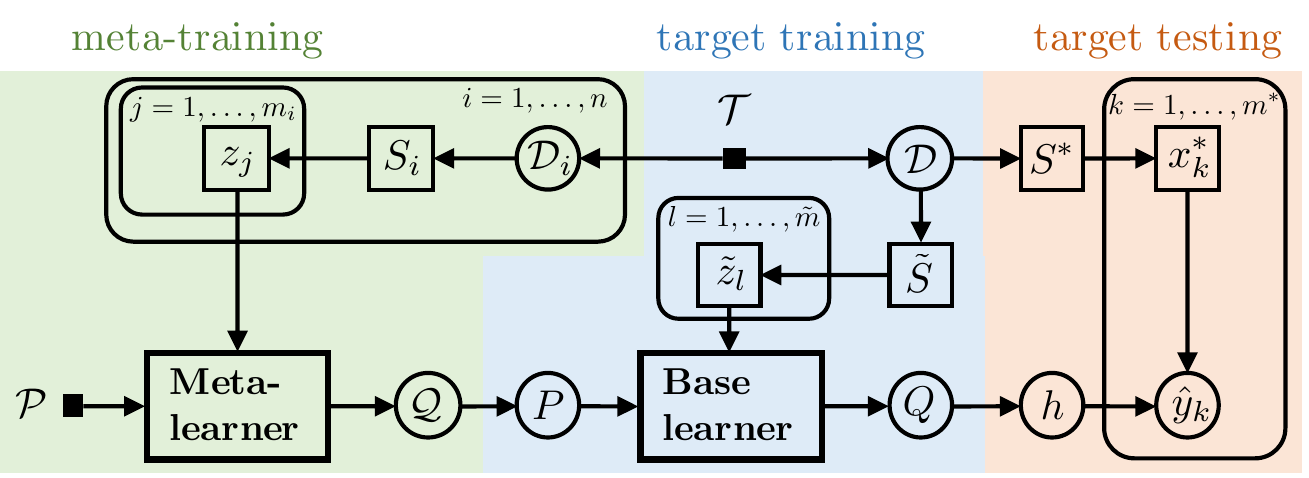}
	\caption{Overview of our meta-learning framework with environment $\mathcal{T}$, task distributions $\mathcal{D}_i$, target task distribution $\mathcal{D}$, hyper-prior $\mathcal{P}$, hyper-posterior $\mathcal{Q}$, target prior $P$, target posterior $Q$, dataset $S$, and data points $z = (x,y)$.}
	\label{fig:overview}
\end{figure}

\added{In this section, we present our main theoretical contributions. First, we describe and formalize meta-learning in the PAC-Bayesian setting, following previous work \citep{pentina2014pac, amit2017meta}. The framework is illustrated in Figure~\ref{fig:overview}.
Then, we present our PAC-Bayesian meta-learning bound and discuss how, under certain instantiations, this bound can be transformed into a useful meta-learning objective. Finally, we introduce the {\em PAC-optimal Hyper-Posterior}, the closed-form solution of our PAC-Bayesian meta-learning problem.
The corresponding proofs can be found in Appendix \ref{appendix:proofs}.}

\subsection{Meta-Learning} \label{sec:meta_learning}
\looseness -1 In the standard learning\added{setting (cf., Section \ref{sec:background}),}we assumed that the learner has prior knowledge in the form of a prior distribution $P$.\added{When the learner experiences a task, in the form of a dataset $S$, then the data are used to update the prior into a posterior $Q$. A \emph{base learner} $Q(S, P)$ can be formalized as a mapping $Q: \calZ^m \times \calM(\calH) \rightarrow \calM(\calH)$ that takes in a dataset and prior and outputs a posterior \citep{amit2017meta}.}Note that the number of samples $m$ may vary between datasets.\added{Since we have not specified the base learner in more detail, the posterior that it produces can in principle be independent of the prior $P$. However, in practice, we typically employ base learners such as Bayesian inference, or more generally, Gibbs learners that output a Gibbs posterior as in (\ref{eq:gibbs_dist}).}

\added{How well such a learner generalizes from a limited dataset $S$ to the entire data distribution $\calD$ typically strongly depends on the prior. For instance, if the prior puts a high probability on hypotheses $h$ that, a priori, describe the data well, the resulting posterior typically has a much smaller generalization error $\calL(Q, \calD)$. Hence, choosing good priors for our learning task is of great importance. However, doing so by hand is often extremely challenging---especially when dealing with complex and uninterpretable hypothesis spaces such as those of neural networks}

\added{Meta-learning offers a solution to this problem by acquiring priors}$P$ in a {\em data-driven manner}, that is, by consulting a set of $n$ statistically related learning tasks $\{\tau_1, ..., \tau_n\}$. We follow the\added{setting}of \citet{baxter2000model} in which all tasks $\tau_i := (\calD_i, S_i)$ share the same data domain $\calZ:=\calX \times \calY$, hypothesis space $\calH$ and loss function $l(h,z)$, but may differ in their (unknown) data distributions\added{$\calD_i \in \calM(\calZ)$}and the number of points $m_i$ in the corresponding dataset $S_i \sim \calD_i^{m_i}$.\footnote{\added{Note that, unlike previous work on data-dependent PAC-Bayesian priors\addedII{\citep[e.g., in][]{parrado12pac, dziugaite2018data, perez2021tighter},} our approach uses data from different tasks rather than from the same data distribution.}}
To simplify our\added{theoretical}exposition, we assume that $m_i=m ~ \forall i$.

\added{Hence, we can employ meta-learning whenever we have access to multiple related datasets, e.g., MRI images labeled for different diagnoses or robotic data collected under different configurations of the robot (e.g., different tools/payloads). Given such a set of datasets $S_1, ..., S_n$, the goal is to {\em learn a prior $P$ with which the base learner generalizes well on learning tasks of the particular problem setting} we are concerned with (e.g., MRI imaging). For this to work, we need to assume that the datasets $S_1, ..., S_n$ are representative of our problem setting which we will henceforth refer to as the {\em environment}. Formally, we assume that each task $\tau_i$ is generated i.i.d.\ by hierarchical sampling: 1) by sampling a data distribution $\calD_i \sim \calT$ from the \emph{environment} $\calT \in \calM(\calM(\calZ))$, a distribution over data distributions, and, 2) by sampling a corresponding dataset $S_i \sim \calD_i^m$. For brevity, we also refer to the distribution of this hierarchical sampling as $\calT_h \in \calM(\calM(\calZ) \times \calZ^m)$ such that $(\calD_i, S_i) \sim \calT_h$. Formally, we want to learn a prior $P$ that leads to small generalization error on}new target tasks $\tau \sim \calT_h$\added{\citep{pentina2014pac}.}

\added{Popular meta-learning approaches phrase this as a bi-level optimization problem \citep[e.g.,][]{finn2017model, amit2017meta}. That is, they optimize a prior towards yielding a small empirical error when given to the base learner. This results in a bi-level optimization since the output of the base learner is typically the solution of an optimization problem in itself that depends on the prior.}   

To extend the PAC-Bayesian analysis to the meta-learning setting, we again consider the notion of probability distributions over hypotheses.\added{The object of learning on a task has previously been a hypothesis $h \in \calH$ over which one presumes a prior distribution $P \in \calM(\calH)$ that is updated into a posterior $Q \in \calM(\calH)$ based on observed data. In meta-learning, in turn, one presumes}a {\em hyper-prior} $\calP \in \calM(\calM(\calH))$, i.e., a distribution over priors $P$. Then, combining the hyper-prior $\calP$ with the datasets $S_1, ..., S_n$ from multiple tasks,\added{the aim is to output a {\em hyper-posterior} $\calQ \in \calM(\calM(\calH))$ which can then be used to draw a prior for learning a new task.}Accordingly, the hyper-posterior's performance is measured via the expected Gibbs error when sampling priors $P$ from $\calQ$ and applying the base learner, the so-called \emph{transfer-error:}
\added{
\begin{equation} \label{eq:transfer_error}
\calL(\calQ, \calT) := \E_{P \sim \calQ}  \E_{(\calD, S) \sim \calT_h} \left[ \calL(Q(S, P), \calD) \right] 
\end{equation}}
While the transfer error is unknown in practice, we can estimate it using the \emph{empirical multi-task error}
\vspaceequation \vspaceequation
\begin{equation}
\vspaceequation 
\hat{\calL}(\calQ, S_1, ..., S_n) := \E_{P \sim \calQ} \left[  \frac{1}{n} \sum_{i=1}^n \hat{\calL} \left( Q(S_i, P), S_i \right) \right] \;.
\end{equation} 

%

\subsection{PAC-Bayesian Meta-Learning bounds}
\looseness -1 We present our first main result: An upper bound on the true transfer error $\calL(\calQ, \calT)$, in terms of the empirical multi-task error $\hat{\calL}(\calQ, S_1, ..., S_n)$ plus several tractable complexity terms.
\begin{theorem} \label{theorem:meta-learning-bound}

Let $Q: \calZ^m \times \calM(\calH) \rightarrow \calM(\calH)$ be a base learner, $\calP \in \calM(\calM(\calH))$ some fixed hyper-prior and\added{$\lambda \geq \sqrt{n}, \beta \geq \sqrt{m}$}. For any confidence level $\delta \in (0, 1]$ the inequality\vspace{-2mm}
\begin{align} \label{eq:meta-learning-bound}
\begin{split}
 \vspaceequation 
\calL(\calQ, \calT) \leq & ~ \hat{\calL}(\calQ, S_1, ..., S_n) + \left(\frac{1}{\lambda} + \frac{1}{n\beta}\right)  D_{KL}(\calQ||\calP) \\ & + \frac{1}{n} \sum_{i=1}^n \frac{1}{\beta} \E_{P \sim \calQ} \left[ D_{KL}(Q(S_i,P) || P)\right] +
\underbrace{\bar{\Psi}^\rom{1}(\beta) + \Psi^\rom{2}(\lambda) + \frac{1}{\sqrt{n}} \log \frac{1}{\delta}}_{:= ~C(\delta, \lambda, \beta)}
\end{split}
\end{align}
holds uniformly over all hyper-posteriors $\calQ \in \calM(\calM(\calH))$ with probability $1-\delta$. 
\added{Here, $\bar{\Psi}^\rom{1}(\beta)$ is an upper bound on the CGF $\Psi^{\rom{1}}(\beta) = \frac{\beta}{m} \log \E_{\calP} \E_{P} \E_{\calD} \left[  e^{ \frac{\beta}{m} (\calL(h, \calD) - l(h, z)) } \right] ~~ \forall ~\calD$ in the support of $\calT$,
and $\Psi^\rom{2}(\lambda) = \frac{n}{\lambda} \log  \E_{\calP} \E_{\calT_h} \left[ e^{ \frac{\lambda}{n} \E_{\calT_h} \E_{\calP} \left[ \calL (Q(P, S), \calD) \right] - \calL (Q(P, S), \calD ) } \right] \leq \bar{\Psi}^\rom{2}(\lambda)$. Additionally, we require that the expectation $\calL(\calQ, \calT)$ exists and is finite.}
%
\end{theorem}

\added{Next, we provide bounds on the CGFs $\Psi^\rom{1}(\beta)$ and $\Psi^\rom{2}(\lambda)$ under specific assumptions about the loss. For that, we denote corresponding upper bounds as $\bar{\Psi}^\rom{1}(\beta)$ and $\bar{\Psi}^\rom{2}(\lambda)$. If the loss function is bounded, an upper bound on the CGFs can be obtained via Hoeffding's inequality and follows \citet{pentina2014pac, amit2017meta}:}

\begin{corollary} \textbf{(Bounded loss)} \label{corr:bounded}
\added{If the loss is bounded in $[a,b]$, Theorem \ref{theorem:meta-learning-bound} holds with
\begin{equation} 
\bar{\Psi}^\rom{1}(\beta) + \bar{\Psi}^\rom{2}(\lambda) = \left( \frac{\lambda}{8n} + \frac{\beta}{8m} \right) (b - a)^2 ~. \label{eq:psi_bound_hoeffding}
\end{equation} 
as upper bounds for the CGFs.}
\end{corollary}

\added{In the case of unbounded loss functions, we require additional assumptions. Inspired by \citet{germain2016pac}, but extended to the meta-learning setting, the following corollary provides a version of Theorem \ref{theorem:meta-learning-bound} that holds under sub-gamma tail assumptions:}

\begin{corollary} \textbf{(Sub-gamma loss)} \label{corr:sub_gamma}
\added{If the loss is sub-gamma with variance factor $s_{\rom{1}}^2$ and scale parameter $c_{\rom{1}}$ under the data distributions $\mathcal{D}$ and hyper-prior $\calP$, and sub-gamma with $s_{\rom{2}}^2$, $c_{\rom{2}}$ under the task distribution $\cal{T}$ and hyper-prior $\calP$ (see Appendix \ref{appendix:proof_theorem_meta_pac_bound}, Step 3 for details) then, Theorem \ref{theorem:meta-learning-bound} holds with
\begin{equation} \label{eq:subgamma_terms_theorem_main_text}
\bar{\Psi}^\rom{1}(\beta) + \bar{\Psi}^\rom{2}(\lambda) = \frac{\beta s_{\rom{1}}^2}{2m(1- (c_{\rom{1}} \beta)/m) } + \frac{\lambda s_{\rom{2}}^2}{2n(1- (c_{\rom{2}} \lambda)/n)} ~.
\end{equation}
as upper bounds for the CGFs if $\lambda < n/c_{\rom{2}}$ and $\beta < m / c_{\rom{1}}$.}
\end{corollary}


\looseness -1 
For bounded losses, Theorem \ref{theorem:meta-learning-bound}\added{together with Corollary \ref{corr:bounded}}provides a structurally similar, but tighter bound than \citet{pentina2014pac}. In particular, by using an improved proof technique, we are able to omit a union-bound argument, allowing us to reduce the negative influence of the confidence parameter $\delta$. In contrast to \citet{pentina2014pac} and \citet{amit2017meta}, our theorem\added{together with Corollary \ref{corr:sub_gamma}}also provides guarantees for \emph{unbounded} loss functions under moment constraints (see Appendix \ref{appendix:proof_theorem_meta_pac_bound} for details). This makes Corollary \ref{corr:sub_gamma} particularly relevant for probabilistic models in which the loss function coincides with the inherently unbounded negative log-likelihood.

\looseness -1 Common choices for $\lambda$ and $\beta$ are either 1) $\lambda = \sqrt{n}$, $\beta = \sqrt{m}$ or 2) $\lambda =n$, $\beta = m$ \citep{germain2016pac}. If we choose  $\lambda = \sqrt{n}$, $\beta = \sqrt{m}$, we obtain consistent bounds, meaning that the gap between the transfer error and the bound vanishes as $n, m \rightarrow \infty$. In the second case ($\lambda =n$, $\beta = m$), the bound always maintains a gap since $C(\delta, n, m)$ does not converge to zero. However, the KL-divergence terms decay faster, which can be advantageous for smaller sample sizes. For instance, despite their lack of consistency, sub-gamma bounds with $\beta = m$ have been shown to be much tighter in simple Bayesian linear regression scenarios with limited data ($m \lesssim 10^4$) \citep{germain2016pac}.


\looseness -1 Previous work \citep{pentina2014pac, amit2017meta} proposes meta-learning algorithms that minimize uniform generalization bounds like the one in (\ref{eq:meta-learning-bound}). However, such bounds explicitly depend on the posterior $Q(S_i,P)$ which is often intractable and the solution of a non-trivial numerical optimization problem; e.g., variational inference in case of Bayesian Neural Networks. Critically, the solution of such an optimization depends on $P$ which changes during the course of meta-learning.
Thus, employing such bounds as a meta-learning objective typically results in a challenging two-level optimization problem wherein $n$ posteriors $Q_i$ and the hyper-posterior $\calQ$ need to be optimized in an interdependent manner. This becomes\added{highly intractable}to solve for rich hypothesis spaces such as neural networks.

\looseness -1 While Theorem~\ref{theorem:meta-learning-bound} holds for any base learner $Q(S,P)$,  we would preferably want to use a base learner that gives us {\em optimal performance guarantees}. As discussed in\addedII{Section~\ref{sec:background},}the Gibbs posterior not only minimizes PAC-Bayesian error bounds, but also generalizes the Bayesian posterior. Assuming a Gibbs posterior as base learner, the bound in (\ref{eq:meta-learning-bound}) can be re-stated in terms of the partition function $Z_\beta(S_i, P)$:
\begin{corollary}
\label{cor:bayesian_learner_PAC_bound}
\looseness -1 When using a Gibbs posterior $Q^*(S_i, P) := P(h) \exp (- \beta \hat{\calL}(S_i,h)) / Z_\beta(S_i, P)$ as a base learner, under the assumptions of Theorem~\ref{theorem:meta-learning-bound}, we have with probability at least $1-\delta$\vspaceequation \vspaceequation
\begin{align}  \label{eq:meta-level_pac_bound_with_mll}
\calL(\calQ, \calT) & \leq ~   - \frac{1}{n} \sum_{i=1}^n \frac{1}{\beta} \E_{P \sim \calQ} \left[\log Z_\beta(S_i, P) \right]  + \left(\frac{1}{\lambda} + \frac{1}{n\beta}\right)  D_{KL}(\calQ||\calP) + C(\delta, \lambda, \beta) \;. \hspace{-5pt} \vspaceequation
\end{align}
\end{corollary}
\added{\begin{remark}
Among all base learners, the Gibbs posterior $Q^*$ achieves the smallest possible value of the bound in (\ref{eq:meta-learning-bound}), i.e., the RHS of (\ref{eq:meta-level_pac_bound_with_mll}) is smaller or equal than the RHS of (\ref{eq:meta-learning-bound}).
\end{remark}}
Since this bound assumes a \emph{PAC-optimal base learner}, it is at least as\added{small}as the bound in (\ref{eq:meta-learning-bound}), which holds for any, potentially sub-optimal, $Q \in \calM(\calH)$. More importantly, (\ref{eq:meta-level_pac_bound_with_mll}) avoids the explicit dependence on $Q(S_i, P)$, {\em turning the previously mentioned\addedII{bi-level}optimization problem into a standard stochastic optimization problem}. Moreover, if we choose the negative log-likelihood as the loss function and\added{$\lambda=n, \beta=m$,}then $\log Z_\beta(S_i, P)$ {\em coincides with the marginal log-likelihood (MLL)}, which is tractable for various popular learning models, such as GPs.

\looseness -1 The bound in (\ref{eq:meta-level_pac_bound_with_mll}) consists of the expected generalized marginal log-likelihood under the hyper-posterior $\calQ$ as well as the KL-divergence term which serves as a {\em regularizer on the meta-level}. As the number of training tasks $n$ grows, the relative weighting of the KL term in (\ref{eq:meta-level_pac_bound_with_mll}) shrinks. This is consistent with the general notion that regularization should be strong if only little data is available and vanish asymptotically as $n, m \rightarrow \infty$. 

\addedII{Finally, Corollary \ref{cor:bayesian_learner_PAC_bound} assumes that the same loss function is used for constructing Gibbs posterior and evaluating the bound. While this is generally the case in regression, in classification, we may want to obtain a bound on the misclassification error while using the negative cross-entropy loss to form a posterior. Hence, to evaluate the bound under a different loss than used for training, we have to resort to the bound in Theorem \ref{theorem:meta-learning-bound}  that holds for arbitrary posteriors.}


\subsection{The PAC-Optimal Hyper-Posterior} 
A natural way to obtain a PAC-Bayesian meta-learning algorithm could be to minimize (\ref{eq:meta-level_pac_bound_with_mll}) with respect to $\calQ$. However, we can go one step further and derive the closed-form solution of the PAC-Bayesian meta-learning problem, i.e., the minimizing hyper-posterior $\calQ^*$.
For that, we exploit once more the insight that the minimizer of (\ref{eq:meta-level_pac_bound_with_mll}) can be written as Gibbs distribution (cf., Lemma~\ref{lemma:optimal_gibbs_posterior}), giving us the following result:
\begin{proposition} \label{proposition:pacoh_optimal_hyper_posterior}
\textbf{(PAC-Optimal Hyper-Posterior)} Given a hyper-prior $\calP$ and datasets $S_1, ..., S_n$, the hyper-posterior minimizing the meta-learning bound in (\ref{eq:meta-level_pac_bound_with_mll}) is given by \vspaceequation
\begin{equation} \label{equation:pacoh_dl_optimal_hyper_posterior}
  \calQ^*(P) = \frac{\calP(P) \exp \left( \frac{\lambda}{n\beta + \lambda}\sum_{i=1}^n \log Z_\beta(S_i, P) \right) }{Z^{\rom{2}}(S_1, ..., S_n, \calP)} \vspaceequation
\end{equation}
with
$
Z^{\rom{2}}(S_1, ..., S_n, \calP) = \E_{P \sim \calP} \left[ \exp \left(  \frac{\lambda}{n\beta + \lambda} \sum_{i=1}^n \log Z_\beta(S_i, P) \right)  \right] \;.
$ \vspaceequation
\end{proposition}
This gives us the {\em closed-form solution of the PAC-Bayesian meta-learning problem}, that is, the {\em PAC-Optimal Hyper-Posterior (PACOH)} $\calQ^*(P)$. In particular, we have a tractable expression for $\calQ^*(P)$ up to the (level-\rom{2}) partition function $Z^{\rom{2}}$, which is constant with respect to $P$. We refer to $\calQ^*$ as PAC-optimal, as it provides the best possible meta-generalization guarantees among all meta-learners in the sense of Theorem \ref{theorem:meta-learning-bound}. These best possible guarantees can be stated as follows:

\begin{corollary} \label{cor:meta_bound_pacoh}
\looseness -1 If we use the PACOH $\calQ^*$ in (\ref{equation:pacoh_dl_optimal_hyper_posterior}) as hyper-posterior, under the same assumptions as in Corollary~\ref{cor:bayesian_learner_PAC_bound}, with probability\added{$\geq 1-\delta$}, the transfer error $\calL(\calQ, \calT)$ is bounded by 
\begin{align}
\added{\calL(\calQ^*, \calT)}  &\leq - \left(\frac{1}{\lambda} + \frac{1}{n\beta}\right) \log Z^{\rom{2}}(S_1, ..., S_n, \calP) + C(\delta, \lambda, \beta) \;. \label{eq:pac_bound_z2}
\end{align}
\end{corollary}

%% file: content/method2.tex
\section{Meta-Learning vs. Per-Task Learning}
\label{sec:method2}
In this section, we aim to understand under which conditions and to what extent meta-learning improves upon per-task learning,that is, simply using the base learner on tasks individually without attempting to transfer knowledge across tasks.

A key challenge in this endeavor is that standard\added{(per-task)}learners typically use different assumptions than meta-learners. For instance, Theorem \ref{theorem:alquier_pac_bound} presumes an exogenously given prior $P \in \calM(\calH)$ whereas in meta-learning (cf. Section \ref{sec:meta_learning}) we infer such prior in an endogenous manner from a set of datasets $S_1, ..., S_n$.\addedII{Suppose we were to assume the best possible prior for per-task learning, i.e., the prior corresponding to the environment's data-generating process $\calT$.}In that case, no meta-learner can possibly improve upon per-task learning. Hence, to be instructive, we must construct our comparison in such a way that both the meta-learner and the learner start with the same prior knowledge.

\looseness -1 Instead of assuming a single prior $P$ as in Theorem \ref{theorem:alquier_pac_bound}, we now assume that the per-task learner is given a distribution over priors, i.e., a  hyper-prior $\calP$. Given priors, sampled from $\calP$, we use the PAC-optimal base learner $Q^*(S, P) := P(h) \exp (- \beta \hat{\calL}(S)i,h)) / Z(S_i, P)$ to infer the posterior for a given dataset $S$. How well this per-task PAC-Bayesian learning approach performs can be quantified by the expected generalization error on (unseen) tasks $\tau \sim \calT_h$, i.e., 
\begin{equation} \label{eq:gen_error_per_task}
\calL(\calP, \calT) := \E_{P \sim \calP}  \E_{(\calD, S) \sim \calT_h} \left[ \calL(Q(S, P), \calD) \right]  ~.
\end{equation}
Note that (\ref{eq:gen_error_per_task}) is similar to the transfer-error in (\ref{eq:transfer_error}) except that the priors are sampled from the hyper-prior $\calP$ rather than a meta-learned hyper-posterior $\calQ$.
We can bound $\calL(\calP, \calT)$ using similar proof techniques as before to obtain the following environment-level generalization bound for per-task learning:

\begin{theorem} \textbf{\added{(Per-Task}Learning Bound)}\label{theorem:single_task_bound_hyper_prior} 
Let the base learner be the Gibbs posterior\added{$Q^*(S_i, P) := P(h) \exp (- \beta \hat{\calL}(S_i,h)) / Z_\beta(S_i, P)$,}$\calP$ some hyper-prior and\added{$\lambda \geq \sqrt{n}, \beta \geq \sqrt{m}$. If we perform per-task learning with priors sampled from $\calP$, then the expected generalization error on tasks $\tau \sim \calT_h$ can be bounded by
\begin{align}  \label{eq:single_task_bound_hyper_prior}
\calL(\calP, \calT) \leq & ~
- \frac{1}{n} \sum_{i=1}^n \frac{1}{\beta} \E_{P \sim \calP} \left[  \log Z_\beta(S_i, P) \right] + C(\delta, \lambda, \beta)
\end{align}
which holds with probability at least $1-\delta$.}
%
\end{theorem}

\looseness - 1 Now, we can compare the generalization bounds for meta-learning with the\added{per-task}learning bound in Theorem~\ref{theorem:single_task_bound_hyper_prior}. For such a comparison to be insightful, the respective bounds should be asymptotically consistent. Thus, we use $\lambda=\sqrt{n}$ and $\beta = \sqrt{m}$ henceforth in this chapter.\addedII{At first glance, we observe that, unlike Corollary \ref{cor:bayesian_learner_PAC_bound}, the per-task learning bound no longer has a KL divergence term on the meta-level. This reflects the absence of meta-learning, i.e., there is no more potential overfitting of the hyper-posterior to the meta-learning tasks, which must be compensated. On the other hand, the first term in (\ref{eq:single_task_bound_hyper_prior}), i.e., the generalized MLL, is now in expectation under the fixed hyper-prior and thus cannot be reduced. Despite this intricate trade-off, we can derive the gap between the PACOH meta-learning bound in Corollary \ref{cor:meta_bound_pacoh} and the per-task learning bound in Theorem \ref{theorem:single_task_bound_hyper_prior} in closed form:}
\begin{proposition} \textbf{(Meta-learning vs.\added{per-task}learning)} \label{prop:single_vs_meta}
\looseness -1 Let $\calP$ be a hyper-prior and $S_1, ..., S_n$ datasets corresponding to tasks sampled from $\calT_h$. \added{We write $Z(S_i, P)$ short for $Z_{n^{1/2}}(S_i, P)$. The PACOH provides smaller generalization bound values than per-task learning with the Gibbs posterior (cf., Theorem \ref{theorem:meta-learning-bound}). In particular, the improvement is given by
\begin{align} \label{eq:improvement_per_task}
\begin{split}
\Delta =  & ~  (\ref{eq:single_task_bound_hyper_prior}) - (\ref{eq:pac_bound_z2}) =  \frac{1}{n} \sum_{i=1}^n \frac{\sqrt{nm}+1}{\sqrt{m}} ~ \Psi_{\calP, \sum_i \log Z(S_i, \cdot)} \left(\frac{1}{\sqrt{nm}+1}\right) \\
 = & ~ \left(\frac{1}{\sqrt{n}} + \frac{1}{n\sqrt{m}}\right) \log \E_{P \sim \calP} \left[ e^{ \frac{1}{\sqrt{nm} + 1} \sum_{i=1}^n \left( \log Z(S_i, P)   -   \E_{P \sim \calP}  \left[ \log Z(S_i, P)   \right] \right) } \right]
\end{split}
\end{align}
and\addedII{is}always non-negative, i.e., $\Delta \geq 0$.}
\end{proposition}

%


\added{\noindent Proposition \ref{prop:single_vs_meta} suggests that the PACOH meta-learning bound always improves upon the per-task learning bound.\addedII{This is natural since meta-learning with the PACOH adapts the hyper-prior into a hyper-posterior so that the corresponding PAC-Bayesian bound is minimized. To what extent the PACOH meta-learning bounds improves upon per-task learning}depends on the following key factors:
\begin{itemize}
\item Hyper-prior informativeness: The CGF in (\ref{eq:improvement_per_task}) 
can be understood as quantifying how much information the hyper-prior already conveys about the tasks $S_1, ..., S_n$, and thus about the environment $\calT$. For instance, if the hyper-prior $\calP$ has a high variance, i.e., is relatively uninformative, then the CGF will be large, suggesting that meta-learning can significantly improve upon per-task learning. On the other extreme, if the hyper-prior is a Dirac measure on a single prior, the CGF will be zero, suggesting that no improvement upon per-task learning is possible. Generally, the more uninformative the hyper-prior $\calP$, the larger the improvement $\Delta$.
\item Number of tasks $n$: $\Delta$ grows with the number of tasks $n$. This reflects that, as the meta-learning algorithm gets more training tasks, it can improve its hyper-posterior and thus improve its generalization on unseen tasks. In contrast, per-task learning benefits from additional training tasks through transfer.
\item Number of samples per task $m$: While the CGF stays roughly constant with $m$\footnote{$\log Z(S, P)$ grows at the order of $\calO(\sqrt{n})$ which cancels with $\calO(1 / \sqrt{m})$ shrinkage of the pre-factor in the exponent}, the pre-factor outside the CGF shrinks with $m$. This reflects that, the more training samples are available per task, the smaller the relative influence of the prior on the posteriors per task. Hence, it becomes increasingly hard for meta-learning to significantly improve the generalization error.
\end{itemize}}
%
In Section \ref{sec:blr_case_study}, we provide an empirical evaluation of how some of these factors affect the improvement\added{of meta-learning over per-task learning.}

Finally, we want to emphasize that Proposition \ref{prop:single_vs_meta} compares upper bounds. Hence, it does not formally guarantee the improvement of meta-learning over per-task learning in all instances. However, since both bounds were obtained with the same proof techniques, we believe that the comparison is instructive and demonstrates how meta-learning can give us better PAC-Bayesian generalization guarantees.\addedII{Moreover, the fact that the gap $\Delta$ exhibits the general behavior one would expect from a meta-learner (three factors above), suggests that it is not a mere difference of vacous bounds, but rather correlates well with the actual empirical improvement.}

%% file: content/lr_case_study.tex
\section{Case Study: Binary Classification and Linear Regression}
\label{sec:blr_case_study}
In this section, we further examine the theoretical results from Sections \ref{sec:method1} and \ref{sec:method2} in the setting of\added{binary classification and linear regression.}First, we numerically compare our bound against previous meta-learning bounds. Since the previous bounds only hold for bounded losses, we compare them in a binary classification setting where the quantity for which we want to provide guarantees is the misclassification error.
Second, we demonstrate how we can apply our PAC-Bayesian analysis from Corollary \ref{cor:meta_bound_pacoh} to unbounded loss functions such as the negative log-likelihood.\added{Inspired by \citet{germain2016pac, shalaeva2020improved},}we choose a simple linear regression setting so that we can derive\added{bounds}for cumulant-generating functions and easily compute the bounds' values.

\subsection{Binary classification}
\label{sec:logistic_regression_setup}
\paragraph{Model.} For the binary classification setting, we consider linear classifiers. Hence, the output space $\calY = \{0, 1\}$ consists of the binary labels and our hypothesis space is $\calH = \{ h_{\bw}(\bx) = \mathbbm{1}(\bw^\top \bx \leq 0) ~| ~ \bw \in \R^d \}$.
The loss we aim to bound is the misclassification error
$\tilde{l}(\bw, \bx, y) = \mathbbm{1}(h_{\bw}(\bx) \neq y) \in [0, 1]$. As prior over weights $\bw$, we use a Gaussian $P_{\mu_P, \sigma_P^2}(\bw) = \calN(\bw | \mu_P, \sigma_P^2)$ with (meta-learnable) mean vector $\mu_P \in \R^2$ and fixed variance $\sigma_P^2$. As hyper-prior we use a zero-centered Gaussian, i.e., $\calP(\mu_P) = \calN(\mu_P | \mathbf{0}, \sigma_{\calP}^2)$ with variance $\sigma_{\calP}^2$.
To construct posteriors $Q_i = Q(S_i, P)$ over the weight vectors $\mathbf{w}$, we use a logistic regression loss $l(\bw, \bx, y) = - y \log g(\bw^\top\bx) - (1-y) \log (1-g(\bw^\top\bx)) $ where  $g(\bz) := 1 / (1 + \exp(-\bz))$ is the sigmoid function. We use the corresponding Gibbs posteriors with $\beta=\sqrt{m}$:
\begin{equation}
    \log Q_i(\bw) = \log P_{\mu_P, \sigma_P^2}(\bw) + \frac{1}{\sqrt{m}} \sum_{j=1}^ml(\bw, \bx_{ij}, y_{ij}) + \text{const.}
\end{equation}
Similarly, we use the PACOH in (\ref{equation:pacoh_dl_optimal_hyper_posterior}) with $\lambda = \sqrt{n}$ as hyper-posterior.\addedII{Note that, to reflect the standard practice in classification, we use the negative log-likelihood loss for training and the misclassification error to evaluate the bound (i.e., Theorem \ref{theorem:meta-learning-bound}).}


\paragraph{Data-generating process.} Each task corresponds to a vector $\bw_i^* \in \R^d$\added{with $d=2$}that is sampled i.i.d.\ from the task distribution $\calT = \calN(\mu_\calT, \sigma^2_\calT \bI)$ with $\mu_\calT=10 \cdot \mathbf{1}$ and $\sigma_\calT= 3 \cdot \mathbf{1}$. The inputs $\bx$ are sampled i.i.d.\ from $p(\bx) = \added{\calU([-1, 1]^d)}$ and the labels are conditionally independent draws from the Bernoulli distribution $p(y=1|\bx) = g(\bw^\top\bx)$.\added{For the evaluation of the bounds we use $m=5$ data points per task.}

\begin{figure}
     \centering
          \begin{subfigure}[b]{0.48\textwidth}
         \centering
         \includegraphics[width=\textwidth]{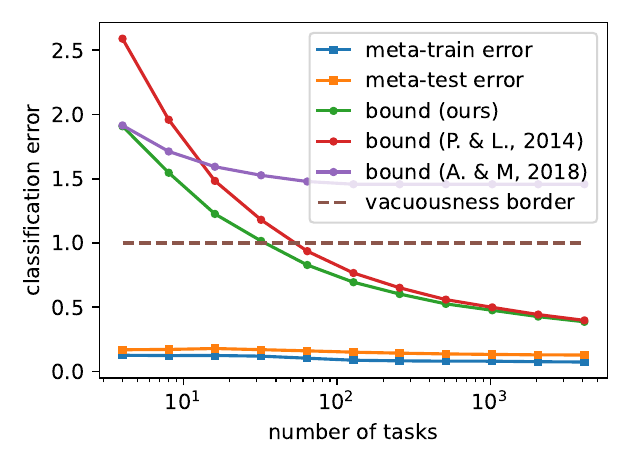}
        \vspace{-20pt}
         \caption{PAC-Bayesian Logistic Regression}
         \label{fig:logistic_regression_bounds}
     \end{subfigure}
     \hfill
     \begin{subfigure}[b]{0.48\textwidth}
         \centering
         \includegraphics[width=\textwidth]{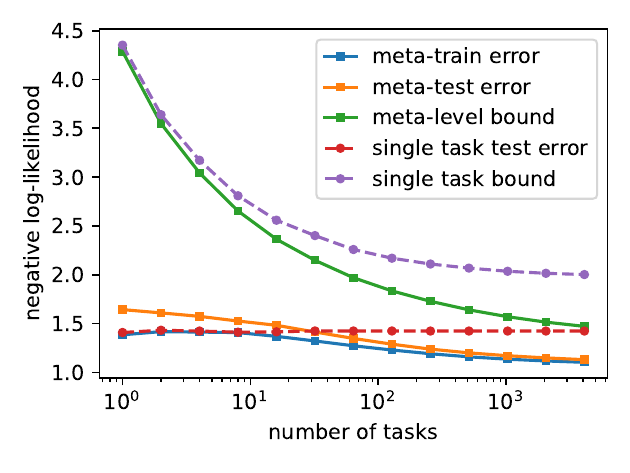}
         \vspace{-20pt}
         \caption{PAC-Bayesian Linear Regression}
         \label{fig:lin_regression_bounds}
     \end{subfigure}
     \hfill
        \caption{\looseness -1 Meta-train and meta-test error as well as corresponding worst-case bounds (e.g., Corollary \ref{cor:meta_bound_pacoh}) for $\delta=0.1$, as functions of the number of meta-training tasks $n$.Left: Comparison of our bound with previous PAC-Bayesian meta-learning bounds for classification. Compared to \citet{pentina2014pac} and \citet{amit2017meta}, the presented PACOH bound is tighter and gives non-vacuous worst-case guarantees for $n > 32$. Right: Meta-level errors and bounds for PAC-Bayesian linear regression as well as the error for single-task learning and the bound of Theorem \ref{theorem:single_task_bound_hyper_prior}. The PACOH meta-learning bound offers meaningful worst-case guarantees for the transfer error which, as $n$ grows, become tighter than bounds for per-task learning.}
        \label{fig:bound_evaluations}
\end{figure}

\paragraph{Comparison of the empirical bounds.}\added{We empirically evaluate the meta-learning bounds as well as the meta-train and -test error in the classification setting with $\sigma_P=10  \cdot \mathbf{1}$ and $\sigma_{\calP} = 20  \cdot \mathbf{1}$. Figure \ref{fig:logistic_regression_bounds} displays the comparison of the bounds and errors across a varying number of tasks $n$.}In particular, we compare the PACOH bound in Corollary \ref{cor:meta_bound_pacoh} with the meta-learning bounds of \citet{pentina2014pac} as well as \citet{amit2017meta}. In the classification setting, the misclassification loss is bounded in $[0, 1]$. Thus, a trivial upper bound is $1$ and any bound value larger than 1 is vacuous. For our setting with $m=5$ number of data points per task, the bound of \citet{amit2017meta} is always vacuous. In contrast, our bound, as well as the one of \citet{pentina2014pac}, provide non-trivial bounds ($< 1$) for more than 30 meta-training tasks. Our PACOH bound is always tighter than \citet{pentina2014pac} which is due to our slightly improved proof technique. 

\subsection{Linear regression}
\subsubsection{Setting}
\label{sec:linear_case_study_setup}
\paragraph{Model.}\added{For linear regression, the}hypothesis space is the family of linear predictors $\calH = \{ h_{\bw}(\bx) = \bw^\top \bx ~| ~ \bw \in \R^d \}$, mapping from the input space $\calX = \R^d$ to the output space $\calY = \R$. Given $(\bx, y) \in \calX \times \calY$ and the model parameters $\bw$, we consider a Gaussian likelihood with observation variance $\sigma^2 \in \R^+$, i.e., $p(y|\bx, \bw) = \calN(y|\bw^\top \bx, \sigma^2)$. Our loss function, the negative log-likelihood, is 
$l(\bw, \bx, y) = \frac{1}{2}\log(2 \pi \sigma^2) + \frac{1}{2\sigma^2}(y-\bw^\top \bx)^2$.
\added{As for the classification setting, we use a Gaussian prior $P_{\mu_P, \sigma_P^2}(\bw) = \calN(\bw | \mu_P, \sigma_P^2)$ with learnable mean $\mu_P$ and fixed variance $\sigma_P^2$ and}a Gaussian hyper-prior, $\calP(\mu_P) = \calN(\added{\mu_\calP}, \sigma^2_{\calP} \bI)$.

\paragraph{Data-generating process.}
\looseness -1 We consider a synthetic meta-learning environment where each task $\tau_i$ corresponds to a\deletedII{linear}vector $\bw_i^* \in \R^d$ that is sampled i.i.d.\ from the task distribution $\calT = \calN(\mu_\calT, \sigma^2_\calT \bI)$ with $\mu_\calT=\frac{1}{5} \cdot \mathbf{1}$ and $\sigma_\calT=\frac{1}{10} \cdot \mathbf{1} $. Each data point corresponding to the $i$-th  task is generated as follows: The input is sampled from a Gaussian $\bx \sim p(\bx) = \calN(0, \sigma_{\bx}^2 \bI)$ with $\sigma_{\bx}^2 = 1$ and the associated output is given by $y = (\bw_i^*)^{\added{\top}} \bx + \epsilon$ where $\epsilon \sim \calN(0, \sigma_{\epsilon}^2)$ is observation noise\added{standard  deviation $\sigma_{\epsilon}=\frac{1}{3}$.}Thus, we can write the conditional label distribution as $p_{\bw_i^*}(y|\bx) = \calN((\bw_i^*)^{\added{\top}} \bx, \sigma_{\epsilon}^2)$. Consequently, the data distribution follows as $\calD_i = p(\bx)  \, p_{\bw_i^*}(y|\bx)$.

\subsubsection{Bounding the cumulant-generating functions for linear regression} 
\label{sec:bounding_cgfs}
To compute the bounds in Theorem \ref{theorem:meta-learning-bound} and Corollary \ref{cor:meta_bound_pacoh}, we need to bound the cumulant-generating functions (CGFs) $\Psi^{\rom{1}}(\beta)$ and $\Psi^{\rom{2}}(\gamma)$. In the classification setting, this was straightforward since the loss function is bounded in $[0,1]$ and we could use (\ref{eq:psi_bound_hoeffding}). However, in the linear regression setting, the loss function is\addedII{typically unbounded. In particular, we use negative log-likelihood with an i.i.d. Gaussian likelihood function, constituting a generalization of the squared loss.}Hence, we need to use the particularities of our data-generating process in Section~\ref{sec:linear_case_study_setup} to derive a bound for the corresponding CGFs.
To give an intuition, the CGFs quantify the difficulty of the learning problem: $\Psi^{\rom{1}}(\beta)$ quantifies the average difficulty of the learning tasks $\tau_1, ..., \tau_n$ with respect to the prior knowledge embedded in the two-level prior, i.e., the hierarchical model over hypotheses comprised of the hyper-prior and prior. $\Psi^{\rom{2}}(\beta)$ can be understood as quantifying the difficulty of the meta-learning environment $\calT$ with respect to the hyper-prior $\calP$. 
We use the same prior and hyper-prior as in the linear regression case above.

\paragraph{Bounding $\Psi^{\rom{1}}(\beta)$:}
First, we aim to bound $\Psi^{\rom{1}}(\beta)$ which is defined as 
\begin{equation}
\Psi^{\rom{1}}(\beta) = \frac{1}{n \beta} \sum_{i=1}^n  \sum_{j=1}^{m} \underbrace{\log \E_{\calP} \E_{P} \E_{\calD_i} \left[  e^{ \frac{\beta}{m} V_{ij}^\rom{1}}\right]}_{\added{:=}\Gamma_{i}^{\rom{1}}(\beta/m)} =  \frac{m}{n \beta} \sum_{i=1}^n \Gamma_i^{\rom{1}}(\underbrace{\beta/m}_{\gamma})\:, \label{eq:Phi_1_2834}
\end{equation}
wherein $\added{V^{\rom{1}}_{ij}}= \calL(h, \calD_i) - l(h, z_{ij}) = \E_{(\bx, y) \sim \calD_i} \left[ l(\bw, \bx, y) \right]- l(\bw, \bx_{ij}, y_{ij})$.\added{Note that, when defining $\Gamma_{i}$, we omit the index $j$ since $V^{\rom{1}}_{ij}, j=1, ..., m$ are distributed identically, and, thus, $\Gamma_{i} = \Gamma_{ij} =\Gamma_{ij'}$.}Writing $\gamma := \beta/m$, we show in Appendix \ref{appendix:bounding_blr_cumulants} that the cumulant-generating function can be bounded as follows:
\begin{equation}
\Gamma_i^{\rom{1}}(\gamma) = \log \E_{\calP} \E_{P} \E_{\calD_i} e^{\gamma V^{\rom{1}}_{ij}} \leq \frac{\gamma^2 s_i^2}{2(1-\gamma c_i)} \added{ ~~~~ \forall \gamma \in (0, 1/c_i)}
\end{equation} 
with  $s_i^2 = \frac{\vartheta_i}{\sigma^2}(\frac{1}{\gamma} - c_i) +  \frac{c_i}{\gamma}$, $c_i = \frac{d}{\sigma^2} \sigma_{\bx}^2 (\sigma_P^2 + \sigma_\calP^2) + \frac{\gamma}{\sigma^4} d  \sigma_{\bx}^2 (\sigma_P^2 + \sigma_\calP^2) \vartheta_i -  \frac{\vartheta_i}{\sigma^2}$ and $\vartheta_i = \sigma_{\bx}^2 ||\bw^*_i||^2  + \sigma_\epsilon^2$. 
Thus, for $\beta = \sqrt{m}$, we have that 
$
\Phi^{\rom{1}}(1 /\sqrt{m}) \leq \frac{1}{n} \sum_{i=1}^n \frac{s_i^2}{2 (\sqrt{m} - c_i) }
$
which is a monotonically decreasing positive function of $m$.

\paragraph{Bounding $\Psi^{\rom{2}}(\lambda)$:} As shown in Appendix \ref{appendix:proof_theorem_meta_pac_bound}, 
\begin{equation} 
\Psi^\rom{2}(\lambda) = \frac{1}{\lambda}  \sum_{i=1}^n \log  \E_{\calP} \E_\calT \left[ e^{ \frac{\lambda}{n} V_i^\rom{2} } \right] = \frac{n}{\lambda} \underbrace{\log  \E_{\calP} \E_\calT \left[ e^{ \frac{\lambda}{n} V^\rom{2} } \right]}_{\Gamma^\rom{2}(\lambda / n)}
\end{equation}
is defined as the sum of the cumulant generating functions of the random variable $V^{\rom{2}}:= \\\E_{(\added{\calD},S)  \sim \calT_h} \E_{P \sim \calP} \left[ \calL (Q(P, S), \calD) \right] - \calL (Q(P, S), \calD )$. To simplify the notation, we define $\kappa := \lambda / n$. The resulting CGF $\Gamma^\rom{2}(\kappa)$ is sub-gamma, i.e.,
\begin{equation}
  \Gamma^\rom{2}(\kappa) = \log \E_{\calT} \E_{\calP} \left[ e^{\kappa V^{\rom{2}}} \right] \leq    \frac{\kappa^2 s_{\rom{2}}^2}{2(1- \kappa c_{\rom{2}})}  \added{ ~~~ \forall \kappa \in (0, 1/c_{\rom{2}})}  
\end{equation}
with $c_{\rom{2}} = \frac{\sigma_\bx^2}{\sigma^2} (\sigma^2_\calP + \sigma_\calT^2)$ and $s_{\rom{2}}^2= \frac{\sigma_\bx^2}{\sigma^2} c_{\rom{2}} ||\mu_\calT||^2 + d c_{\rom{2}}^2$. Thus, for $\lambda = \sqrt{n}$ we have 
$
\Gamma^\rom{2}(1/\sqrt{n}) \leq  \frac{s_{\rom{2}}^2}{2(\sqrt{n}-c_{\rom{2}})}
$, a monotonically decreasing positive function of $n$.

\subsubsection{Empirical evaluation of the bound values}

We now aim to inspect the bound values of the \emph{PACOH} $\calQ^*$. First, we focus on the linear regression setup. We generate our synthetic meta-learning environments as described in Section \ref{sec:linear_case_study_setup} with $d=5$ dimensional features and $m=5$ data points per task. Moreover, we set $\mu_\calP = 0$, $\sigma_\calP^2 = \frac{1}{4}$, $\sigma_P^2 = \frac{1}{25}$
and $\sigma_{\bx}^2 = 1$. Figure \ref{fig:lin_regression_bounds} plots the meta-train and meta-test (transfer) error together with our PAC-Bayesian transfer error bound of
Corollary \ref{cor:meta_bound_pacoh}. In addition, we also display the test error for\added{per-task}learning and its corresponding bound from Theorem \ref{theorem:single_task_bound_hyper_prior}. 
Note that the displayed bounds are worst-case bounds,\added{i.e., hold with high probability over the sampled tasks,}whereas the plotted errors are means. The bounds can be considered fairly tight, giving values\added{in the value range of}the generalization error they bound. As discussed in Section \ref{sec:method2}, for a small number of tasks $n$, the meta-learning bound is\added{almost identical to the per-task learner bound. However, with growing $n$, it quickly improves upon the per-task}learner bound and offers\added{much better performance guarantees}which reflect the benefits of meta-learning.  


Note that the linear and logistic regression settings which we consider here are very simplistic. The main goal of this section is to provide empirical evidence that the presented bounds are correct and behave as expected. 
\deletedII{We are aware that for large neural networks, our PAC-Bayesian bound as well as\addedII{most}previous bounds will most likely be vacuous -- a well-known problem in the PAC-Bayesian literature which we do not claim to solve.}


\subsection{Improvement of meta-learning over per-task learning bounds} \label{sec:empirical_improvement}

\begin{figure}
\centering
\includegraphics[width=1.0\textwidth]{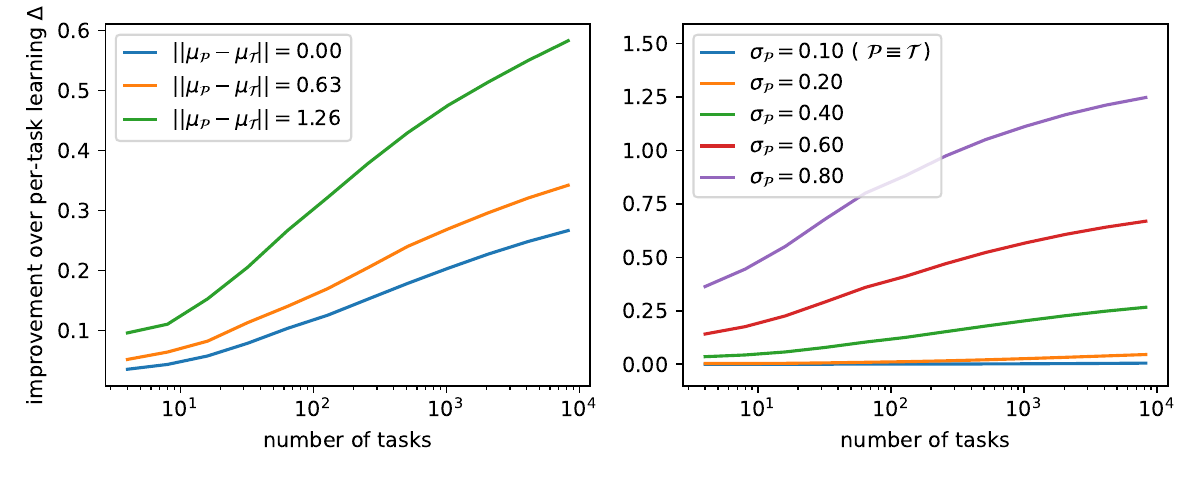}
\vspace{-10pt}
\caption{Improvement of the meta-learning bound over the per-task bound (i.e., $\Delta$ in Proposition \ref{prop:single_vs_meta}) in response to the number of meta-training tasks $n$. Left: Different degrees of misspecification of the hyper-prior mean compared to the true task distribution mean, fixed $\sigma_\calP = 0.4$. Right: Differing hyper-prior variances, fixed $\mu_\calP = 0.2$. The smaller the hyper-prior informativeness, i.e., larger $\sigma_\calP$ and/or larger mean misspecification $||\mu_\calP - \mu_\calT||$, the bigger the improvement of meta-learning over the per-task bound. \vspace{-10pt}}
\label{fig:improvement_plots}
\end{figure}

\looseness -1
Finally, we provide numerical examples of the improvement of the meta-learning over single-task learning bounds, discussed in detail in Section \ref{sec:method2}. Again, we consider the linear regression setting from Section \ref{sec:linear_case_study_setup} with $d=10$ and $m=5$ data points per task. In Figure \ref{fig:improvement_plots}, we plot the difference between meta-learning and single-task learning bounds, i.e., $\Delta$ as defined in (\ref{eq:improvement_per_task}), in response to the number of tasks $n$. In the left sub-figure, we vary the degree of misspecification in the hyper-prior mean relative to the true task distribution mean, i.e., $||\mu_\calP - \mu_\calT||$. As one would expect, the larger the misspecification of the hyper-prior, the more we can gain from meta-learning which is able to correct such misspecification by moving the hyper-posterior mean towards the true task distribution when given enough meta-training tasks. In the right sub-figure, we set $\mu_\calP = \mu_\calT$, i.e., no misspecification, but vary the hyper-prior standard deviation $\sigma_\calP$. Similarly, we observe that, the higher $\sigma_\calP$,\added{the more uninformative is the hyper-prior which leads to a}bigger improvement of the meta-learning bound over the\added{per-task}learning bound. If the variance of the hyper-prior is high, then the meta-learner already provably improves over the single task learner with only 10--20 meta-training tasks. On the other hand, if the hyper-prior is\added{equivalent}to the true task distribution, i.e., $\calP \added{\equiv} \calT$, then it is impossible to improve upon $\calP$ by meta-learning a better $\calQ$.



%% file: content/method_algorithm.tex
\section{Meta-Learning Algorithms based on the PACOH}
\label{sec:algorithm}
\looseness -1 After having introduced the closed-form solution of the PAC-Bayesian meta-learning problem in\addedII{Section~\ref{sec:method1},}we now discuss how to translate the \emph{PACOH} into a practical meta-learning algorithm when employing GPs and BNNs as base learners. For that, we assume a parametric family of priors $\{ P_\phi | \phi \in \Phi  \}$ wherein priors $P_\phi$ are governed by a parameter vector $\phi \in \Phi$. 
Accordingly, we can express the hyper-posterior and hyper-prior as distributions over prior parameters, i.e., $\calQ(\phi) \in \calM(\Phi)$ and $\calP(\phi) \in \calM(\Phi)$ respectively. Later, we will discuss in further detail our particular choice $P_\phi$ for GPs and BNNs, respectively.

\subsection{Approximating the PACOH} \label{sec:approximate_inference}
\looseness -1 Given the hyper-prior and\addedII{generalized marginal log-likelihood (MLL)}function $\log Z_\beta(S_i, P)$, we can compute the PACOH $\calQ^*$ up to the normalization constant $Z^{\rom{2}}$. Such setup lends itself to classical approximate inference methods \citep{Blei2016, Liu2016}.
Thus, Proposition~\ref{proposition:pacoh_optimal_hyper_posterior} yields an entire class of possible meta-learning methods.
We now briefly discuss three tractable approximations of $\calQ^*$ which we evaluate empirically in Section~\ref{sec:experiments}. The corresponding approximate distributions and approximate inference updates are summarized in Table \ref{table:approx_inference_methods}. 

\paragraph{Maximum A Posteriori (MAP).}
This is the simplest and most crude method, which approximates $\calQ^*(\phi)$ by a Dirac measure $\delta_P(\phi^*)$ at the mode of $\calQ^*$, i.e.,
$\phi^*=\argmax_{\phi \in \Phi} \calQ^*(\phi)$.

\paragraph{Variational Inference (VI).} In the case of VI \citep{Blei2016}, we restrict the space of considered hyper-posteriors to a parametric variational family $\calF = \{\tilde{\calQ}_\upsilon | \upsilon \in \calU \} \subset \calM(\Phi)$ wherein the variational posterior $\tilde{\calQ}_\upsilon$ is governed by a parameter vector $\upsilon \in \calU$. Then, we aim to find a $\tilde{\calQ}_\upsilon$ that minimizes the KL-divergence to $\calQ^*$, that is, 
\begin{equation} \label{eq:vi_kl}
\upsilon^* =  \argmin_{\upsilon \in \calU} D_{KL}(\tilde{\calQ}_\upsilon || \calQ^*) \;.
\end{equation}
In fact, it can be shown that the minimizing variational distribution $\tilde{\calQ}_\upsilon$ of (\ref{eq:vi_kl}) is the same as the minimizer of the bound in (\ref{eq:meta-level_pac_bound_with_mll}) under the constraint $\calQ \in \calF$ (see Appendix~\ref{appendix:proof_eqivalence_vi_pac_bound} for proof). Consequently, we can directly use (\ref{eq:meta-level_pac_bound_with_mll}) as an optimization objective. 

\paragraph{Stein Variational Gradient Descent (SVGD).}
SVGD \citep{Liu2016} approximates $\calQ^*$ as a set of $K$ particles, that is, $\tilde{\calQ} =  \frac{1}{K} \sum_{k=1}^K \delta(\phi_k)$.  Initially, we sample $K$ particles $\phi_k \sim \calP$ from the hyper-prior. For notational brevity, we stack the particles into a $K \times \text{dim}(\phi)$ matrix $\bm{\phi} :=[\phi_1, ..., \phi_K]^\top$. Then, the method iteratively transports the set of particles to match $\calQ^*$, by applying a form of functional gradient descent that minimizes  $D_{KL}(\tilde{\calQ} | \calQ^*)$ in the reproducing kernel Hilbert space induced by a kernel function $k(\cdot,\cdot)$. 
In particular, in each iteration, we update the particle matrix using the SVGD update rule:   \vspaceequation
\begin{equation}
    \bm{\phi} \leftarrow \bm{\phi} + \eta ~ \mathbf{K} ~ \nabla_{\bm{\phi}} \mathbf{ln} \calQ^* + \nabla_{\bm{\phi}} \mathbf{K}  \vspaceequation
\end{equation}
where $\nabla_{\bm{\phi}} \mathbf{ln} \calQ^*:= [\nabla_{\phi_1} \log \calQ^*(\phi_1), ..., \nabla_{\phi_K} \log \calQ^*(\phi_K) ]^\top$ denotes the matrix of stacked score gradients, $\mathbf{K} := [k(\phi_k, \phi_{k'})]_{k,k'}$ the kernel matrix induced by a kernel function $k(\cdot,\cdot)$ and $\eta$ the step size for the SVGD updates.

\begin{table} 
\centering
\resizebox{0.81\columnwidth}{!}{
\begin{tabular}{c|c|c}
\toprule
& Approximating Distribution & \texttt{Init\_Approx\_Inference} \\ \hline \\[-1.0em]
 MAP & $\tilde{\calQ} = \delta(\tilde{\phi})$ & $ \tilde{\phi} \sim \calP$ \\[0.5em]
SVGD & $\tilde{\calQ} = \frac{1}{K} \sum_{k=1}^K \delta(\tilde{\phi}_k), ~ \tilde{\bm{\phi}}=\left[\tilde{\phi}_1, ..., \tilde{\phi}_K \right]^\top$& $\tilde{\phi}_k \sim \calP, ~k=1, ..., K$ \\[0.5em]

VI & $\tilde{\calQ}_\upsilon = \calN(\mu_\calQ, \sigma^2_\calQ) ,  ~ \upsilon = (\mu_\calQ, \sigma^2_\calQ)$  & $\upsilon = (\mu_\calP, \sigma^2_\calP) $\\[0.5em] \bottomrule
\end{tabular}} \vspace{12pt}

\resizebox{0.95\columnwidth}{!}{
\begin{tabular}{c|c|c}
\toprule
& \texttt{Sample\_Prior\_Params} & \texttt{Approx\_Inference\_Update} \\ \hline \\[-1.0em]
 MAP & $\phi \leftarrow \tilde{\phi}$ & $ \tilde{\phi} \leftarrow \tilde{\phi} + \eta \nabla_{\phi} \log \calQ^*(\tilde{\phi})$ \\[0.5em]
SVGD & $\phi_k \leftarrow \tilde{\phi}_k $ & $\tilde{\bm{\phi}} \leftarrow \tilde{\bm{\phi}} + \eta ~ \mathbf{K} ~ \nabla_{\bm{\phi}} \mathbf{log} \tilde{\calQ}^* + \nabla_{\bm_{\phi}} \mathbf{K}$ \\[0.5em]

VI & $\phi_k \leftarrow \mu_\calQ + \sigma_\calQ \odot \epsilon , \epsilon \sim \calN(0, \mathbf{I})$  & $\upsilon \leftarrow \upsilon  + \frac{\eta}{K} \sum_{k=1}^K  \nabla_\upsilon \left[ \log \calQ^*(\phi_k) - \log \tilde{\calQ}_\upsilon(\phi_k) \right]$\\[0.5em] \bottomrule
\end{tabular}}
\vspace{-7pt}
\caption{Summary of approximations of the \emph{PACOH} $\calQ^*$ \label{table:approx_inference_methods}}
\end{table}

%

\subsection{Meta-Learning Gaussian Process Priors}
\paragraph{Setup.} In GP regression, each data point corresponds to a tuple $z_{i,j} = (x_{i,j},y_{i,j})$. For the $i$-th dataset, we write $S_i = (\bX_i, \by_i)$, where $\bX_i = (x_{i,1}, ..., x_{i,m_i})^\top$ and $\by_i = (y_{i,1}, ..., y_{i,m_i})^\top$. GPs are a Bayesian method in which the prior $P_\phi(h) = \mathcal{GP} \left(h| m_\phi(x), k_\phi(x, x') \right)$ is specified by a kernel $k_\phi: \calX \times \calX \rightarrow \R$ and a mean function $m_\phi: \calX \rightarrow \R$. 

Since we are interested in meta-learning such prior, we define the mean and kernel function both as parametric functions.
Similar to \citet{wilson2016deep} and \citet{fortuin2019deep}, we instantiate $m_\phi$ and $k_\phi$ as neural networks, and meta-learn the parameter vector $\phi$.
To ensure the positive-definiteness of the kernel, we use the neural network as feature map $\varphi_\phi(x)$ on top of which we apply a squared exponential (SE) kernel.
Accordingly, the parametric kernel reads as $k_\phi(x, x') = \frac{1}{2}\exp \left( - ||\varphi_\phi(x) - \varphi_\phi(x')||_2^2 \right)$. 

As typical for GPs, we assume a Gaussian likelihood $p(\by|\mathbf{h}) = \calN(\by;h(\bx), \sigma^2 I)$ where $\sigma^2$ is observation noise variance.
Moreover, we choose $\lambda = n$, $\beta = m$, so that the closed-form GP posterior coincides with the PAC-optimal posterior $Q^*$. In this case, the empirical loss under the GP posterior $Q^*$ coincides with the negative log-likelihood of regression targets $\by_i$, that is,  $\hat{\calL}(Q^*, S_i) = - \frac{1}{m_i} \log p(\by_i|\bX_i)$. Finally, we use a Gaussian hyper-prior $\calP = \calN(0, \sigma^2_{\calP} I)$ over the GP prior parameters $\phi$.

\paragraph{Algorithm.} 
Depending on which approximate inference method from Section \ref{sec:approximate_inference} we choose, the resulting meta-learning algorithms differ slightly. For this reason, we describe the algorithm in terms of generic operations that are specified in Table \ref{table:approx_inference_methods}.

First, we initialize the approximate hyper-posterior $\tilde{\calQ}$.
Then, in each iteration, we sample $K$ prior parameters\footnote{Note that for MAP inference, we always have $K=1$.} $\phi_k$ from $\tilde{\calQ}$, and compute the corresponding score gradients of $\calQ^*$, 
\begin{align} 
\nabla_{\phi_k} \log \calQ^*(\phi_k) = \nabla_{\phi_k} \log \calP(\phi_k) +  \sum_{i=1}^n  \frac{1}{m_i + 1} \nabla_{\phi_k} \log Z_{m_i} (S_i, P_{\phi_k}) \;.
\end{align}
In our setup, $\log Z_{m_i}(S_i, P_\phi) = \log p(\by_i | \bX_i, \phi)$ is the MLL of the GP which can be computed in closed form, in particular,
\begin{align} \label{eq:mll_gp}
\log p(\by | \bX, \phi) = & - \frac{1}{2} \left( \by- m_{\bX,\phi}) \right)^\top \tilde{K}_{\bX,\phi}^{-1}  \left( \by- m_{\bX,\phi} \right) - \frac{1}{2} \log |\tilde{K}_{\bX, \phi}| - \frac{m_i}{2} \log 2 \pi \:,
\end{align}
where $\tilde{K}_{\bX, \phi} = K_{\bX, \phi} + \sigma^2 I$, with the kernel matrix $K_{\bX, \phi}=(k_\phi(x_l,  x_k))^{m_i}_{l,k=1}$ and mean vector $m_{\bX,\phi} = (m_\phi(x_1),..., m_\phi(x_{m_i}))^\top$. Based on the score gradients $\nabla_{\phi_k} \log \calQ^*(\phi_k)$, we can then update the approximate hyper-posterior $\tilde{Q}$, using the corresponding approximate inference method.  Algorithm~\ref{algo:pacoh_gp} summarizes the resulting generic meta-learning procedure for GPs which we henceforth refer to as \emph{PACOH-GP}.

\begin{algorithm}[t]
\caption{PACOH-GP - Approximate inference of $\calQ^*$}
\label{algo:pacoh_gp}
\begin{algorithmic}
\STATE \textbf{Input:} hyper-prior $\calP$, datasets $S_1, ..., S_n$,  step size $\eta$
\STATE $\tilde{\calQ} \leftarrow \texttt{Init\_Approx\_Inference}()$  \hfill
\WHILE{not converged} 
	\STATE $\{ \phi_1, ..., \phi_K \} \leftarrow \texttt{Sample\_Prior\_Params}(\tilde{\calQ})$
 	\FOR{$k=1,...,K$} 
 	    \FOR{$i=1,...,n$} 
 	        \STATE \added{$\log Z_{m_i}(S_i, P_{\phi_k}) \leftarrow \log p(\by_i | \bX_i, \phi_k)$}
 	    \ENDFOR
     	\STATE $  \nabla_{\phi_k} \log \calQ^*\leftarrow \nabla_{\phi_k} \log \calP +  \sum_{i=1}^{n}  \frac{1}{m_i + 1} \nabla_{\phi_k}  \log Z_{m_i}(S_i, P_{\phi_k})$
     \ENDFOR
	\STATE $\tilde{\calQ} \leftarrow \texttt{Approx\_Inference\_Update}(\tilde{\calQ}, \nabla_{\phi_k} \log \calQ^*, \eta)$ 
\ENDWHILE
\STATE \textbf{Output:} Approximate hyper-posterior $\tilde{\calQ}$ 
\end{algorithmic}
\end{algorithm}


\subsection{Meta-Learning Bayesian Neural Network Priors}

\paragraph{Setup.} Let $h_\theta: \calX \rightarrow \calY$ be a function parametrized by a neural network (NN) with weights $\theta \in \Theta$. Using the NN mapping, we define a conditional distribution $p(y|x,\theta)$. For regression, we may set $p(y|x,\theta) = \calN(y|h_\theta(x), \sigma^2)$, where $\sigma^2$ is the observation noise variance. We treat $\log \sigma$ as a learnable parameter similar to the neural network weights $\theta$ such that a hypothesis coincides with a tuple $h = (\theta, \log \sigma)$.
For classification, we choose $p(y|x,\theta) = \mathrm{Categorical}(\mathrm{softmax}(h_\theta(x)))$. Our loss function is the negative log-likelihood $l(\theta,z) = - \log p(y|x, \theta)$.

\looseness -1 Next, we define a family of priors $\{P_\phi: \phi \in \Phi\}$ over the NN parameters $\theta$. For computational convenience, we employ diagonal Gaussian priors, that is, $P_{\phi_l} = \calN(\mu_{P_k}, \text{diag}(\sigma_{P_k}^2))$ with $\phi:= (\mu_{P_k}, \log \sigma_{P_k})$. Note that we represent $\sigma_{P_k}$ in the log-space to avoid additional positivity constraints. In fact, any parametric distribution that supports re-parametrized sampling and has a tractable log-density (e.g., normalizing flows \citep[cf.,][]{rezende2015variational}) could be used. 
Moreover, we use a zero-centered, spherical Gaussian hyper-prior $\calP := \calN(0,\sigma_{\calP}^2 I)$ over the prior parameters $\phi$.

\paragraph{Approximating the marginal log-likelihood.} 
Unlike for GPs, the (generalized) MLL $\log Z_\beta(S_i, P_\phi) = \log \E_{\theta \sim P_\phi} e^{- \beta_i \hat{\calL}(\theta, S_i)}$ is intractable for BNNs. Estimating and optimizing $\log Z(S_i, P_\phi)$ is not only challenging due to the high-dimensional expectation over $\Theta$ but also due to numerical instabilities inherent in computing $e^{- \beta_i \hat{\calL}(\theta,S_i)}$ when $m_i$ and, thus, $\beta_i$ is large.
\looseness -1 Aiming to overcome these issues, we compute numerically stable Monte Carlo estimates of $\nabla_{\phi} \log Z_{\beta_i}(S_i, P_{\phi_k})$ by combining the LogSumExp (LSE) with the re-parametrization trick \citep{kingma2014auto}. In particular, we draw $L$ samples $\theta_l := f(\phi_k, \epsilon_l) = \mu_{P_k} + \sigma_{P_k} \odot \epsilon_l, ~ \epsilon_l \sim N(0,I) $ and compute the generalized MLL estimate as 
\begin{equation} \label{eq:mll_estimator}
\log \hat{Z}_{\beta_i}(S_i, P_\phi) := ~ \text{LSE}_{l=1}^L\left( - \beta_i  \hat{\calL}(\theta_l, S_i) \right) - \log L  \;.
\end{equation} 
The corresponding gradients follow a softmax-weighted average of score gradients:
\begin{equation} \label{eq:mll_estimator_grad}
\nabla_{\phi} \log \hat{Z}_{\beta_i}(S_i, P_\phi) = - \beta_i \sum_{l=1}^L ~ \underbrace{\frac{e^{- \beta_i  \hat{\calL}(\theta_l,S_i)}}{\sum_{l=1}^L e^{- \beta_i  \hat{\calL}(\theta_l,S_i)} }}_{\text{softmax}}  \underbrace{\nabla_\phi f(\phi, \epsilon_l)^\top}_{\substack{\text{re-param.}\\\text{Jacobian}}} \underbrace{\nabla_{\theta_l} \hat{\calL}(\theta_l,S_i)}_{\text{score}}  \vspaceequation
\end{equation}
Note that $\log \hat{Z}(S_i, P_\phi)$ is a consistent but not an unbiased estimator of $\log Z(S_i, P_\phi)$. The following proposition ensures us that we still minimize a valid upper bound:
\begin{proposition} \label{proposition:mll_estimate_still_upper_bound}
In expectation, replacing $\log Z(S_i, P_\phi)$ in (\ref{eq:meta-level_pac_bound_with_mll}) with the Monte Carlo estimate $\log \hat{Z}(S_i, P_\phi)$ still yields a valid upper bound of the transfer error $\calL(\calQ, \calT)$ for any $L \in \mathbb{N}$, i.e., 
\begin{align*}
\begin{split} 
\calL(\calQ, \calT) \leq (\ref{eq:meta-level_pac_bound_with_mll}) 
 \leq & - \frac{1}{n} \sum_{i=1}^n \frac{1}{\beta} \E_{\calQ} \left[ \E_{\theta_1,...,\theta_L \sim P} \left[ \log \hat{Z}_{\beta}(S_i, P_\phi)  \right] \right] 
  \hspace{-1pt} + \hspace{-1pt} \left(\frac{1}{\lambda} + \frac{1}{n \beta}\right) \hspace{-2pt} D_{KL}(\calQ||\calP) \hspace{-1pt} + \hspace{-1pt} C \;.
\end{split}
\end{align*}
\end{proposition}
Moreover, by the law of large numbers, we have that $\log \hat{Z}(S_i, P)  \xrightarrow[]{\text{a.s.}} \log Z(S_i, P)$ as $L \rightarrow \infty$, that is, for large sample sizes $L$, we recover the original PAC-Bayesian bound in (\ref{eq:meta-level_pac_bound_with_mll}). In the opposite edge case, i.e., $L=1$, the boundaries between tasks vanish meaning that the meta-training data $\{S_1, ..., S_n\}$ is treated as if it were one large dataset $\bigcup_i S_i$ (see Appendix \ref{appendix:properties_mll_estimator} for further discussion). As an alternative to our proposed MLL estimator in (\ref{eq:mll_estimator}), we could use a Russian roulette estimator \citep{kahn1995, luo2020sumo} which would allows us to construct unbiased Monte Carlo estimates of the generalized MLL. However, albeit unbiased, Russian roulette estimators suffer from high variance and are hard to implement efficiently in parallel \citep{matt2016}. Thus, we found (\ref{eq:mll_estimator}) to be the more favorable choice in practice.

\begin{algorithm}[t]
\caption{PACOH-NN - Approximate inference of $\calQ^*$}
\label{algo:pacoh_nn}
\begin{algorithmic}
\STATE \textbf{Input:} hyper-prior $\calP$, datasets $S_1, ..., S_n$,  step size $\eta$
\STATE $\tilde{\calQ} \leftarrow \texttt{Init\_Approx\_Inference}()$  \hfill
\WHILE{not converged} 
	\STATE $\{ \phi_1, ..., \phi_K \} \leftarrow \texttt{Sample\_Prior\_Params}(\tilde{\calQ})$
 	\FOR{$k=1,...,K$} 
 	\STATE $\{\theta_1, ..., \theta_L\} \sim P_{\phi_k}$ \hfill // sample NN-parameters from prior
 	    \FOR{$i=1,...,n$} 
 	                 	\STATE $\log \hat{Z}(S_i, P_{\phi_k}) \leftarrow \text{LSE}_{l=1}^L\left( - \beta_i \hat{\calL}(\theta_l,S_i) \right) - \log L$ \hfill // estimate generalized MLL
 	    \ENDFOR
     		\STATE $  \nabla_{\phi_k} \log \hat{\calQ}^*(\phi_k)  \leftarrow \nabla_{\phi_k} \log \calP(\phi_k) + \sum_{i=1}^n \frac{\lambda}{n\beta_i + \lambda}   \nabla_{\phi_k} \log \hat{Z}_{\beta_i}(S_i, P_{\phi_k})$ \hfill // compute score
     \ENDFOR
	\STATE $\tilde{\calQ} \leftarrow \texttt{Approx\_Inference\_Update}(\tilde{\calQ}, \nabla_{\phi_k} \log \calQ^*(\phi_k), \eta)$ 
\ENDWHILE
\STATE \textbf{Output:} Approximate hyper-posterior $\tilde{\calQ}$ 
\end{algorithmic}
\end{algorithm}

\paragraph{Algorithm.} \looseness -1  Algorithm \ref{algo:pacoh_nn} summarizes the proposed meta-learning method, henceforth referred to as \emph{PACOH-NN}. It follows similar steps as \emph{PACOH-GP} but uses the proposed generalized MLL estimator $\log \hat{Z}_{\beta_i}(S_i, P_\theta)$ instead of a closed-form solution. To estimate the hyper-posterior score 
$
\nabla_{\phi_{k'}} \log \hat{\calQ}^*(\phi_{k'}) = \nabla_{\phi_k} \log \calP(\phi_k) + \sum_{i=1}^n \frac{\lambda}{n\beta_i + \lambda}   \nabla_{\phi_k} \log \hat{Z}_{\beta_i}(S_i, P_{\phi_k}),
$
we can use mini-batching over tasks, only estimating the generalized MLL for a random subset of $n_{bs} \leq n$ tasks and adjusting the (truncated) sum of MLL gradients by the factor $\frac{n}{n_{bs}}$.

If we use the mini-batched version in conjunction with the SVGD approximate inference, the resulting algorithm (see Algorithm \ref{algo:pacoh_nn_batched} in Appendix \ref{appendix:pacoh_nn})
maintains $K$ particles to approximate the hyper-posterior, and in each forward step samples $L$ NN-parameters (of dimensionaly $|\Theta|$) per prior particle that are deployed on a mini-batch of $n_{bs}$ tasks to estimate the score of $\calQ^*$. As a result, the total space complexity is on the order of $\mathcal{O}(|\Theta|K + L)$ and the computational complexity of the algorithm for a single iteration is $\mathcal{O}(K^2 + K L n_{bs})$.

A key advantage of \emph{PACOH-NN} over previous methods for meta-learning BNN priors \citep[e.g.][]{pentina2014pac, amit2017meta} is that it turns the previously nested optimization problem into a much simpler stochastic optimization problem. This makes meta-learning not only much more stable but also more scalable. In particular, we do not need to explicitly compute the task posteriors $Q_i$ and can use mini-batching over tasks. As a result, the space and compute complexity do not depend on the number of tasks $n$. In contrast, \emph{MLAP} \citep{amit2017meta} has a memory footprint of $\mathcal{O}(|\Theta|n)$ making meta-learning prohibitive for more than 50 tasks.

\looseness -1 A central feature of \emph{PACOH-NN} is that is comes with principled meta-level regularization in form of the hyper-prior $\calP$, which combats overfitting to the meta-training tasks \citep{qin2018rethink}. As we show in our experiments, this allows us to successfully perform meta-learning with as little as 5 tasks. This is unlike the majority of popular meta-learners \citep[e.g.]{finn2017model, yoon2018bayesian, garnelo2018neural}, which rely on a large number of tasks to generalize well on the meta-level \citep{qin2018rethink, rothfuss2020pacoh}.

\clearpage

%% file: content/experiments.tex
\section{Experiments}
\label{sec:experiments}


\looseness -1 We now empirically evaluate the \emph{PACOH} algorithms introduced in\addedII{Section~\ref{sec:algorithm}.}In particular, we evaluate the \emph{PACOH} approach with GPs and NNs as base learners, i.e., \emph{PACOH-GP} and \emph{PACOH-NN}, as well as the different variational approximations of $\calQ$, i.e., \emph{MAP, SVGD} and, in case of PACOH-GP, also \emph{VI}. 
Comparing them to existing meta-learning approaches on various regression and classification environments, we demonstrate that our \emph{PACOH}-based methods: (i) outperform previous meta-learning algorithms in {\em predictive accuracy}, (ii) improve the calibration of {\em uncertainty estimates}, (iii) are much more {\em scalable} than previous PAC-Bayesian meta-learners, and (iv) effectively combat {\em meta-overfitting}. Finally, we showcase how meta-learned \emph{PACOH-NN} priors can be harnessed in a real-world {\em sequential decision making} task concerning peptide-based vaccine development.

\vspacesubcaption
\subsection{Experiment Setup} \label{sec:exp_setup}
\vspacesubcaption
\paragraph{Baselines.} 
We use a \emph{Vanilla GP} with squared exponential kernel and a \emph{Vanilla BNN} with a zero-centered, spherical Gaussian prior and SVGD posterior inference \citep{Liu2016} as baselines.
Moreover, we compare our proposed approach against various popular meta-learning algorithms, including model-agnostic meta-learning (\emph{MAML}) \citep{finn2017model}, Bayesian MAML (\emph{BMAML}) \citep{yoon2018bayesian} and the PAC-Bayesian approach by \citet{amit2017meta} (\emph{MLAP}). For the regression task experiments, we also compare with two versions of \emph{MR-MAML} \citep{yin2020meta} which add an information bottleneck regularization to the NN weights (W) or the activations (A) to prevent meta-overfitting in MAML. Additionally, we consider neural processes (\emph{NPs}) \citep{garnelo2018neural} and a GP with neural-network-based mean and kernel function, meta-learned by maximizing the marginal log-likelihood (\emph{MLL-GP}) \citep{fortuin2019deep}.
Among all, \emph{MLAP} is the most similar to our approach as it is neural-network-based and minimizes PAC-Bayesian bounds on the transfer error. However, unlike \emph{PACOH-NN}, it relies on nested optimization of the task posteriors $Q_i$ and the hyper-posterior $\calQ$. \emph{MLL-GP}  is similar to \emph{PACOH-GP} insofar that it also maximizes the sum of marginal log-likelihoods $\log Z_m(S_i, P_\phi)$ across tasks. However, unlike \emph{PACOH-GP}, it lacks any form of meta-level regularization.

\paragraph{Regression environments.} 
\looseness -1 We consider two synthetic and four real-world meta-learning environments for \emph{regression}. As synthetic environments, we employ \emph{Sinusoids} of varying amplitude, phase, and slope as well as a 2-dimensional mixture of \emph{Cauchy} distributions plus random GP functions. 
As real-world environments, we use datasets corresponding to different calibration sessions of the Swiss Free Electron Laser (\emph{SwissFEL}) \citep{milne2017swissfel, kirschner2019swissfel}, as well as data from the \emph{PhysioNet} 2012 challenge, which consists of time series of electronic health measurements from intensive care patients \citep{silva2012predicting}, in particular, the Glasgow Coma Scale (\emph{GCS}) and the hematocrit value (\emph{HCT}). Here, the different tasks correspond to different patients. Moreover, we employ the Intel Berkeley Research Lab temperature sensor dataset (\emph{Berkeley-Sensor}) \citep{intel_sensor_data} where the tasks require auto-regressive prediction of temperature measurements corresponding to sensors installed in different locations of the building. Further details can be found in Appendix~\ref{appendix:meta-envs}.

\begin{figure}
    \centering
    \vspace{-4pt}
    \includegraphics[trim={1.5cm 0 1.1cm 0}, width=0.99\textwidth]{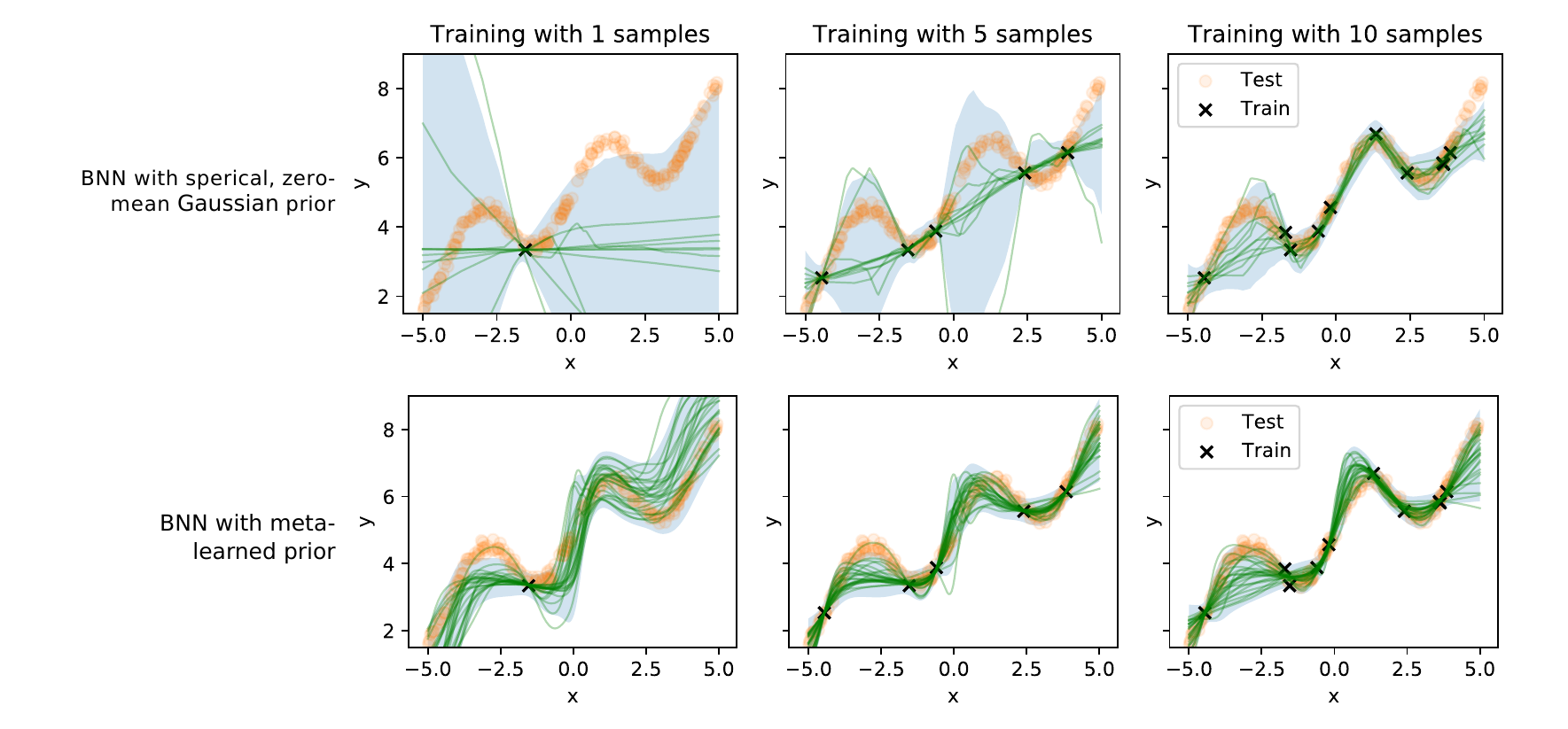} 
    \vspace{-8pt}
    \caption{\looseness -1 BNN posterior predictions with (top) standard Gaussian prior vs. (bottom) meta-learned prior. Meta-learning conducted on the \emph{Sinusoids} environment. The meta-learned PACOH-NN-SVGD prior conveys useful inductive bias, improving the BNN predictions. \vspace{-4pt}}
    \label{fig:sine_ilustrations}
\end{figure}

\paragraph{Classification environments.}
We conduct experiments with the multi-task \emph{classification} environment \emph{Omniglot} \citep{lake2015human}, consisting of handwritten letters across 50 alphabets. Unlike previous work \citep[e.g.,][]{finn2017model}, we do not perform data augmentation and do not recombine letters of different alphabets. This preserves the data's original structure, where each task corresponds to classifying letters within an alphabet. In particular, one task corresponds to 5-way 5-shot classification of letters within an alphabet. This leaves us with much fewer tasks (30 train, 20 test tasks), making the environment more challenging and interesting for uncertainty quantification. This also allows us to include \emph{MLAP} in the experiment, which hardly scales to more than 50 tasks. 


\vspacesubcaption
\subsection{Experiment Results} \label{sec:exp_reg_class}
\vspacesubcaption

\paragraph{Qualitative example.} Figure \ref{fig:sine_ilustrations} illustrates \emph{BNN} predictions on a sinusoidal regression task with a standard Gaussian prior as well as a \emph{PACOH-NN} prior meta-learned with 20 tasks from the \textit{Sinusoids} environment. We observe that the standard Gaussian prior provides poor inductive bias, not only leading to bad mean predictions away from the test points but also to poor 95\% confidence intervals (blue shaded areas). In contrast, the meta-learned \emph{PACOH-NN} prior encodes useful inductive bias towards sinusoidal function shapes, leading to better predictions and uncertainty estimates, even with minimal training data.

\input{tables/avg_rmse_table.tex}

\input{tables/cal_err_regression_table.tex}

\paragraph{PACOH improves the predictive accuracy.}
\looseness -1 Using the meta-learning environments and baseline methods that we introduced in\added{Section~\ref{sec:exp_setup},}we perform a comprehensive benchmark study. Table~\ref{tab:reg_rmse} reports the results on the regression environments in terms of the root mean squared error (RMSE) on unseen test tasks. Table~\ref{tab:classification} reports the classification accuracy for the Omniglot classification environments. For the image classification problems, we only consider the neural network-based methods, since GPs generally perform poorly on high-dimensional problems. 
When comparing the different variational approximations of the PACOH, the SVGD approximation gives the lowest meta-test errors in the majority of environments.
Throughout the benchmark study, \emph{PACOH-NN} and \emph{PACOH-GP} {\em consistently perform best or are among the best methods}. Similarly, \emph{PACOH-NN} achieves the {\em highest accuracy} in the Omniglot classification environment (see Table \ref{tab:classification}). Overall, this demonstrates that the introduced meta-learning framework is not only theoretically sound, but also yields state-of-the-art empirical performance in practice.

\paragraph{PACOH improves the predictive uncertainty.}
\looseness -1 We hypothesize that by acquiring the prior in a principled data-driven manner (e.g., with \emph{PACOH}), we can improve the quality of the GP's and BNN's uncertainty estimates. To investigate the effect of meta-learned priors on the uncertainty estimates of the base learners, we compute the probabilistic predictors' calibration errors, reported in Table~\ref{tab:cal_err_reg} and~\ref{tab:classification}. In regression, the {\em calibration error} measures the discrepancy between the predicted confidence regions and the actual frequencies of test data points in the respective areas \citep{Kuleshov2018}. In the case of classification, it measures how well the probability of the predicted class reflects the corresponding misclassification rate \citep{guo2017calibration}.
Note that, since \emph{MAML} only produces point predictions, the concept of calibration does not apply to it. We observe that meta-learning priors with \emph{PACOH-NN} consistently improves the Vanilla BNN's uncertainty estimates. Similarly, \emph{PACOH-GP} yields a lower calibration error than the Vanilla GP in the majority of the environments. For meta-learning environments where the task similarity is high, like \emph{SwissFEL} and \emph{Berkeley-Sensor}, the improvement is substantial. 
\input{tables/classification_table.tex}
%

\paragraph{PACOH combats meta-overfitting.}
\begin{figure}
\centering
\includegraphics[width=0.85\linewidth]{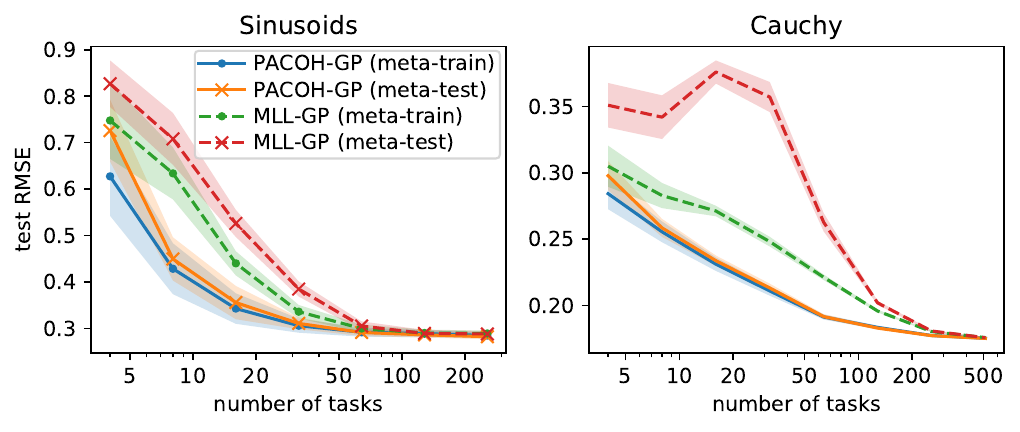}\vspace{-2pt}
\caption{ \looseness -1 Test RMSE on meta-training and meta-testing tasks as a function of the number of meta-training tasks for PACOH-GP-MAP  and MLL-GP. The performance gap between the train and test tasks demonstrates overfitting in the MLL method, while PACOH performs consistently better and barely overfits. \vspace{-14pt}} 
\label{fig:meta_overfitting}
\end{figure}
\looseness -1 
As \citet{qin2018rethink} and \citet{yin2020meta} point out, many  popular meta-learners \citep[e.g.,][]{finn2017model, garnelo2018neural} require a large number of meta-training tasks to generalize well. When presented with only a limited number of tasks, such algorithms suffer from severe meta-overfitting, adversely impacting their performance on unseen tasks from $\calT$. This can even lead to {\em negative transfer}, such that meta-learning actually hurts the performance when compared to standard learning. In our experiments, we also observe such failure cases: For instance, in the classification environment (Table \ref{tab:classification}), \emph{MAML} fails to improve upon the Vanilla BNN. Similarly, in the regression environments (Table \ref{tab:cal_err_reg}) we find that \emph{NPs}, \emph{BMAML}, and \emph{MLL-GP} often yield worse-calibrated predictive distributions than the Vanilla BNN and GP, respectively.
In contrast, thanks to its theoretically principled construction, \emph{PACOH-NN} is able to achieve positive transfer even when the tasks are diverse and small in number. In particular, the hyper-prior acts as a meta-level regularizer by penalizing complex priors that are unlikely to convey useful inductive bias for unseen learning tasks.

\looseness -1 To investigate the importance of meta-level regularization through the hyper-prior in more detail, we compare the performance of our proposed method \emph{PACOH-GP} to \emph{MLL-GP} \citep{fortuin2019deep}, which also maximizes the sum of GP marginal log-likelihoods across tasks but has no hyper-prior nor meta-level regularization. 
Figure \ref{fig:meta_overfitting} shows that MLL-GP performs significantly better on the meta-training tasks than on the meta-test tasks in both of our synthetic regression environments. This gap between meta-train performance and meta-test performance signifies overfitting on the meta-level.
In contrast, our method hardly exhibits this gap and consistently outperforms MLL-GP. As expected, this effect is particularly pronounced when the number of meta-training tasks is small (i.e., less than 100) and vanishes as $n$ becomes large.
Once more, this demonstrates the importance of meta-level regularization, and shows that our proposed framework effectively addresses the problem of meta-overfitting.

\paragraph{PACOH is scalable.} 
\begin{figure}
\centering
\includegraphics[width=.75\textwidth]{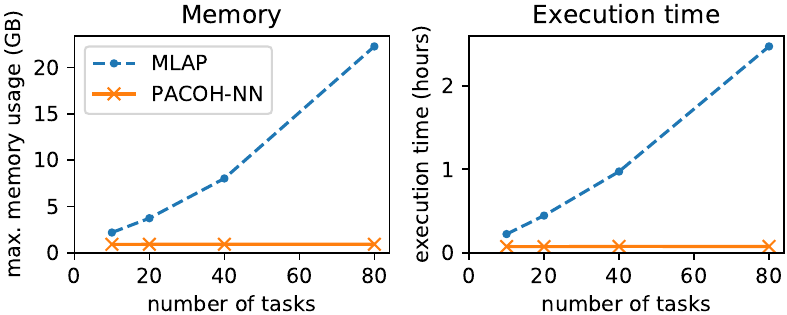}
\caption{Comparison of \emph{PACOH-NN} and \emph{MLAP} in memory footprint and compute time, as the number of meta-training task grows. \emph{PACOH-NN} scales much better in the number of tasks than \emph{MLAP}.}
\label{fig:complexity_analysis}
\end{figure}
A key feature of the proposed approach is that it drastically simplifies PAC-Bayesian meta-learning by converting the bi-level minimization of a\added{PAC-Bayesian}bound (cf., Eq. \ref{eq:meta-learning-bound}) into a variational approximation of the closed-form \emph{PACOH}.
Hence, unlike \emph{MLAP} \citep{amit2017meta}, \emph{PACOH-NN} does not need to maintain posteriors $Q_i$ for the meta-training tasks and can use mini-batching on the task level. 
Thus, it is {\em computationally much faster and more scalable} than previous PAC-Bayesian meta-learners. This is reflected in its computation and memory complexity, discussed in Section \ref{sec:method2}. Figure \ref{fig:complexity_analysis} showcases this computational advantage during meta-training with \emph{PACOH-NN} and \emph{MLAP} on the \emph{Sinusoids} environment with varying number of tasks, reporting the maximum memory requirements, as well as the training time. While \emph{MLAP's} memory consumption and compute time grow linearly and become prohibitively large even for less than 100 tasks, \emph{PACOH-NN} maintains a constant memory and compute load as the number of tasks grow.

\vspacesubcaption
\subsection{Meta-Learning for Sequential Decision Making} \label{sec:bandit}
\vspacesubcaption
 

\added{In Section \ref{sec:exp_reg_class}, we have considered supervised learning problems. Now, we go one step further and study our PACOH approach in the context of sequential decision-making where the predictions and uncertainty estimates are used to interactively collect the training data.}

\added{We consider a Bayesian Optimization (BO) problem where the goal is to optimize a (black-box) target function with as few function queries as possible \citep[see, e.g.,][]{shahriari2015taking}. To do so efficiently, BO approaches typically use a probabilistic model of the objective function together with an uncertainty-aware acquisition function to decide where to sequentially query the objective function.} 

\added{In particular, we study a problem from molecular biology:}The goal is to discover peptides\added{that bind} to major histocompatibility complex class-I molecules (MHC-I). MHC-I molecules present fragments of proteins from within a cell to T-cells, allowing the immune system to distinguish between healthy and infected cells.
Following the setup of \citet{krause2011contextual}, the considered BO problem corresponds to searching for maximally binding peptides, a vital step in the design of peptide-based vaccines.\added{In each iteration, the experimenter (i.e., the BO algorithm) chooses to test one peptide among the pool of more than 800 candidates and receives its binding affinity as evaluation of the objective function.}

\added{We have a number of such BO tasks that}differ in their targeted MHC-I allele, corresponding to different genetic variants of the MHC-I protein. We use data from \citet{widmer2010inferring}, which contains the standardized binding affinities ($\text{IC}_{50}$ values) of different peptide candidates (encoded as 45-dimensional feature vectors) to the MHC-I alleles.\added{Since the feature space is high-dimensional, this BO problem is very challenging. We aim to investigate whether PACOH allows us to transfer useful knowledge across alleles and, thus, accelerate the optimization for new alleles.}

\begin{figure}
\centering
\vspace{-5pt}
\includegraphics[width=0.75\textwidth]{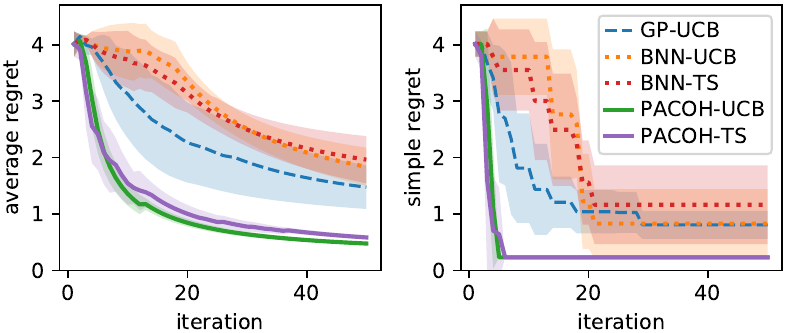}
\vspace{-3pt}
\caption{\looseness -1 MHC-I peptide design task: Regret for different priors (GP, standard BNN prior and meta-learned \emph{PACOH-NN} prior) and bandit algorithms (UCB and TS). A meta-learned \emph{PACOH-NN} prior substantially improves the regret, compared to a standard BNN/GP prior. \vspace{-4pt}}
 \label{fig:bandit_mhc}
\end{figure}

\looseness -1\added{In our experiment, we use the datasets for 5 alleles (tasks) to meta-learn a BNN prior with \emph{PACOH-NN} and leave the most genetically dissimilar allele (A-6901) as the test BO task. In particular, we use a BNN with meta-learned prior as a probabilistic model of the objective function. To pick the next protein candidates to evaluate based on the BNN's predictions, we employ the UCB \citep{auer2002using} and Thompson-Sampling (TS) \citep{thompson1933likelihood} BO algorithms. We refer to the respective approaches as {\em PACOH-UCB} and {\em PACOH-TS}. As baselines, we compare} against a BNN with zero-centered Gaussian prior (\emph{BNN-UCB/TS}) and a Gaussian process as dynamics model (\emph{GP-UCB}) \citep{srinivas2009gaussian}.\added{For further details we refer to Appendix \ref{appx:vaccine_exp}.}

\looseness -1 Figure~\ref{fig:bandit_mhc} reports the respective average regret and simple regret over 50 iterations.
Unlike the bandit algorithms with standard BNN/GP prior, \emph{PACOH-UCB/TS} reaches near-optimal regret within less than 10 iterations, and after 50 iterations still maintains a significant performance advantage. This highlights the importance of \emph{transfer (learning)} for solving real-world problems and demonstrates the effectiveness of \emph{PACOH-NN} to this end. While the majority of meta-learning methods rely on a large number of meta-training tasks \citep{qin2018rethink}, \emph{PACOH-NN} allows us to achieve promising positive transfer, even in complex real-world scenarios with only a\addedII{few of tasks (in this case 5).}

%% file: tables/avg_rmse_table.tex
\begin{table*}[t]
\centering
\resizebox{\textwidth}{!}{\begin{tabular}{l||c|c|c|c|c}
& \multicolumn{1}{c|}{Cauchy} & \multicolumn{1}{c|}{SwissFel} & \multicolumn{1}{c|}{Physionet-GCS} & \multicolumn{1}{c|}{Physionet-HCT} & \multicolumn{1}{c}{Berkeley-Sensor} \\
\hline
Vanilla GP & $0.275 \pm 0.000$ & $0.876 \pm 0.000$ & $2.240 \pm 0.000$ & $2.768 \pm 0.000 $ & $0.276 \pm 0.000 $ \\
Vanilla BNN \citep{Liu2016} & $0.327 \pm 0.008$ & $0.529 \pm 0.022$ & $2.664 \pm 0.274$ & $3.938 \pm 0.869$ & $0.109 \pm 0.004$ \\
\hline
MLL-GP \citep{fortuin2019deep} & $0.216 \pm 0.003$ & $0.974 \pm 0.093$ & $1.654 \pm 0.094$ & $2.634 \pm 0.144$ & $0.058 \pm 0.002$  \\
MLAP \citep{amit2017meta}& $0.219 \pm 0.004$ & $0.486 \pm 0.026$ & $2.009 \pm 0.248$ & $2.470 \pm 0.039$ & $0.050 \pm 0.005$ \\
MAML \citep{finn2017model} & $0.219 \pm 0.004$ & $0.730 \pm 0.057$ & $1.895 \pm 0.141$ & $2.413 \pm 0.113$ & $\mathbf{0.045 \pm 0.003}$ \\ 
MR-MAML (W) \citep{yin2020meta} &   $0.227 \pm 0.002$  &   $0.483 \pm 0.021$ &   $1.643 \pm 0.174$ &   $\mathbf{2.306 \pm  0.162}$ &  $0.059 \pm 0.004$ \\
MR-MAML (A) \citep{yin2020meta} &   $0.228 \pm 0.003$  &   $0.555 \pm 0.045$ &   $1.556 \pm 0.179$ &   $\mathbf{2.275 \pm 0.155}$ &  $0.066 \pm 0.006$ \\ 
BMAML \citep{yoon2018bayesian}  & $0.225 \pm 0.004$ & $0.577 \pm 0.044$ & $1.894 \pm 0.062$ & $2.500 \pm 0.002$ & $0.073 \pm 0.014$ \\
NP \citep{garnelo2018neural}  & $0.224 \pm 0.008$ & $0.471 \pm 0.053$ & $2.056 \pm 0.209$ & $2.594 \pm 0.107$ & $0.079 \pm 0.007$ \\  \hline
PACOH-GP-MAP (ours) & $0.213 \pm 0.003$ & $0.486 \pm 0.055$ & $1.492 \pm 0.091$ & $2.574 \pm 0.058$ & $0.056 \pm 0.001$ \\
PACOH-GP-SVGD (ours) & $0.209  \pm 0.008$ & $0.376 \pm 0.024$ & $1.498 \pm 0.081$ & $\mathbf{2.361 \pm 0.047}$ & $0.065 \pm 0.005$ \\ 
PACOH-GP-VI (ours) & $0.218 \pm  0.003$ & $0.387 \pm 0.019$ & $\mathbf{1.472 \pm 0.012}$ & $2.695 \pm 0.040 $ & $0.095 \pm 0.010$ \\\hline
PACOH-NN-MAP (ours) & $0.202 \pm 0.003$ & $0.375 \pm 0.004$ & $1.564 \pm 0.200$ & $2.480 \pm 0.042$ & $0.047 \pm 0.001$ \\
PACOH-NN-SVGD (ours) & $\mathbf{0.195 \pm 0.001}$ & $\mathbf{0.372 \pm 0.002}$ & $1.561 \pm 0.061$ & $\mathbf{2.405 \pm 0.017}$ & $\mathbf{0.043 \pm 0.001}$ \\ 

\end{tabular}}
\vspace{-3pt}
\caption{Comparison of standard and meta-learning algorithms in terms of test RMSE in 5 meta-learning environments for regression. We report the mean and standard deviation across 5 random seeds. \emph{PACOH} achieves the best performance across the environments.  \vspace{-4pt}
} \label{tab:reg_rmse}
\end{table*}

%% file: tables/cal_err_regression_table.tex
\begin{table*}[t]
\centering
\resizebox{\textwidth}{!}{
\begin{tabular}{l||c|c|c|c|c}
& \multicolumn{1}{c|}{Cauchy} & \multicolumn{1}{c|}{SwissFel} & \multicolumn{1}{c|}{Physionet-GCS} & \multicolumn{1}{c|}{Physionet-HCT} & \multicolumn{1}{c}{Berkeley-Sensor} \\
\hline
Vanilla GP & $0.087 \pm 0.000$ & $0.135 \pm 0.000$ & $0.268 \pm 0.000$ & $\mathbf{0.277 \pm 0.000}$ & $0.119 \pm 0.000 $ \\
Vanilla BNN  \citep{Liu2016} & $0.055 \pm 0.006$ & $0.085 \pm 0.008$ & $0.277 \pm 0.013$ & $0.307 \pm 0.009$ & $0.179 \pm 0.002$ \\

\hline
MLL-GP \citep{fortuin2019deep} & $0.059 \pm 0.003$ & $0.096 \pm 0.009$ & $ 0.277 \pm 0.009$ & $0.305 \pm 0.014$ & $0.153 \pm 0.007$  \\
MLAP \citep{amit2017meta} & $0.086 \pm 0.015$ & $0.090 \pm 0.021$ & $0.343 \pm 0.017$ & $0.344 \pm 0.016$ & $0.108 \pm 0.024$ \\
BMAML \citep{yoon2018bayesian}  & $0.061 \pm 0.007$ & $0.115 \pm 0.036 $ & $0.279 \pm  0.010$ & $0.423 \pm 0.106$ & $0.161 \pm 0.013$ \\ 
NP \citep{garnelo2018neural} & $0.057 \pm 0.009$ & $0.131 \pm 0.056$ & $0.299 \pm 0.012$ & $0.319 \pm 0.004$ & $0.210 \pm 0.007$ \\ \hline
PACOH-GP-MAP (ours) & $ 0.058 \pm 0.006$ & $0.055 \pm 0.005$ & $0.267 \pm 0.010$ & $0.298 \pm 0.003$ & $0.155 \pm 0.004$ \\
PACOH-GP-SVGD (ours) & $0.056 \pm 0.004$ & $0.038 \pm 0.006$ & $\mathbf{0.262 \pm  0.004}$ & $0.296 \pm 0.003$ & $0.098 \pm 0.005$ \\ 
PACOH-GP-VI (ours) & $0.057 \pm 0.002$ & $0.046 \pm 0.016$ & $0.265 \pm 0.002$ & $0.311 \pm  0.005$ & $0.089 \pm 0.003$ \\ \hline
PACOH-NN-MAP (ours) & $0.051 \pm 0.002$ & $0.031 \pm 0.003$ & $0.268 \pm 0.015$ & $0.306 \pm 0.003$ & $\mathbf{0.063 \pm 0.016}$ \\
PACOH-NN-SVGD (ours) & $\mathbf{0.046 \pm 0.001}$ & $\mathbf{0.027 \pm 0.003}$ & $0.267 \pm 0.005$ & $0.302 \pm 0.003$ & $\mathbf{0.067 \pm 0.005}$ \\

\end{tabular}
}
\vspace{-3pt}
\caption{\looseness -1 Comparison of standard and meta-learning methods in terms of the test calibration error in 5 regression environments. We report the mean and standard deviation across 5 random seeds. \emph{PACOH} yields the best uncertainty calibration in the majority of environments. \label{tab:cal_err_reg} \vspace{-10pt}} 
\end{table*}

%% file: tables/classification_table.tex
\begin{table*}[t]
\centering 
\resizebox{0.93\textwidth}{!}{
\begin{tabular}{l||c|c||c|c}
& \multicolumn{2}{c||}{Accuracy} & \multicolumn{2}{c}{Calibration error} \\ \hline
& Omniglot 2-shot & Omniglot 5-shot & Omniglot 2-shot & Omniglot 5-shot \\
\hline
BNN  \citep{Liu2016}& $0.6709 \pm 0.006 $  & $0.795 \pm 0.006$  & $0.173 \pm 0.009$ & $0.135 \pm 0.009$  \\ \hline
MLAP-M \citep{amit2017meta} & $0.635 \pm 0.015$ & $0.804 \pm 0.0168$ & $0.108 \pm 0.008$ & $0.119 \pm 0.019$  \\
MLAP-S \citep{amit2017meta} & $0.615 \pm 0.037$ & $0.700 \pm 0.0135 $ & $0.129 \pm 0.018$ & $0.108 \pm 0.010$ \\ \hline

FO-MAML \citep{nichol2018firstorder} & $0.429 \pm 0.047$ & $0.590 \pm 0.010$ & N/A & N/A  \\
MAML \citep{finn2017model}  & $0.571 \pm 0.018$ & $0.693 \pm 0.013$ & N/A & N/A  \\
\hline
BMAML \citep{yoon2018bayesian}  & $0.651 \pm 0.008$  & $0.764 \pm 0.025$  & $0.132 \pm 0.007$ & $0.191 \pm 0.018$  \\ \hline
PACOH-NN-SVGD (ours) & $0.733 \pm 0.009$ & $\mathbf{0.885\pm 0.090}$ &  $\mathbf{0.094 \pm 0.004}$ & $0.091 \pm 0.010$ \\
PACOH-NN-MAP (ours) & $\mathbf{0.735 \pm 0.010}$ & $0.866 \pm 0.005$ & $0.099 \pm 0.009$ & $\mathbf{0.075 \pm 0.006}$  \\
\end{tabular}}
\caption{Comparison of meta-learning algorithms in terms of test accuracy and calibration error on the \emph{Omniglot} environment with 2-shot and 5-shot 5-way-classification tasks.
We report the mean and standard deviation across 5 random seeds. \emph{PACOH} achieves the best performance, while yielding the best uncertainty calibration.}
\label{tab:classification}
\end{table*}

%
%

%% file: content/discussion.tex
\vspacecaption
\section{Summary and Critical Discussion}
\vspacecaptionlow
\label{sec:conclusion}

\looseness -1 This paper provides a theoretical analysis of generalization in meta-learning, studying the factors that drive positive transfer and improvement over single-task learning. We present novel PAC-Bayesian generalization bounds on the transfer error and derive the PAC-optimal hyper-posterior (PACOH) in closed form. This transforms PAC-Bayesian meta-learning from a previously\added{difficult bi-level optimization}problem into\added{simple}approximate inference on the PACOH.

\added{
However, this comes at a price. While Theorem \ref{theorem:meta-learning-bound} holds for arbitrary choices of base learners and hyper-posteriors, the bound for Gibbs learners (Corollary \ref{cor:bayesian_learner_PAC_bound}) and for the PACOH (Corollary \ref{cor:meta_bound_pacoh}) only hold when we employ the exact Gibbs posterior and PAC-optimal hyper-posterior. In most practical settings, however, we can only approximate these distributions. Hence, the respective bounds formally do not hold for the algorithms proposed in Section \ref{sec:algorithm}. Additionally, for complex and over-parametrized hypothesis spaces such as those of neural networks, PAC-Bayesian bounds are well-known to be loose or even vacuous. This is most likely also the case for the bounds in this paper.
}

\looseness -1 \added{Nonetheless, the presented PACOH framework gives rise to a range of meta-learning algorithms that are practical and scalable. As demonstrated in the experiments, the proposed algorithms achieve}state-of-the-art performance in terms of predictive accuracy and the quality of uncertainty estimates across both regression and classification tasks.
Thanks to their foundation in learning theory,\added{our meta-learning}methods are equipped with principled meta-level regularization which allows them to achieve positive transfer with as little as five\added{learning}tasks.
As\added{shown}by our experiments on vaccine design, our meta-learned priors can be effectively employed in realistic sequential decision-making problems. 
Overall, we believe that our approach provides an important step towards understanding generalization in meta-learning and, in practice, reliably learning useful inductive biases when meta-training data is scarce.

\clearpage

%% file: content/acknowledgements.tex
\section*{Acknowledgements}
\looseness -1 This research was supported by the European Research Council (ERC) under the EU's Horizon 2020 research and innovation program grant agreement no.\ 815943.
Jonas Rothfuss was supported by an Apple Scholars in AI/ML fellowship.
Vincent Fortuin was supported by a PhD Fellowship from the Swiss Data Science Center, a Postdoc.Mobility Fellowship from the Swiss National Science Foundation, a Research Fellowship from St John's College Cambridge, and a Branco Weiss Fellowship.
We thank Parnian Kassraie for her valuable feedback.

%% file: content/supplement_theory.tex
\newpage

\beginsupplement

\appendix

\section*{Appendix}

\section{Proofs and Derivations} \label{appendix:proofs}

\added{Various proofs in this section follow along with \citet{rothfuss2020pacoh}, i.e., the short conference version of this paper. In particular, the proofs of Corollary \ref{cor:bayesian_learner_PAC_bound} and Proposition \ref{proposition:pacoh_optimal_hyper_posterior} are almost identical to \citet{rothfuss2020pacoh}. The proofs of Theorem \ref{theorem:meta-learning-bound}, Corollary \ref{corr:bounded}, and Corollary \ref{corr:sub_gamma} have seen significant revisions to make them easier to follow and more rigorous. The proofs of Corollary \ref{cor:meta_bound_pacoh}, Theorem \ref{theorem:single_task_bound_hyper_prior}, and Proposition \ref{prop:single_vs_meta} are new additions.}

\subsection{Proof of Theorem~\ref{theorem:meta-learning-bound}}
\label{appendix:proof_theorem_meta_pac_bound}
\added{
A key tool in the PAC-Bayesian framework which we also use in our proofs is the change of measure inequality:
\begin{lemma} \label{lemma:concentration_sum_of_rvs_2}  \textbf{(\citet{csiszar1975divergence, donsker1975})}
Let $f$ be a random variable taking values in a set $A$ and $g: A \rightarrow \R$ a function. For distributions $\pi, \rho$ $\in \mathcal{M}(A)$ and any $\lambda > 0$, if $\E_{f\sim\rho} \left[ g(f) \right]$ exists and is finite, we have that
\begin{align} \label{eq:change_of_measure_simple}
\E_{f\sim\rho} \left[ g(f) \right] \leq \frac{1}{\lambda} \left( D_{KL}(\rho||\pi) + \log \E_{f\sim\pi} \left[ e^{ \lambda  g(f) }  \right] \right) .
\end{align}
\end{lemma}}
\added{
Lemma \ref{lemma:concentration_sum_of_rvs_2} follows from the positivity of the KL-divergence and is a special case of the convex duality of the KL divergence (see e.g., \citet[][Appendix A]{seeger2002pac} for a proof). Hence, it is fairly general and only requires the integrability of $g(f)$ under $\rho$. Importantly, it also holds for random functions $g$, a property we will use in our proof.}

To prove Theorem \ref{theorem:meta-learning-bound}, we need to bound the difference between \emph{transfer error} $\calL(\calQ, \calT)$ and the \emph{empirical multi-task error} $\hat{\calL}(\calQ, S_1, ..., S_n)$. To this end, we\added{make use of}an intermediate quantity, the \emph{expected multi-task error}:
\begin{equation}
\tilde{\calL}(\calQ, \tau_1, ..., \tau_n) = \E_{P \sim \calQ} \left[ \frac{1}{n} \sum_{i=1}^n  \calL(Q(S_i, P), \calD_i) \right] 
\end{equation}
In the following, we invoke Lemma \ref{lemma:concentration_sum_of_rvs_2} twice. First, in step 1, we bound the difference between $\tilde{\calL}(\calQ, \tau_1, ..., \tau_n)$ and $\hat{\calL}(\calQ, S_1, ..., S_n)$, then, in step 2, the difference between $\calL(\calQ, \calT)$ and $\tilde{\calL}(\calQ, \tau_1, ..., \tau_n)$. Finally, in step 3, we combine both results and bound the cumulant-generating functions that arise from applying (\ref{eq:change_of_measure_simple}).

\paragraph{Step 1 (Task-specific generalization)}
\added{First, we bound the expected generalization error across the observed tasks $\tau_i=(\calD_i, S_i)$, $i=1,...,n$, when using a hyper-posterior $\calQ \in \calM(\calM(\calH))$ and a base learner $Q: \calZ^{m} \times \calM(\calH) \rightarrow \calM(\calH)$. Remember from Section \ref{sec:meta_learning} that the base learner takes as an input a prior distribution $P \in \calM(\calH)$ as well as a dataset $S_i \sim \calD^{m}_i$ of size $m$ and outputs a posterior distribution $Q_i=Q(S_i, P)$ over hypotheses $h \in \calH$. For brevity, we also write $Q_i$ short for $Q(S_i, P)$ in the following.}

\added{
\textit{Instantiations for Lemma \ref{lemma:concentration_sum_of_rvs_2}:} To use Lemma \ref{lemma:concentration_sum_of_rvs_2}, we set $A = \calM(\calH) \times \calH^n$, and, correspondingly, $f=(P, h_1, ..., h_n) \in A$. We can interpret this as a joint two-level hypothesis of a prior $P$ (i.e., the hypothesis of the meta-learner) and $n$ base hypotheses $h_i$, one for each task. Denoting $P^n$ as product distribution of $n$ prior distributions $P$, we set $\pi = (\calP, P^n)$ which is a distribution over the joint two-level hypotheses corresponding to hierarchical sampling via $P \sim \calP$ and $h_i \sim P ~ \forall i = 1, ..., n$. Furthermore, we denote $Q^n = Q_1 \cdots Q_n$ as product distribution of the task-specific posteriors and set $\rho = (\calQ, Q^n)$ as a distribution over the joint two-level hypotheses corresponding to hierarchical sampling via $P \sim \calQ$ and $h_i \sim Q_i=Q(S_i, P) ~ \forall i = 1, ..., n$.
Finally, we set $g(f) = g(P, h_1, ..., h_n) = \frac{1}{n} \sum_{i=1}^n \calL(h_i, \calD_i) - \hat{\calL}(h_i, S_i)$ and $\lambda = \gamma$.}

\added{
\textit{Integrability conditions for Lemma \ref{lemma:concentration_sum_of_rvs_2}:} Finally, we check whether the integrability conditions for the application of Lemma \ref{lemma:concentration_sum_of_rvs_2} are satisfied.
By assumption, the expectation 
\begin{equation}
    \calL(\calQ, \calT) = \E_{\calD \sim \calT} \E_{S \sim \calD^m} \E_{ P \sim \calQ} \left[\calL(Q(S, P), \calD) \right]
\end{equation}
is finite. Thus, the probability that we sample a $(\calD, S) \sim \calT_h$ so that either $\E_{ P \sim \calQ} \left[  \calL(Q(S, P), \calD) \right]$ or $\E_{ P \sim \calQ} \left[  \hat{\calL}(Q(S, P), S) \right]$ are infinite or undefined must be zero. Hence, we have with probability 1 that every summand in 
\begin{equation}
   \E_{f \sim \rho} [g(f)] =  \frac{1}{n} \sum_{i=1}^n \E_{P \sim \calQ} \left[  \calL(Q(S_i, P), \calD_i)\right]  - \frac{1}{n} \sum_{i=1}^n \E_{P \sim \calQ} \left[ \hat{\calL}(Q(S_i, P), S_i) \right]  
\end{equation}
is finite, and, thus that the overall sum, i.e., $\E_{f \sim \rho} [g(f)]$ is finite.}

\added{\textit{Application of Lemma \ref{lemma:concentration_sum_of_rvs_2}:}
Finally, we can invoke Lemma \ref{lemma:concentration_sum_of_rvs_2} to obtain that, with probability 1, for any $\gamma > 0$,
\begin{align} \label{eq:sdkjfbskjdf}
\begin{split}
    \E_{(P, h_1, ..., h_n) \sim (\calQ, Q^n)} \left[ \frac{1}{n} \sum_{i=1}^n \calL(h_i, \calD_i) - \hat{\calL}(h_i, S_i) \right] \leq \frac{1}{\gamma} \bigg( D_{KL}((\calQ,  Q^n) || (\calP, P^n)) \\
    + \log \E_{(P, h_1, ..., h_n) \sim (\calP, P^n)} \left[ e^{ \gamma \left( \frac{1}{n} \sum_{i=1}^n \calL(h_i, \calD_i) - \hat{\calL}(h_i, S_i) \right) }  \right] \bigg) \;.
\end{split}
\end{align}
By using the definitions of $\calL(Q_i, \calD_i)$ and $\hat{\calL}(Q_i, S_i)$, and that $h_i \sim P$ are i.i.d., we can write (\ref{eq:sdkjfbskjdf}) as
\begin{align}
\begin{split} \label{eq:kjsglkdbsgklsbe}
    \E_{P \sim \calQ} \left[ \frac{1}{n} \sum_{i=1}^n \calL(Q_i, \calD_i) - \hat{\calL}(Q_i, S_i) \right] \leq \frac{1}{\gamma} D_{KL}((\calQ,  Q^n) || (\calP, P^n)) \\
    + \frac{1}{\gamma} \log \E_{P \sim \calP} \E_{h \sim P} \left[ e^{  \frac{\gamma}{n} \sum_{i=1}^n \calL(h, \calD_i) - \hat{\calL}(h, S_i)}  \right] \bigg)
\end{split}
\end{align}
}

\added{Next we follow the arguments of \citet{pentina2014pac}}and use the above definitions to rewrite the KL-divergence term as follows:
\begin{align}
    D_{KL}\left[ (\calQ, Q^n) || (\calP, P^n) \right]  &= \E_{P \sim \calQ} \left[ \E_{h_1 \sim Q_1} ... \E_{h_n \sim Q_n}\left[ \log \frac{\calQ(P)  \prod_{i=1}^n Q_i(h_i)}{\calP(P)  \prod_{i=1}^n  P(h_i)}\right] \right] \\
    &= \E_{P \sim \calQ} \left[ \log \frac{\calQ(P)}{\calP(P)}\right] + \sum_{i=1}^n \E_{P \sim \calQ} \left[ \E_{h \sim Q_i} \left[ \log \frac{Q_i(h)}{P(h)}\right] \right] \\
    &= D_{KL}(\calQ||\calP) + \sum_{i=1}^n \E_{P \sim \calQ} \left[ D_{KL}(Q_i || P)\right] \label{eq:two_level_kl}
\end{align}
By inserting (\ref{eq:two_level_kl}) into (\ref{eq:kjsglkdbsgklsbe}) and using the definitions of $\tilde{\calL}(\calQ, \tau_1, ..., \tau_n)$ and $\hat{\calL} (\calQ, S_1, ..., S_n)$ we obtain a bound on the expected multi-task error:
\begin{align} \label{eq:step1_bound}
\begin{split}
    \tilde{\calL}(\calQ, \tau_1, ..., \tau_n)  \leq \hat{\calL} (\calQ, S_1, ..., S_n) + \frac{1}{\gamma} D_{KL}(\calQ||\calP) + \frac{1}{\gamma} \sum_{i=1}^n \E_{P \sim \calQ} \left[ D_{KL}(Q_i || P)\right] \\
    + \frac{1}{\gamma} \underbrace{\log \E_{P \sim \calP} \E_{h \sim P} \left[ e^{  \frac{\gamma}{n} \sum_{i=1}^n \calL(h, \calD_i) - \hat{\calL}(h, S_i)}  \right] }_{\coloneqq \Upsilon^{\rom{1}}(\gamma)} 
\end{split}
\end{align}

\looseness -1 \paragraph{Step 2 (Task environment generalization)}
\added{
Now, we apply Lemma~\ref{lemma:concentration_sum_of_rvs_2} on the meta-level to bound the difference between the transfer error $\calL(\calQ, \calT)$ and the expected multi-task error $\tilde{\calL}(\calQ, \tau_1, ..., \tau_n)$. For that, we use the following instantiations for Lemma~\ref{lemma:concentration_sum_of_rvs_2}: $A = \calM(\calH)$, $f = P$, $\rho = \calQ$, $\pi = \calP$ and $g(f) = g(P) = \frac{1}{n} \sum_{i=1}^n \E_{\calD \sim \calT} \E_{S \sim \calD^m} \left[ \calL(Q(S,P), \calD) \right] - \calL(Q(S_i,P), \calD_i)$. Again, we have to check the integrability conditions: By assumption, the expectation $\calL(\calQ, \calT)$ is finite and, thus, $\E_{f \sim \rho} [g(f)]$ must be finite with probability 1 over sampled tasks $\tau \sim \calT_h$ (proof analogous to Step 1).}
Hence, we obtain that, with probability 1, for all $\lambda > 0$, 
\begin{align} \label{eq:step_2_bound}
\begin{split}
\calL(\calQ, \calT) \leq & ~ \tilde{\calL}(\calQ, \tau_1, ..., \tau_n) + \frac{1}{\lambda} D_{KL}(\calQ || \calP)  \\ & ~ + \frac{1}{\lambda} \underbrace{\log \E_{P \sim \calP} \left[ e^{ \frac{\lambda}{n} \sum_{i=1}^n \E_{\calD \sim \calT} \E_{S \sim \calD^m} \left[ \calL(Q(S,P), \calD) \right] - \calL(Q(S_i,P), \calD_i) }  \right] }_{\coloneqq \Upsilon^\rom{2}(\lambda)}.
\end{split}
\end{align}

Combining (\ref{eq:step1_bound}) with (\ref{eq:step_2_bound}), we obtain
\begin{align}
\begin{split}
\calL(\calQ, \calT) \leq &~ \hat{\calL}(\calQ, S_1, ..., S_n)  + \left(\frac{1}{\lambda} + \frac{1}{\gamma} \right) D_{KL}(\calQ||\calP) \\
& +  \frac{1}{\gamma} \sum_{i=1}^n  \E_{P \sim \calQ} \left[ D_{KL}(Q_i || P)\right] +  \frac{1}{\gamma} \Upsilon^{\rom{1}}(\gamma) +  \frac{1}{\lambda} \Upsilon^\rom{2}(\lambda)
\end{split}
\end{align}

\paragraph{Step 3 (Bounding the cumulant-generating functions)}
Finally, we aim to bound the random quantity $\frac{1}{\gamma} \Upsilon^{\rom{1}}(\gamma) +  \frac{1}{\lambda} \Upsilon^\rom{2}(\lambda)$. Note that the randomness of $\Upsilon^{\rom{1}}(\gamma)$ is governed by the random data points $z_{i,j}$ sampled i.i.d.\ from the respective data distribution $\calD_i$, while $\Upsilon^\rom{2}(\lambda)$ is governed by random tasks sampled from the environment $\calT$. 
    
    First, we factor out $\sqrt{n}$ from $\gamma$ and $\lambda$, obtaining
    \begin{equation} \label{eq:factor_out_sqrt_n}
    \frac{1}{\gamma} \Upsilon^{\rom{1}}(\gamma) +  \frac{1}{\lambda} \Upsilon^\rom{2}(\lambda) = \frac{1}{\sqrt{n}} \left( \frac{\sqrt{n}}{\gamma} \Upsilon^{\rom{1}}(\gamma) +  \frac{\sqrt{n}}{\lambda} \Upsilon^\rom{2}(\lambda) \right)
    \end{equation}
    Next, we proceed by bounding the inner part on the RHS, i.e., $\frac{\sqrt{n}}{\gamma} \Upsilon^{\rom{1}}(\gamma) +  \frac{\sqrt{n}}{\lambda} \Upsilon^\rom{2}(\lambda)$. Using Markov's inequality, we have 
    \begin{equation}
    e^{\frac{\sqrt{n}}{\gamma} \Upsilon^{\rom{1}}(\gamma) +  \frac{\sqrt{n}}{\lambda} \Upsilon^\rom{2}(\lambda)}  \leq  \frac{\E \left[ e^{ \frac{\sqrt{n}}{\gamma} \Upsilon^{\rom{1}}(\gamma) +  \frac{\sqrt{n}}{\lambda} \Upsilon^\rom{2}(\lambda)} \right]}{\delta}
    \end{equation}
    and thus
    \begin{equation} \label{eq:high_prob_bound_generic}
     \frac{\sqrt{n}}{\gamma} \Upsilon^{\rom{1}}(\gamma) +  \frac{\sqrt{n}}{\lambda} \Upsilon^\rom{2}(\lambda) \leq \log \E \left[ e^{ \frac{\sqrt{n}}{\gamma} \Upsilon^{\rom{1}}(\gamma) +  \frac{\sqrt{n}}{\lambda} \Upsilon^\rom{2}(\lambda)} \right] - \log \delta
\end{equation}
with probability at least $1-\delta$. Next, we bound the expectation, i.e., 

\begin{align} \label{eq:dskjfkjsdfb}
\E \left[ e^{ \frac{\sqrt{n}}{\gamma} \Upsilon^{\rom{1}}(\gamma) +  \frac{\sqrt{n}}{\lambda} \Upsilon^\rom{2}(\lambda)} \right] = ~ \E_{\calT} \left[ e^{\frac{\sqrt{n}}{\lambda} \Upsilon^\rom{2}(\lambda)}  ~ \E_{\calD_1} ... \E_{\calD_1} \left[  \left(e^{\Upsilon^{\rom{1}}(\gamma)}\right)^{\frac{\sqrt{n}}{\gamma} } \right] \right] ~. 
\end{align}
\added{If $\gamma \geq \sqrt{n}$ and $x \mapsto x^{\sqrt{n}/\gamma}$ is a concave function, and we can obtain an upper bound on (\ref{eq:dskjfkjsdfb}) by using Jensen's inequality to move the exponent in (\ref{eq:dskjfkjsdfb}) outside the inner expectation. Further, we denote $V_i^\rom{2} := \E_{\calT} \left[ \calL (Q_i, \calD)  \right] - \calL(Q_i, \calD_i )$ as i.i.d.\ realizations of the random variable under the task distribution.  Similarly, for each $i=1, ....,n$ and $j=1,....,m$, we denote the independent realizations of $\calL(h, \calD_i) - l(h, z_{ij})$ as $V_{ij}^\rom{1}$. Hence, we can write 
\begin{align} 
    (\ref{eq:dskjfkjsdfb}) \leq & ~ \E_{\calT} \left[ e^{\frac{\sqrt{n}}{\lambda} \Upsilon^\rom{2}(\lambda)}  ~ \E_{\calD_1} ... \E_{\calD_n} \left[ e^{\Upsilon^{\rom{1}}(\gamma)} \right]^{\frac{\sqrt{n}}{\gamma}} \right] \\
    = & ~ \E_{\calT} \bigg[ \E_{\calP} \left[ e^{\frac{\lambda}{n} \sum_{i=1}^n V_i^{\rom{2}}} \right]^{\sqrt{n}/\lambda}  \cdot  \E_{\calP} \E_{P}  \E_{\calD_1} ... ~\E_{\calD_n} \left[ e^{ \frac{\gamma}{nm} \sum_{i=1}^n \sum_{j=1}^m V_{ij}^{\rom{1}}}\right]^{\sqrt{n}/\gamma}  \bigg] \\
    = & ~ \E_{\calT} \bigg[ \E_{\calP} \left[ \prod_{i=1}^n e^{\frac{\lambda}{n}  V_i^{\rom{2}}} \right]^{\sqrt{n}/\lambda}  \cdot \E_{\calP} \E_{P}  \E_{\calD_1} ... ~\E_{\calD_n} \left[  \prod_{i=1}^n \prod_{j=1}^m  e^{ \frac{\gamma}{nm} V_{ij}^{\rom{1}}}\right]^{\sqrt{n}/\gamma}  \bigg] ~. \label{eq:sdkjnfkjsdnfkdsjf}
\end{align}
Following the arguments of \citet[][Proof of Lemma 2]{pentina2014pac} and \citet[][Proof of Corollary 4]{germain2016pac}, given a fixed $h$ sampled via $h \sim P$, $P\sim\calP$, the $V_{ij}^\rom{1}$ are independent. Similarly, given a $P$ sampled via $P \sim \calP$, the $V^\rom{2}_i$ are independent. Hence, we can write (\ref{eq:sdkjnfkjsdnfkdsjf}) as
\begin{align} \label{eq:lknlnskfjdngskdfgnkdf}
    (\ref{eq:dskjfkjsdfb})  \leq ~ &  \E_{\calT}  \bigg[ \prod_{i=1}^n \E_{\calP} \left[  e^{\frac{\lambda}{n} V_i^\rom{2}} \right]^{\sqrt{n}/\lambda}   \left( \prod_{i=1}^n \prod_{j = 1}^{m} \E_{\calP} \E_{P}  \E_{\calD_i} e^{ \frac{\gamma}{nm} V_{ij}^\rom{1}}\right)^{\sqrt{n}/\gamma}  \bigg] ~.
\end{align}}

\added{Next, we set $\gamma = \beta n$ and use the uniform bound $\bar{\Psi}^{\rom{1}}(\beta) \geq \Psi^{\rom{1}}(\beta) = \frac{m}{\beta} \log \E_{\calP} \E_{P} \E_{\calD} \left[  e^{ \frac{\beta}{m} V^{\rom{1}} } \right] ~~ \forall ~\calD$ in the support of $\calT$. Crucially, if we upper bound $\prod_{i=1}^n \prod_{j = 1}^{m} \E_{\calP} \E_{P}  \E_{\calD_i} e^{ \frac{\gamma}{nm} V_{ij}^\rom{1}}$ by 
$e^{\beta n \bar{\Psi}^{\rom{1}}(\beta)}$, the upper bound no longer depends on the tasks sampled from $\calT$, allowing us to move it outside the expectation. As a result, we obtain
\begin{align} \label{eq:ijlknfjkndgkjslb}
    (\ref{eq:dskjfkjsdfb})  \leq ~ &  \E_{\calT}  \bigg[ \prod_{i=1}^n \E_{\calP} \left[  e^{\frac{\lambda}{n} V_i^\rom{2}} \right]^{\sqrt{n}/\lambda}  e^{\sqrt{n} \bar{\Psi}^{\rom{1}}(\beta) } \bigg] \\
    = ~ &  \prod_{i=1}^n \E_{\calT}   \E_{\calP} \left[  e^{\frac{\lambda}{n} V_i^\rom{2}} \right]^{\sqrt{n}/\lambda} e^{\sqrt{n} \bar{\Psi}^{\rom{1}}(\beta)} ~.
\end{align}
Further, with $\Psi^\rom{2}(\lambda) = \frac{n}{\lambda} \log  \E_{\calP} \E_{\calT} \left[ e^{ \frac{\lambda}{n}  V_i^\rom{2} } \right]$, we have that 
\begin{align} \label{eq:ljkdnglsnldfgnfd}
    (\ref{eq:dskjfkjsdfb})  \leq ~ & e^{\sqrt{n} \Psi^\rom{2}(\lambda)} \cdot e^{\sqrt{n} \bar{\Psi}^{\rom{1}}(\beta)}.
\end{align}
Finally, we insert (\ref{eq:ljkdnglsnldfgnfd}) into (\ref{eq:high_prob_bound_generic}) to obtain that
\begin{align} \label{eq:ksdfjhkdsbf}
\frac{1}{\gamma} \Upsilon^{\rom{1}}(\gamma) +  \frac{1}{\lambda} \Upsilon^\rom{2}(\lambda)
\leq  \bar{\Psi}^{\rom{1}}(\beta) +  \Psi^\rom{2}(\lambda)  - \frac{1}{\sqrt{n}}\log \delta
\end{align}
with probability at least $1-\delta$ which concludes our high-probability bound.}

\subsection{Proof of Corollary \ref{corr:bounded}}

\added{If the loss function $l(h_i, z_{ij})$ is bounded in $[a,b]$, we can apply Hoeffding's lemma to $\Psi^{\rom{1}}(\beta)$ and $\Psi^\rom{2}(\lambda)$. In particular, we have that
\begin{equation}
\Psi^{\rom{1}}(\beta) = \frac{m}{\beta} \log \E_{\calP} \E_{P} \E_{\calD} \left[  e^{ \frac{\beta}{m} V^{\rom{1}} } \right]  \leq \frac{\beta^2 (b-a)^2}{ 8 m^2} = \bar{\Psi}^{\rom{1}}(\gamma) ~~~ \forall ~\calD ~.
\end{equation}
Similarly, we can bound $\Psi^\rom{2}(\lambda)$ as
\begin{equation}
    \Psi^\rom{2}(\lambda) \leq \frac{\lambda^2 (b-a)^2}{8 n^2} = \bar{\Psi}^\rom{2}(\lambda) ~.
\end{equation}
When unserting these bounds on the CGFs into (\ref{eq:ksdfjhkdsbf}), we have that Theorem \ref{theorem:meta-learning-bound} holds with 
\begin{equation}
 C(\beta, \lambda, \delta) = \frac{\lambda (b-a)^2}{8 n} +  \frac{\beta (b-a)^2}{8 m} - \frac{1}{\sqrt{n}}\log \delta ~.
\end{equation}}

\subsection{Proof of Corollary \ref{corr:sub_gamma}} 
\added{First, we bound the CGF corresponding to $\bar{\Psi}^{\rom{1}}(\beta)$. If the loss is sub-gamma with variance factor $s_{\rom{1}}^2$ and scale parameter $c_{\rom{1}}$ under the two-level prior and any data distribution $\calD$ in the support of $\calT$, i.e., 
\begin{equation} \label{eq:dskfdshjfb}
    \E_{P \sim \calP} \E_{h \sim P}  \E_{z \sim \calD} \left[ e^{\nu (\calL(\calD, h) - l(h, z)) } \right] \leq \exp \left( \frac{\nu^2 s_{\rom{1}}^2}{2(1- \nu c_{\rom{1}})} \right) ~~ \forall ~ \nu \in (0, 1 / c_{\rom{1}})
\end{equation}
then, with $\nu := \beta / m$, we have that, if $\beta < \frac{m}{c_{\rom{1}}}$,
\begin{equation} \label{eq:kjnkjdfbgsdmnbgfd}
    \frac{m}{\beta} \log \E_{\calP} \E_{P} \E_{\calD} \left[  e^{ \frac{\beta}{m} V^{\rom{1}} } \right]  \leq \frac{m}{\beta} \frac{\beta^2 s_{\rom{1}}^2}{2 m^2(1- \frac{c_{\rom{1}} \beta}{m})}  = \frac{\beta s_{\rom{1}}^2}{2 m(1- \frac{c_{\rom{1}} \beta}{m})}  = \bar{\Psi}^{\rom{1}}(\beta) ~~~ \forall ~\calD ~.
\end{equation}
Second, we bound the remaining CGF corresponding to $\Psi^\rom{2}(\lambda)$. For that, we use the assumption that the random variable $V_i^\rom{2} := \E_{{\calT_h}} \left[ \calL (Q_i, \calD)  \right] - \calL(Q_i, \calD_i )$ is sub-gamma with variance factor $s_{\rom{2}}^2$ and scale parameter $c_{\rom{2}}$ under the hyper-prior $\calP$ and the task distribution ${\calT_h}$. That is, its moment generating function can be bounded by that of a Gamma distribution $\Gamma(s_{\rom{2}}^2, c_{\rom{2}})$:
\begin{equation}
 \E_{(\calD, S) \sim {\calT_h}} \E_{P \sim \calP} \left[  e^{\kappa  \E_{(D,S)  \sim {\calT_h}} \left[ \calL (Q(P, S), \calD) \right] - \calL (Q(P, S), \calD )}\right] \leq   \exp \left(\frac{\kappa^2 s_{\rom{2}}^2}{2(1- c_{\rom{2}} \kappa)} \right) ~ \forall \kappa \in ( 0, 1 / c_{\rom{2}} ) ~.
\end{equation}
Using the sub-gamma assumption with $\kappa := \lambda / n$, we obtain that
\begin{equation}
     \Psi^\rom{2}(\lambda) \leq  \frac{\lambda^2 s^2_{\rom{2}}}{2 n^2 (1 - \frac{\lambda c_{\rom{2}}}{n})}  ~, \label{eq:kjndf2832ksdjcjsd}
\end{equation}
under the condition that $\lambda < \frac{n}{c_{\rom{2}}}$.}

\added{Finally, we insert (\ref{eq:kjndf2832ksdjcjsd}) and (\ref{eq:kjnkjdfbgsdmnbgfd}) into 
(\ref{eq:ksdfjhkdsbf}) to obtain that, if $\beta < m / c_{\rom{1}}$ and $\lambda \leq n / c_{\rom{2}}$, Theorem \ref{theorem:meta-learning-bound} holds with 
\begin{align}
C(\lambda, \beta, \delta) = \frac{\beta s_{\rom{1}}^2}{2  m (1- \frac{ c_{\rom{1}} \beta}{m})} + \frac{\lambda s^2_{\rom{2}}}{2 n (1 - \frac{\lambda c_{\rom{2}}}{n})} - \frac{1}{\sqrt{n}} \log \delta
\end{align}
with probability at least $1-\delta$. This concludes our high-probability bound on the CGFs in the case of sub-gamma tail assumptions.}
\subsection{Proof of Corollary \ref{cor:bayesian_learner_PAC_bound}}
\added{The proof of Corollary \ref{cor:bayesian_learner_PAC_bound} is inspired by \citet{germain2016pac}.}When we choose the posterior $Q$ as the optimal Gibbs posterior $Q^*_i := Q^*(S_i, P)$, we have that 
\begin{align}
& \hat{\calL}(\calQ, S_1, ..., S_n) + \frac{1}{n} \sum_{i=1}^n \frac{1}{\beta} \E_{P \sim \calQ} \left[ D_{KL}(Q^*_i || P)\right] \\
=~& \frac{1}{n} \sum_{i=1}^n \left( \E_{P \sim \calQ} \E_{h \sim Q^*_i} \left[ \hat{\calL}(h, S_i) \right]  + \frac{1}{\beta} \left( \E_{P \sim \calQ} \left[ D_{KL}(Q^*_i || P)\right] \right) \right) \label{eq:task_objective_in_bound} \\
=~& \frac{1}{n} \sum_{i=1}^n \frac{1}{\beta} \left(\E_{P \sim \calQ} \E_{h \sim Q^*_i} \left[  \beta \hat{\calL}(h, S_i)  +  \log \frac{P(h) e^{-  \beta \hat{\calL}(h, S_i) }}{P(h) Z_\beta(S_i, P)} \right] \right) \\
=~& \frac{1}{n} \sum_{i=1}^n \frac{1}{\beta} \left(- \E_{P \sim \calQ} \left[ \log Z_\beta(S_i, P)\right] \right) \;.
\end{align}

This allows us to write the inequality in (\ref{eq:meta-learning-bound}) as 
\begin{align} \label{eq:meta_pac_bound_mll_appendix}
\calL(\calQ, \calT)  \leq ~  & - \frac{1}{n} \sum_{i=1}^n \frac{1}{\beta} \E_{P \sim \calQ} \left[\log Z_\beta(S_i, P) \right]  + \left(\frac{1}{\lambda} + \frac{1}{n\beta}\right)  D_{KL}(\calQ||\calP) + C(\delta, \lambda, \beta) \;.
\end{align}

According to Lemma~\ref{lemma:optimal_gibbs_posterior}, the Gibbs posterior $Q^*(S_i, P)$ is the minimizer of (\ref{eq:task_objective_in_bound}), in particular $\forall P \in \calM(\calH), \forall i=1, ..., n:$
\begin{equation}
Q^*(S_i, P) = \frac{P(h)e^{- \beta \hat{\calL}(h,S_i)}}{Z_\beta(S_i,P)} = \argmin_{Q \in \calM(\calH)} \E_{h \sim Q} \left[ \hat{\calL}(h, S_i) \right]  + \frac{1}{\beta} D_{KL}(Q || P)  \;.
\end{equation}

Hence, we can write
\begin{align*}
\calL(\calQ, \calT)  \leq & - \frac{1}{n} \sum_{i=1}^n \frac{1}{\beta} \E_{P \sim \calQ} \left[\log Z_\beta(S_i, P) \right]  + \left(\frac{1}{\lambda} + \frac{1}{n\beta}\right)  D_{KL}(\calQ||\calP) + C(\delta, \lambda, \beta) \\
= & ~ \frac{1}{n} \sum_{i=1}^n \E_{P \sim \calQ} \left[\min_{Q \in \calM(\calH)} \hat{\calL}(Q, S_i)  + \frac{1}{\beta}  D_{KL}(Q || P) \right]  + \left(\frac{1}{\lambda} + \frac{1}{n\beta}\right)  D_{KL}(\calQ||\calP) + C(\delta, n, \overline{m}) \\
\leq 
& \frac{1}{n} \sum_{i=1}^n \E_{P \sim \calQ} \left[\hat{\calL}(Q, S_i)  + \frac{1}{\beta}  D_{KL}(Q || P) \right] + \left(\frac{1}{\lambda} + \frac{1}{n\beta}\right)  D_{KL}(\calQ||\calP) + C
\end{align*}
which proves that the bound for Gibbs-optimal base learners in (\ref{eq:meta_pac_bound_mll_appendix}) and (\ref{eq:meta-level_pac_bound_with_mll}) is\added{smaller}than the bound in Theorem~\ref{theorem:meta-learning-bound} which holds uniformly for all $Q \in \calM(\calH)$.

\subsection{Proof of Proposition~\ref{proposition:pacoh_optimal_hyper_posterior}: PAC-Optimal Hyper-Posterior}
\label{appendix:proof_optimal-hyper-posterior}
An objective function corresponding to (\ref{eq:meta-level_pac_bound_with_mll}) reads as
\begin{equation} \label{eq:meta-objective_with_mll}
J(\calQ) =  - \E_{\calQ} \left[ \frac{\lambda}{n\beta + \lambda} \sum_{i=1}^n \log Z(S_i, P) \right] + D_{KL}(\calQ||\calP) \;.
\end{equation}
To obtain $J(\calQ)$, we omit all additive terms from (\ref{eq:meta-level_pac_bound_with_mll}) that do not depend on $\calQ$ and multiply by the scaling factor $\frac{\lambda n \beta}{n \beta + \lambda}$. Since the described transformations are monotone, the minimizing distribution of $J(\calQ)$, that is,
\begin{equation}
\calQ^* = \argmin_{\calQ \in \calM(\calM(\calH))} J(\calQ) \;,
\end{equation}
is also the minimizer of (\ref{eq:meta-level_pac_bound_with_mll}). More importantly, $J(\calQ)$ is structurally similar to the generic minimization problem in Lemma \ref{lemma:optimal_gibbs_posterior}. Hence, we can invoke Lemma~\ref{lemma:optimal_gibbs_posterior} with $A = \calM(\calH)$, $g(a) = - \sum_{i=1}^n \log Z(S_i, P)$, $\beta = \frac{1}{\sqrt{n\overline{m}} + 1}$, to show that the optimal hyper-posterior is
\begin{equation}
\calQ^*(P) = \frac{\calP(P) \exp \left( \frac{\lambda}{n\beta + \lambda}\sum_{i=1}^n \log Z_\beta(S_i, P) \right) }{Z^{\rom{2}}(S_1, ..., S_n, \calP)} \;,
\end{equation}
wherein 
$
 Z^{\rom{2}}(S_1, ..., S_n, \calP) = \E_{P \sim \calP} \left[ \exp \left( \frac{\lambda}{n\beta + \lambda} \sum_{i=1}^n \log Z_\beta(S_i, P) \right)  \right] \;.
$ $\hfill \Box$

\subsection{Proof of Corollary \ref{cor:meta_bound_pacoh}}
The proof follows the same scheme as the proof of Corollary \ref{cor:bayesian_learner_PAC_bound}. However, for completeness, we'll state it the following:
If we choose $\calQ = \calQ^*$, the first two terms of the PAC-Bayes bound in (\ref{eq:meta-level_pac_bound_with_mll}) can be re-arranged as follows:
\begin{align}
& - \frac{1}{n} \sum_{i=1}^n \frac{1}{\beta} \E_{P \sim \calQ} \left[\log Z_\beta(S_i, P) \right]  + \left(\frac{1}{\lambda} + \frac{1}{n\beta}\right)  D_{KL}(\calQ^*||\calP) \\
 = & \E_{P \sim \calQ} \left[ - \frac{1}{n} \sum_{i=1}^n \frac{1}{\beta} \log Z_\beta(S_i, P)   + \left(\frac{1}{\lambda} + \frac{1}{n\beta}\right) \log \frac{\calP(P) \exp \left( \frac{\lambda}{n\beta + \lambda}\sum_{i=1}^n \log Z_\beta(S_i, P) \right) }{\calP(P) Z^{\rom{2}}(S_1, ..., S_n, \calP)} \right] \\
  = &  \E_{P \sim \calQ} \left[  - \left(\frac{1}{\lambda} + \frac{1}{n\beta}\right) \log Z^{\rom{2}}(S_1, ..., S_n, \calP) \right]
  =   - \left(\frac{1}{\lambda} + \frac{1}{n\beta}\right)\log Z^{\rom{2}}(S_1, ..., S_n, \calP) \label{eq:lnz2} 
\end{align}
Hence, if we insert (\ref{eq:lnz2}) in (\ref{eq:meta-level_pac_bound_with_mll}), we obtain
\begin{align}
\calL(\calQ, \calT)  &\leq - \left(\frac{1}{\lambda} + \frac{1}{n\beta}\right) \log Z^{\rom{2}}(S_1, ..., S_n, \calP) + C(\delta, \lambda, \beta) \;. \label{eq:pac_bound_z2_appdx}
\end{align}
which concludes the proof. $\hfill \Box$


\subsection{Proof of Theorem \ref{theorem:single_task_bound_hyper_prior}}

\paragraph{Step 1: Uniform bound for any $Q_i \in \calM(\calH)$.}
In the following, we bound the generalization error of a base learner $Q(S, P)$  with priors $P \sim \calP$ when applied to a\added{new task $\tau \sim \calT_h$, i.e., 
\begin{equation}
    \calL(\calP, \calT) =  \E_{P \sim \cal\calP}  \E_{(\calD, S) \sim \calT_h} \left[ \calL(Q(S, P), \calD) \right] ~.
\end{equation}
Similar to the proof of Theorem \ref{theorem:meta-learning-bound}, we first bound the intermediary average generalization error across a set of given tasks $\{\tau_1, ..., \tau_n \}$, i.e.,}
$$
\calL(\calP, \tau_1, ..., \tau_n) =  \E_{P \sim \calP} \left[\frac{1}{n} \sum_{i=1}^n \calL (Q(P, S_i), \calD_i) \right] .
$$
However, unlike in the meta-learning case, the priors are directly sampled from the hyper-prior $\calP$ instead of a data-dependent hyper-posterior $\calQ$. Thus, we only need to apply the change-of-measure inequality once.
In particular, we apply Lemma~\ref{lemma:concentration_sum_of_rvs_2} with the following instantiations. We set $A = \calM(\calH) \times \calH^n$, $f = (P, h_1, ..., h_n) \in A$, $\pi = (\calP, P^n)$ and $\rho = (\calP, Q^n)$ as a distribution over the joint two-level hypotheses corresponding to hierarchical sampling via $P \sim \calP$ and $h_i \sim Q_i=Q(S_i, P) ~ \forall i = 1, ..., n$. Furthermore, we set $g(f) = g(P, h_1, ..., h_n) = \frac{1}{n} \sum_{i=1}^n \calL(h_i, \calD_i) - \hat{\calL}(h_i, S_i)$ and $\lambda = \gamma$. Hence, we obtain that, for $\gamma > 0$,
%
\begin{align}
\begin{split}
\calL(\calP, \tau_1, ..., \tau_n) \leq & ~
\frac{1}{n} \sum_{i=1}^n \E_{P \sim \calP} \left[ \calL(Q_i,S_i) \right] +  \frac{1}{\gamma} \sum_{i=1}^n \E_{P \sim \calP} \left[ D_{KL}(Q_i || P)\right] \\
&+  \frac{1}{\gamma} \underbrace{\log \E_{P \sim \calP} \E_{h \sim P}  \left[ e^{ \frac{\gamma}{n} \sum_{i=1}^n (\calL(h, \calD_i) - \hat{\calL}(h, S_i))}\right]}_{\Upsilon^{\rom{1}}(\gamma)} ~.\\ 
\end{split} \label{eq:dksjbfkjsdfs}
\end{align}
\added{Next, we bound the difference between $\calL(\calP, \calT_h)$ and $\calL(\calP, \tau_1, ..., \tau_n)$. Unlike the meta-learning bound, we do not require the change of measure lemma since there is no meta-learned hyper-posterior. Instead, we can simply use Jensen's inequality to obtain that
\begin{align} \label{eq:sdkjnflkjzbgkjh}
 \calL(\calP, \calT) - \calL(\calP, \tau_1, ..., \tau_n) \leq & \frac{1}{\lambda}\underbrace{\log \E_{P \sim \calP} \left[ e^{ \frac{\lambda}{n} \sum_{i=1}^n \E_{\calD \sim \calT} \E_{S \sim \calD^m} \left[ \calL(Q(S,P), \calD) \right] - \calL(Q(S_i,P), \calD_i) }  \right]}_{\Upsilon^{\rom{2}}(\lambda)} ~.
\end{align}
Combining (\ref{eq:sdkjnflkjzbgkjh}) with (\ref{eq:dksjbfkjsdfs}) we have that
\begin{align}
    \calL(\calP, \calT) \leq \frac{1}{n} \sum_{i=1}^n \E_{P \sim \calP} \left[ \calL(Q_i,S_i) \right] +  \frac{1}{\gamma} \sum_{i=1}^n \E_{P \sim \calP} \left[ D_{KL}(Q_i || P) \right] + \Upsilon^{\rom{1}}(\gamma) + \Upsilon^{\rom{2}}(\lambda) ~.
\end{align}}
\added{
Now, it remains to bound (with high probability) the CGFs $\Upsilon^{\rom{1}}(\gamma) + \Upsilon^{\rom{2}}(\lambda)$ which are identical to those in the proof of Theorem \ref{theorem:meta-learning-bound}. Using the results from there and $\gamma:= n \beta$ we have with probability at least $1-\delta$ that 
\begin{align}
    \calL(\calP, \calT) \leq & ~\frac{1}{n} \sum_{i=1}^n \E_{P \sim \calP} \left[ \calL(Q_i,S_i) \right] +  \frac{1}{n\beta} \sum_{i=1}^n \E_{P \sim \calP} \left[ D_{KL}(Q_i || P) \right]   \\ & ~ + \underbrace{\bar{\Psi}^{\rom{1}}(\beta) + \Psi^{\rom{2}}(\lambda) + \frac{1}{\sqrt{n}} \log \frac{1}{\delta}}_{ = C(\beta, \lambda, \delta)}.
\end{align}
}

\paragraph{Step 2: Bound for Gibbs learner.}
\looseness -1 Now, we assume a Gibbs distribution as base learner $Q^*(S_i, P) := \frac{P(h)e^{- \beta \hat{\calL}(h,S_i)}}{Z_\beta(S_i,P)}$. Following the same steps as in Appendix \ref{cor:bayesian_learner_PAC_bound}, we obtain
\begin{align} 
\label{eq:step2_bound}
\added{\calL(\calP, \calT)}\leq & ~
- \frac{1}{n \beta} \sum_{i=1}^n \E_{P \sim \calP} \left[  \log Z_\beta(S_i, P) \right] +\added{C(\beta, \lambda, \delta)}
\end{align}
which concludes the proof of the bound. $\hfill \Box$

\subsection{Proof of Proposition \ref{prop:single_vs_meta}}




\looseness -1 To show that meta-learning improves upon single-task learning, we study the difference between (\ref{eq:pac_bound_z2}) and (\ref{eq:single_task_bound_hyper_prior}), i.e., $\Delta = (\ref{eq:single_task_bound_hyper_prior}) - (\ref{eq:pac_bound_z2})$. After removing terms that cancel out, we have that
%
\added{\begin{align} \label{eq:condition_meta_learning_better}
\Delta = & \left(\frac{1}{\lambda} + \frac{1}{n\beta}\right) \log Z^{\rom{2}}(S_1, ..., S_n, \calP)  -  \frac{1}{n} \sum_{i=1}^n \frac{1}{\beta} \E_{P \sim \cal\calP} \left[\log Z_\beta(S_i, P) \right] \\
= & \left(\frac{1}{\lambda} + \frac{1}{n\beta}\right) \bigg[ \log \E_{P \sim \calP} \left[ \exp \left( \frac{\lambda}{n\beta + \lambda} \sum_{i=1}^n \log Z_\beta(S_i, P) \right)  \right] \\
& - \E_{P \sim \calP} \left[ \frac{\lambda}{n\beta + \lambda}  \sum_{i=1}^n  \log Z_\beta(S_i, P) \right] \bigg] \\
    =& \left(\frac{1}{\lambda} + \frac{1}{n\beta}\right) \log \E_{P \sim \calP} \left[ \exp \left( \frac{\lambda}{n\beta + \lambda} \sum_{i=1}^n \left( \log Z_\beta(S_i, P)   -   \E_{P \sim \calP}  \left[ \log Z_\beta(S_i, P)   \right] \right) \right) \right] 
\end{align}
Using $\lambda = \sqrt{m}$ and $\beta = \sqrt{m}$, we can re-write this as 
\begin{align} \label{eq:condition_meta_learning_better2}
\Delta = \left(\frac{1}{\sqrt{m}} + \frac{1}{n \sqrt{m}}\right) \log \E_{P \sim \calP} \left[ \exp \left( \frac{1}{\sqrt{mn} + 1} \sum_{i=1}^n \left( \log Z(S_i, P)   -   \E_{P \sim \calP}  \left[ \log Z(S_i, P)   \right] \right) \right) \right] 
\end{align}
where we have abbreviated $Z_{\sqrt{n}}(S_i, P)$ as $Z(S_i, P)$.}

\added{Finally, we show that $\Delta$ is non-negative via Jensen's inequality. In particular, the non-negativity follows from
\begin{align}
 & \log \E_{P \sim \calP} \left[ e^{ \frac{1}{\sqrt{nm} + 1} \sum_{i=1}^n \left( \log Z_\beta(S_i, P)   -   \E_{P \sim \calP}  \left[ \log Z_\beta(S_i, P)   \right] \right) } \right] \\
\geq & ~ \frac{1}{\sqrt{nm} + 1}  \sum_{i=1}^n \E_{P \sim \calP} \left[  \left( \log Z_\beta(S_i, P)   -   \E_{P \sim \calP}  \left[ \log Z_\beta(S_i, P)   \right] \right)  \right]  \\
 = & ~ 0 ~.
\end{align}}

\subsection{Proof of the Equivalence of Variational Inference and Minimization of the PAC-Bayes Meta-Learning Bound}
\label{appendix:proof_eqivalence_vi_pac_bound}

\addedII{Here, we show that performing variational inference \citep{Blei2016} on the PACOH $\calQ^*$ is equivalent to minizing the PAC-Bayesian meta-learning bound in (\ref{eq:meta-level_pac_bound_with_mll}). A similar connection for per-task learning has previously been pointed out by \citet{thakur2019unifying}.}We can write the optimal variational distribution $\tilde{\calQ}$ with respect to $\calQ^*$ as
\begin{align} 
\tilde{\calQ} & = \argmin_{\calQ \in \calF} ~ D_{KL}(\calQ || \calQ^*) 
= \argmin_{\calQ \in \calF} ~ \E_{P \sim \calQ}  \left[ \log \calQ(P) - \log \calQ^*(P)\right] \\
& =  \argmin_{\calQ \in \calF} ~ \E_{P \sim \calQ}  \left[ \log \calQ(P) - \log \calP(P) -  \left(  \frac{\beta}{\beta + \lambda} \sum_{i=1}^n \frac{1}{\beta_i} \log Z(S_i, P) \right) + \log Z^{\rom{2}} \right] \\
& =  \argmin_{\calQ \in \calF}  ~ - \frac{1}{n} \sum_{i=1}^n \frac{1}{\beta_i} \E_{P \sim \calQ}  \left[  \log Z(S_i, P)  \right] + \left(\frac{1}{\lambda} + \frac{1}{\beta}\right) D_{KL}(\calQ || \calP) \;.
\label{eq:proof_eqivalence_vi_pac_bound_step4}
\end{align}
%
Now it is straightforward to see that (\ref{eq:proof_eqivalence_vi_pac_bound_step4}) is the same as the meta-learning PAC-Bayes bound in (\ref{eq:meta-level_pac_bound_with_mll}) up to the constant $C(\delta, n, \beta)$. Hence, we can conclude that variational inference with respect to $\calQ^*$ is equivalent to minimizing (\ref{eq:meta-level_pac_bound_with_mll}) over the same variational family $\calF$. 

%% file: content/supplement_blr.tex
\section{Bounding the CGF for linear regression}
\label{appendix:bounding_blr_cumulants}

\added{The setting and analysis in this section is inspired by \citet{germain2016pac} who provide CGF bounds for per-task learning with linear regression. However, we bound the CGFs for meta-learning which is significantly more involved due to the hierarchical nature of the priors and data-generating process.}

\paragraph{Assumptions:} Regression problem with $\calX \times \calY \subset \R^d \times \R$, family of linear predictors: $\calH = \{ h_{\bw}(\bx) = \bw^\top \bx ~| ~ \bw \in \R^d \} $, Gaussian prior $P_{\mu_P, \sigma^2_P} = \calN(\mu_P, \sigma^2_P \bI)$, Gaussian hyper-prior $\calP(\mu_p) = \calN(0, \sigma^2_{\calP} \bI)$, Loss function $l(\bw, \bx, y) = \frac{1}{2} \ln(2\pi \sigma^2) + \frac{1}{2\sigma^2}(\bw^\top \bx - y)^2$.

\paragraph{Data generating process:}
$\bw_i^* \sim \calN(\mu_\calT, \sigma^2_\calT \bI)$, $\bx \sim p(\bx) = \calN(0, \sigma_{\bx}^2)$, $y = {\bw_i^*}^\top \bx + \epsilon$ where $\epsilon \sim \calN(0, \sigma_{\epsilon}^2)$. Thus, we can write the conditional label distribution as $p_{\bw_i^*}(y|\bx) = \calN({\bw_i^*}^\top  \bx, \sigma_{\epsilon}^2)$. Accordingly, the data distribution follows as $\calD_i = p(\bx) \, p_{\bw_i^*}(y|\bx)$.

\paragraph{Bounding} $\mathbf{\Psi^{\rom{1}}(\gamma)}$:
We aim to bound 
\begin{equation}
\Phi^{\rom{1}}(\beta) = \frac{1}{n \beta} \sum_{i=1}^n  \sum_{j=1}^{m} \ln \E_{\calP} \E_{P} \E_{\calD_i} \left[  e^{ \frac{\beta}{m} V_{ij}^\rom{1}}\right]
\end{equation}
For that, we focus on the cumulant-generating function of $V_{ij} = \calL(h, \calD_i) - l(h, z) = \E_{(\bx, y) \sim \calD_i} \left[ l(\bw, \bx, y) \right]- l(\bw, \bx, y)$, i.e.,

\begin{equation}
\Gamma^{\rom{1}}(\gamma) = \ln \E_{\calP} \E_{P} \E_{\calD_i} e^{\gamma V_i} \leq \frac{\gamma^2 s^2}{2(1-\gamma c) } ~, \quad \forall \gamma \in (0, 1/c) ~.
\end{equation}

\begin{align}
\Gamma^{\rom{1}}(\gamma) = & \ln \E_{\bw} \E_{(\bx, y)} \exp \left(  \frac{\gamma}{2\sigma^2} \left(  \E_{(\bx, y)} \left[ l(\bw, \bx, y) \right]- l(\bw, \bx, y) \right) \right) \\
= & \ln \E_{\bw} \E_{(\bx, y)}  \exp \left(\frac{\gamma}{2\sigma^2} \left( \sigma_{\bx}^2 ||\bw - \bw^*_i||^2  + \sigma_\epsilon^2 -  (\bw^\top \bx - y)^2  \right) \right) \\
 = & \ln  \E_{\bw} \E_{\bx} \E_{\epsilon} \exp \left(\frac{\gamma}{2\sigma^2} \left( \sigma_{\bx}^2 ||\bw - \bw^*_i||^2  + \sigma_\epsilon^2 -  ((\bw - \bw^*_i)^\top \bx  + \epsilon)^2  \right) 
 \right) \\
  = & \ln  \E_{\bw} \left[ \exp \left(\frac{\gamma}{2\sigma^2} \left(\sigma_{\bx}^2 ||\bw - \bw^*_i||^2  + \sigma_\epsilon^2 \right) \right)  \underbrace{\E_{\bx} \E_{\epsilon} \exp \left( - \frac{\gamma}{2 da\sigma^2} ((\bw - \bw^*_i)^\top \bx  + \epsilon)^2\right)}_{(*)}  \right] \label{eq:log_laplace_11}
\end{align}
\vspace{-12pt}
\begin{align}
(*) = & \E_{\bx} \E_{\epsilon}  \exp \left(- \frac{\gamma}{2\sigma^2} ((\bw - \bw^*_i)^\top \bx  + \epsilon)^2  \right) \\
= &\left(1 +  \gamma \frac{\sigma^2_\epsilon}{\sigma^2}\right)^{-\frac{1}{2}} \E_{\bx}  \exp \left( - \frac{\frac{\gamma}{2\sigma^2}}{1 + \gamma \frac{\sigma^2_\epsilon}{\sigma^2}} ((\bw - \bw^*_i)^\top \bx )^2 \right) \\
= &\left(1 + \gamma \frac{\sigma^2_\epsilon}{\sigma^2}\right)^{-\frac{1}{2}} \left( 1 + \frac{\frac{\gamma}{\sigma^2}}{1 +  \gamma \frac{\sigma^2_\epsilon}{\sigma^2} } ||\bw - \bw^*_i||^2 \sigma^2_\bx \right)^{-\frac{1}{2}} \\
= & \left(1 + \gamma \frac{\sigma^2_\epsilon}{\sigma^2} + \frac{\gamma}{\sigma^2} ||\bw - \bw^*_i||^2 \sigma^2_\bx \right)^{-\frac{1}{2}} \label{eq:second_mgf_term_closed_form2}
\end{align}

Now, we can insert (\ref{eq:second_mgf_term_closed_form2}) into (\ref{eq:log_laplace_11}):

\begin{align}
\Gamma^{\rom{1}}(\gamma) = & \ln  \E_{\bw \sim \calN(0,\sigma_P^2 I_d)} \left[ \frac{\exp \left(\frac{\gamma}{2 \sigma^2} \left( \sigma_{\bx}^2 ||\bw - \bw^*_i||^2  + \sigma_\epsilon^2 \right) \right)}{(1 + \gamma \frac{\sigma^2_\epsilon}{\sigma^2} + \frac{\gamma}{\sigma^2} ||\bw - \bw^*_i||^2 \sigma^2_\bx)^{\frac{1}{2}}} \right] \\
\leq &  \frac{\gamma}{2 \sigma^2} \left( \sigma_{\bx}^2 ||\bw^*_i||^2  + \sigma_\epsilon^2 \right) + \ln  \E_{\bw \sim \calN(0,\sigma_P^2 \bI_d)} \left[ \frac{\exp \left(\frac{\gamma}{2 \sigma^2} \sigma_{\bx}^2 ||\bw||^2  \right)}{(1 + \gamma \frac{\sigma^2_\epsilon}{\sigma^2} + \gamma \frac{\sigma_\bx^2}{\sigma^2} ||\bw - \bw^*_i||^2 )^{\frac{1}{2}}} \right]\\
\leq &  \frac{\gamma}{2 \sigma^2} \left( \sigma_{\bx}^2 ||\bw^*_i||^2  + \sigma_\epsilon^2 \right) + \ln  \E_{w \sim \calN(0,\sigma_P^2)} \left[ \frac{\exp \left(\frac{\gamma}{2 \sigma^2}  \sigma_{\bx}^2 d w^2  \right)}{(1 + \gamma \frac{\sigma^2_\epsilon}{\sigma^2}  + \gamma \frac{ \sigma^2_\bx}{\sigma^2} ||\bw^*_i||^2 + \gamma \frac{\sigma_{\bx}^2}{\sigma^2} d w^2)^{\frac{1}{2}}} \right] \label{eq:345345}
\end{align}

\begin{lemma} \label{lemma:bound_bessel2}
Let $a, b \in \R^+ $ and $x \sim \calN(0, \sigma^2)$ be a Gaussian random variable with mean $0$ and variance $ \sigma^2$. Then, we have
\begin{equation}
\E_{x \sim \calN(0, \sigma^2)} \left[ \frac{e^{a x^2}}{(b + 2ax^2 )^{\frac{1}{2}}} \right] < \frac{1}{(b -2 b a \sigma^2 ))^{\frac{1}{2}}}
\end{equation}
\end{lemma}
\begin{proof}
\begin{equation}
E_{x \sim \calN(0, \sigma^2)} \left[ \frac{e^{a x^2}}{(b + 2ax^2 )^{\frac{1}{2}}} \right] = \frac{e^{\frac{b}{8} \left(\frac{1}{a \sigma^2}-2\right)} K_0\left(\frac{b}{8} \left(\frac{1}{a \sigma^2}-2\right)\right)}{\sqrt{4 a \pi \sigma^2 }} \label{eq:closed_form_expectation_mgf_bessel2}
\end{equation}
wherein $K_v(y)$ is the modified Bessel function of the second kind.  In particular, for $v=0$ we have
\begin{equation}
K_0(y) = \int_0^\infty e^{-y \cosh(t)}dt < \frac{\sqrt{\pi} e^{-y}}{\sqrt{2y}} \label{eq:upper_bound_bessel2}
\end{equation}
For a proof of the inequality in (\ref{eq:upper_bound_bessel2}), we refer to \citet{yang2017approximating}. Using (\ref{eq:upper_bound_bessel2}) to upper-bound (\ref{eq:closed_form_expectation_mgf_bessel2}), we obtain
\begin{align}
E_{x \sim \calN(0, \sigma^2)} \left[ \frac{e^{a x^2}}{(b + 2ax^2)^{\frac{1}{2}}} \right] <& \frac{1}{\sqrt{ b (1 -2 a \sigma^2 ) }}
\end{align}
which concludes the proof.
\end{proof}


Next, we invoke Lemma \ref{lemma:bound_bessel2} with $a=\frac{\gamma}{2\sigma^2} \sigma_{\bx}^2 d$ and $b= 1 + \frac{\gamma}{\sigma^2} \sigma^2_\epsilon +  \frac{\gamma}{\sigma^2} \sigma^2_\bx ||\bw^*_i||^2 $, such that 
\begin{align}
(\ref{eq:345345}) \leq &  \frac{\gamma}{2 \sigma^2} \left( \sigma_{\bx}^2 ||\bw^*_i||^2  + \sigma_\epsilon^2 \right) - \frac{1}{2}\ln \left(b (1 -2 a (\sigma_P^2 + \sigma_\calP^2)) \right) \\
 = &  \frac{\gamma}{2 \sigma^2} \underbrace{\left( \sigma_{\bx}^2 ||\bw^*_i||^2  + \sigma_\epsilon^2 \right)}_{\vartheta_i} - \frac{1}{2}\ln \left(\left(1 + \frac{\gamma}{\sigma^2} \underbrace{\left( \sigma_{\bx}^2 ||\bw^*_i||^2  + \sigma_\epsilon^2 \right)}_{\vartheta_i}\right) \left(1 -\frac{\gamma}{\sigma^2} d \sigma_{\bx}^2 (\sigma_P^2 + \sigma_\calP^2)\right) \right) \\
  = &  \frac{\gamma}{2\sigma^2} \vartheta_i - \frac{1}{2}\ln \left((1 + \frac{\gamma}{\sigma^2} \vartheta_i - \frac{\gamma}{\sigma^2} d \sigma_{\bx}^2 (\sigma_P^2 + \sigma_\calP^2) - \frac{\gamma^2}{\sigma^4} d  \sigma_{\bx}^2 (\sigma_P^2 + \sigma_\calP^2) \vartheta_i \right) \\
 \leq & \frac{\gamma}{2 \sigma^2} \vartheta_i + \frac{ \frac{\gamma}{2 \sigma^2} d \sigma_{\bx}^2 (\sigma_P^2 + \sigma_\calP^2) +  \frac{\gamma^2}{2\sigma^4} d  \sigma_{\bx}^2 (\sigma_P^2 + \sigma_\calP^2) \vartheta_i - \frac{\gamma}{2 \sigma^2} \vartheta_i}{1 + \frac{\gamma}{\sigma^2} \vartheta_i -  \frac{\gamma}{\sigma^2} d \sigma_{\bx}^2 (\sigma_P^2 + \sigma_\calP^2) - \frac{\gamma^2}{\sigma^4} d  \sigma_{\bx}^2 (\sigma_P^2 + \sigma_\calP^2) \vartheta_i } \\
  = & \frac{\gamma}{2\sigma^2} \vartheta_i + \frac{ \frac{\gamma}{2} c_i}{1 - \gamma c_i } 
 = \frac{  \gamma^2 \left( \frac{\vartheta_i}{\sigma^2}(\frac{1}{\gamma} - c_i) +  \frac{c_i}{\gamma} \right)}{2(1 - \gamma c_i) } 
       =  \frac{  \gamma^2 s_i^2}{2(1 - \gamma c_i) } 
\end{align}

with  $s_i^2 = \frac{\vartheta_i}{\sigma^2}(\frac{1}{\gamma} - c_i) +  \frac{c_i}{\gamma}$, $c_i = \frac{d}{\sigma^2} \sigma_{\bx}^2 (\sigma_P^2 + \sigma_\calP^2) + \frac{\gamma}{\sigma^4} d  \sigma_{\bx}^2 (\sigma_P^2 + \sigma_\calP^2) \vartheta_i -  \frac{\vartheta_i}{\sigma^2}$ and $\vartheta_i = \sigma_{\bx}^2 ||\bw^*_i||^2  + \sigma_\epsilon^2$.


\paragraph{Bounding $\Psi^{\rom{2}}(\gamma)$:}

To show that the cumulant-generating function of the random variable $V^{\rom{2}}:= \E_{(D,S)  \sim \calT} \E_{P \sim \calP} \left[ \calL (Q(P, S), \calD) \right] - \calL (Q(P, S_i), \calD_i )$ is {\em sub-gamma}, we aim to find values $s_{\rom{2}}^2 > 0$ and parameter $c_{\rom{2}} > 0$ such that 
\begin{equation}
\Psi^{\rom{2}}(\gamma) = \ln \E_{(\calD, S) \sim \calT} \E_{P \sim \calP} \left[ \exp \left( \gamma V^{\rom{2}} \right) \right] \leq   \exp \left(\frac{\gamma^2 s_{\rom{2}}^2}{2(1- c_{\rom{2}} \gamma)} \right) \quad \forall \gamma \in ( 0, 1 / c_{\rom{2}} )
\end{equation}
\begin{align*}
\Psi^{\rom{2}}(\lambda) =& \ln \E_{\calT} \E_{\calP}  \exp \left( \frac{\lambda}{2\sigma^2}  \left( \E_{\calT} \E_{\calP} \left[ \calL (Q(P, \calS), \calD) \right] - \calL (Q(P, \calS), \calD ) \right) \right) \\
= & 
\ln \E_{\calT} \E_{\calP}  \exp \bigg( \frac{\lambda}{2\sigma^2}  \left( \E_{\calT} \E_{\calP} \left[ E_{\bw \sim Q} \left[ \sigma_\bx^2 ||\bw^* - \bw||^2  \right]\right]  - \E_{\bw \sim Q} \left[ \sigma_\bx^2 ||\bw^* - \bw||^2 \right] \right) \bigg)  \\
\leq & 
\ln \E_{\calT} \E_{\calP}  \exp \bigg( \frac{\lambda}{2\sigma^2}  \left( \E_{\calT} \E_{\calP} \left[ E_{\bw \sim P} \left[ \sigma_\bx^2 ||\bw^* - \bw||^2  \right]\right]  - \E_{\bw \sim P} \left[ \sigma_\bx^2 ||\bw^* - \bw||^2 \right] \right) \bigg)  \\
=& \ln \E_{\calT} \E_{\calP} \exp \bigg( \frac{\lambda}{2\sigma^2}  \left( \left[ \sigma_\bx^2( ||\mu_\calT||^2 + d\sigma^2_\calT + d \sigma^2_\calP) \right]  -( \sigma_\bx^2 ||\bw^* - \mu_P||^2 ) \right) \bigg)  \\
=& 
\frac{\lambda}{2\sigma^2}   \sigma_\bx^2( ||\mu_\calT||^2 + d\sigma^2_\calT + d \sigma^2_\calP) + \ln \E_{\calT} \E_{\calP} \exp \bigg( - \frac{\lambda}{2\sigma^2} \sigma_\bx^2 ||\bw^* - \mu_P||^2 \bigg)  \\
=& 
\frac{\lambda}{2\sigma^2}   \sigma_\bx^2( ||\mu_\calT||^2 + d\sigma^2_\calT + d \sigma^2_\calP) + \ln \left(1 + \frac{\lambda}{\sigma^2} \sigma^2_\bx \sigma^2_\calP \right)^{-\frac{d}{2}} \E_{\calT} \exp \bigg( -\frac{ \frac{\lambda}{2\sigma^2} \sigma_\bx^2 ||\bw^*||^2}{1 + \frac{\lambda}{\sigma^2} \sigma_\bx^2 \sigma^2_\calP} \bigg)  \\
=& 
\frac{\lambda}{2\sigma^2}   \sigma_\bx^2( ||\mu_\calT||^2 + d\sigma^2_\calT + d \sigma^2_\calP) -\frac{d}{2} \ln \left(1 + \lambda \frac{\sigma_\bx^2}{\sigma^2}   (\sigma^2_\calP +  \sigma^2_\calT) \right)  -\frac{ \frac{\lambda}{2\sigma^2} \sigma_\bx^2 ||\mu_\calT||^2}{1 + \lambda\frac{\sigma_\bx^2}{\sigma^2}  (\sigma^2_\calP + \sigma_\calT^2)}  \\
\leq&
\frac{\lambda}{2\sigma^2}   \sigma_\bx^2( ||\mu_\calT||^2 + d\sigma^2_\calT + d \sigma^2_\calP) - \frac{d \frac{\lambda}{2\sigma^2} \sigma^2_\bx (\sigma^2_\calP +  \sigma^2_\calT)}{1 +  \lambda\frac{\sigma_\bx^2}{\sigma^2}   (\sigma^2_\calP +  \sigma^2_\calT)}  -\frac{ \frac{\lambda}{2\sigma^2} \sigma_\bx^2 ||\mu_\calT||^2}{1 + \lambda\frac{\sigma_\bx^2}{\sigma^2}   (\sigma^2_\calP + \sigma_\calT^2)}  \\
=& 
\frac{(\frac{\lambda}{2\sigma^2}   \sigma_\bx^2 ||\mu_\calT||^2 + \lambda \frac{d}{2}  c_{\rom{2}}) (1 + \lambda c_{\rom{2}}) -  \lambda \frac{d}{2} c_{\rom{2}} - \frac{\lambda}{2\sigma^2} \sigma_\bx^2 ||\mu_\calT||^2}{1 + \lambda c_{\rom{2}}} \\
=&
\frac{\lambda^2 (  \frac{\sigma_\bx^2}{\sigma^2} c_{\rom{2}} ||\mu_\calT||^2 + d c_{\rom{2}}^2) }{2(1 + \lambda c_{\rom{2}})}  =
\frac{\lambda^2 s_{\rom{2}}^2}{2(1 + \lambda c_{\rom{2}})} 
\end{align*}

with $c_{\rom{2}} = \frac{\sigma_\bx^2}{\sigma^2} (\sigma^2_\calP + \sigma_\calT^2)$ and $s_{\rom{2}}^2= \frac{\sigma_\bx^2}{\sigma^2} c_{\rom{2}} ||\mu_\calT||^2 + d c_{\rom{2}}^2$.

%% file: content/supplement_method.tex
\section{PACOH-GP algorithm details} \label{appendix:pacoh}

\subsection{Meta-training with PACOH-GP}

\paragraph{Prior parametrization.} When meta-learning a GP prior, we instantiate the GP's mean $m_\phi$ and kernel function $k_\phi$ as neural networks, where the parameter vector $\phi$ can be meta-learned. To ensure the positive-definiteness of the kernel, we use the neural network as feature map $\varphi_\phi(x)$ on top of which we apply a squared exponential (SE) kernel. Accordingly, the parametric kernel reads as $k_\phi(x, x') = \frac{1}{2}\exp \left( - ||\varphi_\phi(x) - \varphi_\phi(x')||_2^2 \right)$.
Both $m_\phi(x)$ and $\varphi_\phi(x)$ are fully-connected neural networks with 4 layers with each 32 neurons and $\tanh$ non-linearities. The parameter vector $\phi$ represents the weights and biases of both neural networks.
As hyper-prior we choose a zero-mean isotropic Gaussian $\calP(\phi) = \calN(0, \sigma_{\calP}^2 I)$.

\paragraph{Estimating the hyper-posterior score.}
To estimate $\nabla_\phi \calQ^*(\phi)$, we use mini-batching on the task level. In each iteration, we sample a mini-batch of $H \leq n$ datasets $S_1, ..., S_H$ and form an unbiased estimate of the hyper-posterior score as follows:
\begin{equation}
\tilde{\nabla}_{\phi} \log \calQ^*(\phi) = \frac{n}{H} \cdot \sum_{h=1}^H \frac{1}{m_h + 1} \nabla_{\phi} \log Z(S_h, P_{\phi})  + \nabla_{\phi} \log \calP(\phi) \;.
\end{equation}
Here, $\nabla_{\phi} \log Z(S_h, P_{\phi})$ is the derivative of the closed-form marginal log-likelihood (see Eq. \ref{eq:mll_gp}) w.r.t.\ the prior parameters $\phi$.

\paragraph{MAP approximation.} A maximum a-posteriori (MAP) approximation of  $\calQ^*$ is the simplest way to obtain a practical meta-learning algorithm from our PAC-Bayesian theory. In particular, it approximates the $\calQ^*(\phi)$ by a Dirac measure $\delta_\phi(\phi^*)$ on the prior parameter vector $\phi^*$ that maximizes $\calQ^*$, i.e., $\phi^* = \argmax_\phi \calQ^*(\phi)$. To find $\phi^*$, we initially randomly initialize $\phi$, and then optimize $\phi$ by performing gradient descent on $\tilde{\nabla}_{\phi} \log \calQ^*(\phi)$.

\paragraph{SVGD approximation.}
\looseness -1 SVGD \citep{Liu2016} approximates $\calQ^*$ as a set of particles $\hat{\calQ} = \{P_1, ..., P_K\}$. In our setup, each particle corresponds to the parameters of the GP prior, i.e., $\hat{\calQ} =  \{\phi_1, ..., \phi_K\}$. Initially, we sample random priors $\phi_k \sim \calP$ from our hyper-prior. Then, the SVGD iteratively transports the set of particles to match $\calQ^*$, by applying a form of functional gradient descent that minimizes  $D_{KL}(\hat{\calQ} | \calQ^*)$ in the reproducing kernel Hilbert space induced by $k(\cdot,\cdot)$. We choose a squared exponential kernel with length scale (hyper-)parameter $\ell$, i.e., $k(\phi ,\phi') = \ \exp \left( - \frac{|| \phi - \phi'||_2^2}{2 \ell} \right)$. In each iteration, the particles are updated by
\begin{equation*}
\phi_k \leftarrow \phi_k + \eta_t \psi^*(\phi_k) ~, \quad \text{with} \quad \psi^*(\phi) = \frac{1}{K} \sum_{l=1}^K \left [k(\phi_l, \phi) \nabla_{\phi_l} \log \calQ^*(\phi_l) +  \nabla_{\phi_l} k(\phi_l, \phi) \right] \;.
\end{equation*}
\textbf{VI approximation. }
When aiming to approximate $\calQ^*$ via variational inference, we consider the variational family of Gaussians with diagonal covariance matrices over our prior parameters $\phi$. In particular, we have variational hyper-posteriors of the form
$$\tilde{\calQ}_\upsilon(\phi) = \calN(\phi; \mu_\calQ, \sigma^2_\calQ), ~ \text{with} ~ \upsilon = (\mu_\calQ, \log \sigma_\calQ)$$ 
where we parameterize the variance of $\tilde{\calQ}$ in the log-space to avoid a positivity constraint. The resulting VI loss follows as

\begin{equation} \label{eq:negative_elbo}
J^{\text{VI}}(\upsilon) = - \E_{\phi \sim \calQ_\upsilon} \left [  \frac{\tilde{m}}{\tilde{m} + 1} \sum_{i=1}^n \frac{1}{m_i} \log Z(S_i, P_\phi)  + \log \calP(\phi) - \log \calQ_\upsilon(\phi)\right] \;.
\end{equation}
\looseness -1 Here, $\tilde{m}$ is the harmonic mean of dataset sizes $m_1, ..., m_n$.
To estimate the gradients of $J^{\text{VI}}(\upsilon)$ w.r.t. $\upsilon$, we employ a pathwise gradient estimator, also known as reparametrization trick. That is, we sample a set of $K$ prior parameters $\phi_k := \mu_{\calQ} + \sigma_{\calQ} \epsilon_k, ~ \epsilon_k \sim \calN(0,I)$ as well as a mini-batch of $H$ datasets $S_1, ..., S_H$ and compute an unbiased gradient estimate of (\ref{eq:negative_elbo}) as follows:
\begin{equation} \label{eq:negative_elbo_grad}
\nabla_\upsilon J^{\text{VI}}(\upsilon) \approx - \frac{1}{K} \sum_{k=1}^L \nabla_{\mu_{\calQ}, \sigma_{\calQ}} \left( \frac{n}{H} \cdot \frac{\tilde{m}}{\tilde{m} + 1} \sum_{h=1}^H\frac{1}{m_h} \log Z(S_h, P_{\phi_k})  + \log \calP(\phi_k) - \log \calQ_\upsilon(\phi_k)\right) \;.
\end{equation}
During gradient descent with $\nabla_\upsilon J^{\text{VI}}(\upsilon)$, we employ the adaptive learning rate method Adam. Due to the double stochasticity (mini-batches of tasks and mini-batches of $\phi_k \sim \calQ_\upsilon$), we found that in practice the gradient estimates of the marginal log-likelihood term in (\ref{eq:negative_elbo_grad}) are very noisy whereas the second and third term (meta-level KL-divergence) are subject to less variance. As a result, the less noisy gradients of the KL-divergence dominate during gradient-descent, pushing the VI posterior towards the prior which in turn leads to a higher entropy of $\calQ_\upsilon$ and even noisier gradient estimates for the marginal log-likelihood term. To counteract this explosion in hyper-posterior entropy, we add a weight $0 < \eta < 1$ in front of $ \log \calP(\phi) - \log \calQ_\upsilon(\upsilon)$ which effectively down-scales the effect of $D_{KL}(\calQ_\phi ||\calP)$ and improves results significantly. Such tempering of the prior often help when the Gaussian prior (here hyper-prior) is misspecified and has been studied in e.g. \citet{fortuin2021bayesian}.
\subsection{Meta-Testing / target training with PACOH-GP}
\looseness -1 Meta-learning with PACOH gives us an approximation of $\calQ^*$. In target-testing (see Figure \ref{fig:overview}), the base learner is instantiated with the meta-learned prior $P_\phi$, receives a dataset $\tilde{S}=(\tilde{\bX}, \tilde{\by})$ from an unseen task $\calD \sim \calT$ and outputs a posterior $Q$ as product of its inference. In our GP setup, $Q$ is the GP posterior and the predictive distribution $\hat{p}(y^*|x^*, \tilde{\bX}, \tilde{\by}, \phi)$ is a Gaussian \citep[for details, see][]{rasmussen2003gaussian}.
Since the meta-learner outputs $\calQ$, a distribution over priors, we may obtain different predictions for different priors $P_\phi \sim \calQ$, sampled from the hyper-posterior. To obtain a predictive distribution we empirically marginalize $\calQ$. That is, we draw a set of prior parameters $\phi_1, ..., \phi_K \sim \calQ$ from the hyper-posterior, compute their respective predictive distributions $\hat{p}(y^*|x^*, \tilde{\bX}, \tilde{\by}, \phi_k)$ and form an equally weighted mixture:
\begin{equation} \label{eq:predictive_mixture}
    \hat{p}(y^*|x^*, \tilde{\bX}, \tilde{\by}, \calQ) = \E_{\phi \sim \calQ} \left[ \hat{p}(y^*|x^*, \tilde{\bX}, \tilde{\by}, \phi) \right] \approx \frac{1}{K} \sum_{k=0}^{K} \hat{p}(y^*|x^*, \tilde{\bX}, \tilde{\by}, \phi_k) ~, \quad \phi_k \sim \calQ 
\end{equation}
Since we are concerned with GPs, (\ref{eq:predictive_mixture}) coincides with a mixture of Gaussians. As one would expect, the mean prediction under $\calQ$ (i.e., the expectation of (\ref{eq:predictive_mixture})), is the average of the mean predictions corresponding to the sampled prior parameters $\phi_1, ..., \phi_K$. In case of PACOH-GP-VI, we sample $K=100$ priors from the variational hyper-posterior $\tilde{\calQ}$. For PACOH-GP-SVGD, samples from the hyper-posterior correspond to the $K=10$ particles. PACOH-GP-MAP can be viewed as a special case of SVGD with $K=1$, that is, only one particle. Thus, $\hat{p}(y^*|x^*, \tilde{\bX}, \tilde{\by}, \calQ) \approx \hat{p}(y^*|x^*, \tilde{\bX}, \tilde{\by}, \phi^{MAP})$ is a single Gaussian.

\section{PACOH-NN algorithm details} \label{appendix:pacoh_nn}

\looseness -1 Here, we summarize and further discuss our proposed meta-learning algorithm \emph{PACOH-NN}. An overview of our proposed framework is illustrated in Figure \ref{fig:overview}. Overall, it consists of the two stages \emph{meta-training} and \emph{meta-testing}, which we explain in more details in the following.
\subsection{Meta-training with PACOH-NN} The hyper-posterior distribution $\calQ$ that minimizes the upper bound on the transfer error is given by
$
   \calQ^*(P) \propto \calP(P) \exp\left( \sum_{i=1}^n  \frac{\lambda}{n\beta_i + \lambda} \log \tilde{Z}(S_i, P)  \right) \vspaceequation
$.
Here, we no longer assume that $m = m_i ~ \forall i=1,...,n$, which was done in the theory to maintain notational brevity. Thus, we use a different $\beta_i$ for each task as we want to set $\beta_i = m_i$ or $\beta_i = \sqrt{m_i}$.
Provided with a set of datasets $S_1, ..., S_n$, the meta-learner minimizes the respective meta-objective, in the case of \emph{PACOH-NN-SVGD}, by performing SVGD on the $\calQ^*$. \emph{PACOH-NN-MAP} can be considered a special case of the SVGD-based approximation of $\calQ^*$ with only one particle, i.e., $K=1$. Algorithm \ref{algo:pacoh_nn} outlines the required steps in more detail.
%
%
%
%
%
Alternatively, to estimate the score of $\nabla_{\phi_k} \tilde{\calQ}^*(\phi_k)$, we can use mini-batching at both the task and the dataset level. Specifically, for a given meta-batch size of $n_{bs}$ and a batch size of $m_{bs}$, we get Algorithm \ref{algo:pacoh_nn_batched}.

\begin{algorithm}
\caption{PACOH-NN-SVGD: mini-batched meta-training}
\label{algo:pacoh_nn_batched}
\begin{algorithmic}
\STATE \textbf{Input:} hyper-prior $\calP$, datasets $S_1, ..., S_n$
\STATE \textbf{Input:} kernel function $k(\cdot, \cdot)$, SVGD step size $\eta$, number of particles $K$
\STATE $\{ \phi_1, ..., \phi_K\} \sim \calP$ \hfill // Initialize prior particles
\WHILE{not converged} 
    \STATE $\{T_1, ..., T_{n_{bs}}\} \subseteq [n]$ \hfill // sample $n_{bs}$ tasks uniformly at random
 	\FOR{$i=1, ..., n_{bs}$}
 	    \STATE $\tilde{S}_i \leftarrow \{z_1, ..., z_{m_{bs}}\} \subseteq S_{T_i}$ \hfill // sample $m_{bs}$ datapoints from $S_{T_i}$ uniformly at random
 	\ENDFOR
 	
 	\FOR{$k=1,...,K$} 
     	\STATE $\{\theta_1, ..., \theta_L\} \sim P_{\phi_k}$ \hfill // sample NN-parameters from prior
     	\FOR{$i=1, ..., n_{bs}$}
     	    \STATE $\log \tilde{Z}(\tilde{S}_i, P_{\phi_k}) \leftarrow \text{LSE}_{l=1}^L\left( - \beta_i  \hat{\calL}(\theta_l, \tilde{S}_i) \right) - \log L$ \hfill // estimate generalized MLL
     	\ENDFOR
     	\STATE $  \nabla_{\phi_k} \log \tilde{\calQ}^*(\phi_k)  \leftarrow \nabla_{\phi_k} \log \calP(\phi_k) + \frac{n}{n_{bs}}\sum_{i=1}^{n_{bs}} \frac{\lambda}{n\beta_i + \lambda} \nabla_{\phi_k} \log \tilde{Z}(S_i, P_{\phi_k})$ \hfill // compute score
	\ENDFOR
	
	\STATE $\phi_k \leftarrow \phi_k + \frac{\eta}{K} \sum_{k'=1}^K \left [k(\phi_{k'}, \phi_k) \nabla_{\phi_{k'}} \log \tilde{\calQ}^*(\phi_{k'}) +  \nabla_{\phi_{k'}} k(\phi_{k'}, \phi_k) \right] \; \forall{k \in [K]}$ \hfill // SVGD
\ENDWHILE
\STATE \textbf{Output:} set of priors $\{ P_{\phi_1}, ..., P_{\phi_K} \}$
\end{algorithmic}
\end{algorithm}

\subsection{Meta-testing / target training with PACOH-NN} \label{appx:meta_testing}
\looseness -1 The result of meta-training with {\em PACOH-NN} is a set of neural network priors $\{ P_{\phi_1}, ..., P_{\phi_K} \}$. To understand how good these meta-learned priors are, we need to instantiate our base learner with these priors and evaluate its performance on an unseen learning task $\tau = (\calD, m) \sim \calT$ when given a corresponding training dataset $\tilde{S} \sim \calD^m$. In case of neural networks, our base learner forms a generalized Bayesian posterior $Q^*(S,P_\phi)$ over neural networks parameters $\phi$. Since this $Q^*(S,P_\phi)$ is intractable for neural networks, we employ SVGD to approximate it---a standard procedure in the context of Bayesian Neural Networks. 
Algorithm \ref{algo:pacoh_nn_target_training} details the steps of the approximating procedure---referred to as \emph{target training}---when performed via SVGD. For a data point $x^*$, the respective predictor outputs a probability distribution given as $\tilde{p}(y^*| x^*, \tilde{S}) \leftarrow \frac{1}{K \cdot L}\sum_{k=1}^K \sum_{l=1}^L p(y^*|h_{\theta^k_l}(x^*))$. We evaluate the quality of the predictions on a held-out test dataset $\tilde{S}^* \sim \calD$ from the same task, in a \emph{target testing} phase (see Appendix \ref{appendix:exp_methodology}).

\begin{algorithm}
\caption{PACOH-NN: meta-testing}
\label{algo:pacoh_nn_target_training}
\begin{algorithmic}
\STATE \textbf{Input:} set of priors $\{ P_{\phi_1}, ..., P_{\phi_K} \}$, target training dataset $\tilde{S}$
\STATE \textbf{Input:} kernel function $k(\cdot, \cdot)$, SVGD step size $\nu$, number of particles $L$
\FOR{$k=1, ..., K$}
 	\STATE $\{\theta^k_1, ..., \theta^k_L\} \sim P_{\phi_k}$ \hfill // initialize NN posterior particles from $k$-th prior
 	\WHILE{not converged} 
 		\FOR{$l=1,...,L$}
 		\STATE $  \nabla_{\theta^k_l} Q^*(\theta^k_l))  \leftarrow \nabla_{\theta^k_l} \log P_{\phi_k}(\theta^k_l)) + \beta ~ \nabla_{\theta^k_l}  \calL(l, \tilde{S})$ \hfill // compute score
 		\ENDFOR
 		\STATE $\theta^k_l \leftarrow \theta^k_l +  \frac{\nu}{L} \sum_{l'=1}^L \left [k(\theta^k_{l'}, \theta^k_l) \nabla_{\theta^k_{l'}} \log Q^{*}(\theta^k_{l'}) +  \nabla_{\theta^k_{l'}} k(\theta^k_{l'}, \theta^k_l) \right] \forall{l \in [L]}$ \hfill // update
 	\ENDWHILE
\ENDFOR
\STATE \textbf{Output:} a set of NN parameters
$\bigcup_{k=1}^K \{\theta_1^k\, ..., \theta_L^k\}$
\end{algorithmic}
\end{algorithm}

\subsection{Properties of the score estimator} \label{appendix:properties_mll_estimator}

Since the marginal log-likelihood of BNNs is intractable, we have replaced it by a numerically stable Monte Carlo estimator $\log \tilde{Z}_{\beta}(S_i, P_\phi)$ in (\ref{eq:mll_estimator}), in particular
\begin{equation}
    \log \tilde{Z}_\beta(S_i, P_\phi) := \log  \frac{1}{L} \sum_{l=1}^{L} e^{- \beta  \hat{\calL}(\theta_l,S_i)} = \text{LSE}_{l=1}^L\left( - \beta \hat{\calL}(\theta_l, S_i) \right) - \log L ~, ~~ \theta_l \sim P_\phi ~.
\end{equation}
\looseness -1 Since the Monte Carlo estimator involves approximating an expectation of an exponential, it is not unbiased. However, we can show that replacing $\log Z_{\beta}(S_i, P_\phi)$ by the estimator $\log \tilde{Z}_{\beta}(S_i, P_\phi)$, we still minimize a valid upper bound on the transfer error (see Proposition \ref{proposition:mll_estimate_still_upper_bound}).

\begin{proposition} \label{proposition:mll_estimate_still_upper_bound_app}
In expectation, replacing $\log Z_{\beta}(S_i, P_\phi)$ in (\ref{eq:meta-level_pac_bound_with_mll}) by the Monte Carlo estimate $\log \tilde{Z}_{\beta}(S_i, P) := \log  \frac{1}{L} \sum_{l=1}^{L} e^{- \beta \hat{\calL}(\theta_l,S_i)}, ~ \theta_l \sim P$ still yields a valid upper bound of the transfer error. In particular, it holds that
\begin{align}
\calL(\calQ, \calT) & \leq   - \frac{1}{n} \sum_{i=1}^n \frac{1}{\beta} \E_{P \sim \calQ} \left[\log Z(S_i, P) \right]   + \left(\frac{1}{\lambda} + \frac{1}{n \beta}\right)  D_{KL}(\calQ||\calP) + C  \hspace{-4pt} \label{eq:normal_pac_bound} \\
\begin{split} \label{eq:estimator_upper_bound}
& \leq - \frac{1}{n} \sum_{i=1}^n \frac{1}{\beta} \E_{P \sim \calQ} \left[ \E_{\theta_1,...,\theta_L \sim P} \left[ \log \tilde{Z}(S_i, P)  \right] \right]  + \left(\frac{1}{\lambda} + \frac{1}{n \beta}\right) D_{KL}(\calQ||\calP) + C .
\end{split}
\end{align}
\end{proposition}
\begin{proof}
First, we show that:
\begin{align}
\label{eq:ineq_mll_estimator}
\E_{\theta_1,...,\theta_L \sim P} \left[\log \tilde{Z}_{\beta}(S_i, P) \right] &= \E_{\theta_1,...,\theta_L \sim P} \left[ \log \frac{1}{L} \sum_{l=1}^{L}  e^{- \beta \hat{\calL}(\theta_l,S_i)} \right] \nonumber \\
&\leq \log \frac{1}{L} \sum_{l=1}^{L} \E_{\theta_l \sim P} \left[  e^{- \beta \hat{\calL}(\theta_l,S_i)} \right] = \log Z_\beta(S_i, P)
\end{align}
which follows directly from Jensen's inequality and the concavity of the logarithm. Now, Proposition \ref{proposition:mll_estimate_still_upper_bound_app} follows directly from (\ref{eq:ineq_mll_estimator}).
\end{proof}

\looseness -1 By the law of large numbers, it is straightforward to show that as $L \rightarrow \infty$, the $\log \tilde{Z}(S_i, P) \xrightarrow[]{\text{a.s.}} \log Z(S_i, P) $, i.e., the estimator becomes asymptotically unbiased and we recover the original PAC-Bayesian bound (i.e., (\ref{eq:estimator_upper_bound})  $\xrightarrow[]{\text{a.s.}}$ (\ref{eq:normal_pac_bound})). Also, it is noteworthy that the bound in (\ref{eq:estimator_upper_bound}) we get by our estimator is, in expectation, tighter than the upper bound when using the na\"ive estimator 
$
\log \hat{Z}_\beta(S_i, P) := - \beta ~ \frac{1}{L} \sum_{l=1}^{L}   \hat{\calL}(\theta_l,S_i) \quad \theta_l \sim P_\phi
$
which can be obtained by applying Jensen's inequality to $\log \E_{\theta \sim P_\phi} \left[ e^{- \beta \hat{\calL}(\theta,S_i)} \right]$. In the edge case $L=1$, our LSE estimator $\log \tilde{Z}_\beta(S_i, P)$ falls back to this na\"ive estimator and coincides in expectation with $\E [ \log \hat{Z}_\beta(S_i, P)] = - \beta ~ \E_{\theta \sim P}   \hat{\calL}(\theta_l,S_i)$. As a result, we effectively minimize the looser upper bound
\begin{align*}
\calL(\calQ, \calT) & \leq   \frac{1}{n} \sum_{i=1}^n \E_{\theta \sim P}  \left[  \hat{\calL}(\theta,S_i) \right]  + \left(\frac{1}{\lambda} + \frac{1}{n \beta}\right)   D_{KL}(\calQ||\calP) + C(\delta, n, \overline{m}) . \\
& =  \E_{\theta \sim P}  \left[  \frac{1}{n} \sum_{i=1}^n \frac{1}{m_i} \sum_{j=1}^{m_i} - \log p(y_{ij} | x_{ij}, \theta) \right]  + \left(\frac{1}{\lambda} + \frac{1}{n \beta}\right)   D_{KL}(\calQ||\calP) + C(\delta, n, \overline{m}) 
\end{align*}
Since the boundaries between the tasks vanish in the edge case of $L=1$, i.e., all data-points are treated as if they would belong to one dataset, we should choose $L > 1$. In our experiments, we used $L=5$ and found the corresponding approximation to be sufficient.
%
%

%% file: content/supplement_exps.tex
\section{Experiments}\label{appendix:exps}
\subsection{Meta-Learning Environments}
\label{appendix:meta-envs}
\looseness -1 We provide further details on the meta-learning environments used in Section~\ref{sec:experiments}.
Information about the numbers of tasks and samples per environments can be found in Table~\ref{tab:num_tasks_samples}.
\paragraph{Sinusoids.}
\begin{table}[htbp]
\centering
\begin{tabular}{l|ccccc}
 & Sinusoid & Cauchy & SwissFEL & Physionet & Berkeley \\ \hline
 $n$ & 20 & 20 & 5 & 100 & 36 \\
 $m_i$ & 5 & 20 & 200 & 4 - 24  & 288 \\ 
\end{tabular}
\caption{\looseness -1 Number of tasks $n$ and samples per task $m_i$ for the meta-learning environments.}
\label{tab:num_tasks_samples}
\end{table}
Each task of the sinusoid environment corresponds to a parametric function
\begin{equation}
   f_{a, b, c, \beta} (x) = \beta \cdot x + a \cdot \sin(1.5 \cdot (x - b)) + c \;,
\end{equation}
which, in essence, consists of an affine as well as a sinusoid function. Tasks differ in the function parameters $(a, b, c, \beta)$ that are sampled from the task environment $\calT$ as follows:
\begin{equation}
a  \sim \calU(0.7, 1.3), \quad  b  \sim \calN(0, 0.1^2), \quad  c  \sim \calN(5.0, 0.1^2), \quad \beta  \sim \calN(0.5, 0.2^2) \;. \label{eq:param_sampling_sinusoid}
\end{equation}
\begin{figure}[t] \label{fig:simulated_tasks}
\begin{subfigure}{0.5\textwidth}
        \centering
        \includegraphics[width=0.85\textwidth]{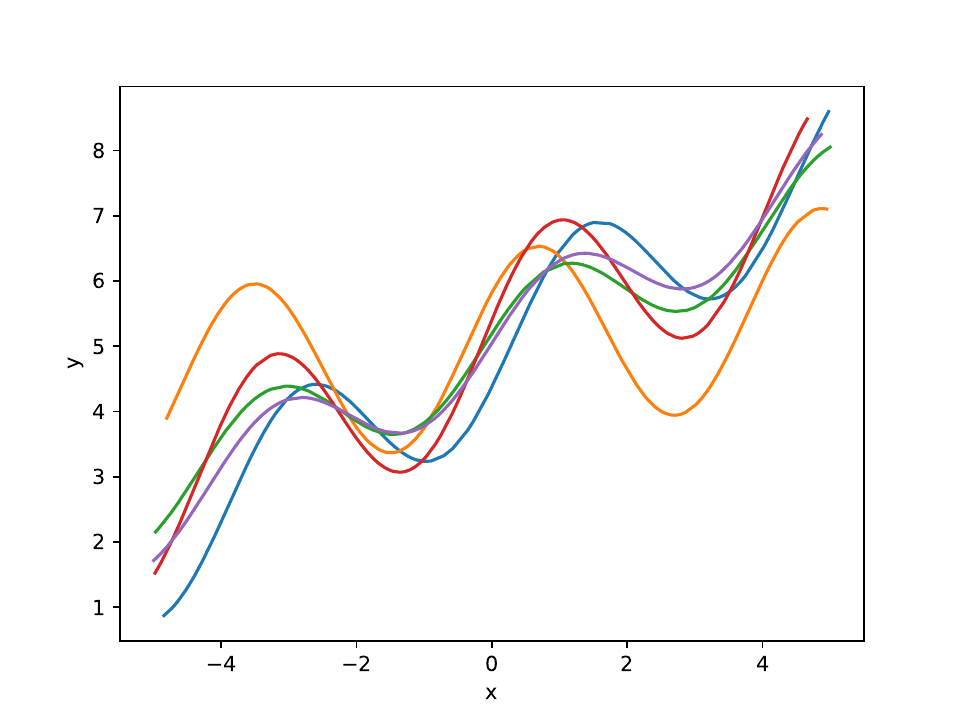}
        \vspace{-6pt}
        \caption{Sinusoid tasks \vspace{-6pt}} \label{fig:sin_tasks}
    \end{subfigure}%
    ~
 \begin{subfigure}{0.5\textwidth}
        \centering
        \includegraphics[width=0.85\textwidth]{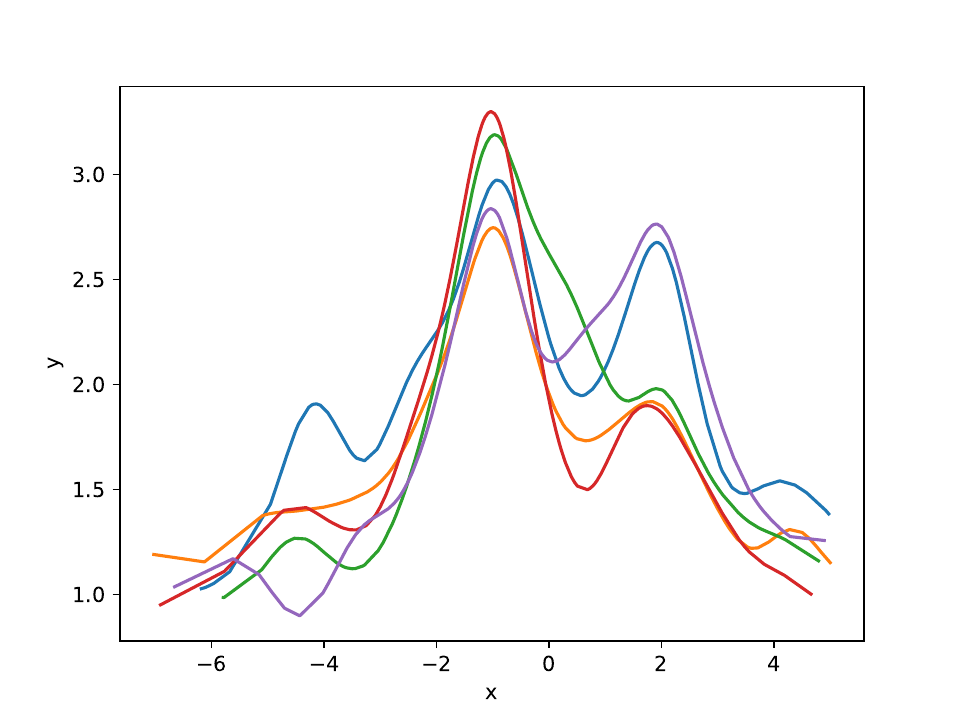}
        \vspace{-6pt}
        \caption{Cauchy tasks \vspace{-6pt}} \label{fig:cauchy_tasks}
    \end{subfigure}%
  \caption{Depiction of tasks (i.e., functions) sampled from the Sinusoid and Cauchy task environment, respectively. Note that the Cauchy task environment is two-dimensional ($\text{dim}(\calX) = 2$), while (b) displays a one-dimensional projection.}
\end{figure}
Figure~\ref{fig:sin_tasks} depicts functions $f_{a, b, c, \beta}$ with parameters sampled according to (\ref{eq:param_sampling_sinusoid}). To draw training samples from each task, we draw $x$ uniformly from $\calU(-5, 5)$ and add Gaussian noise with standard deviation $0.1$ to the function values $f(x)$:
\begin{equation}
x \sim \calU(-5, 5)~, \qquad y \sim \calN(f_{a, b, c, \beta} (x), 0.1^2) \;.
\end{equation}

\paragraph{Cauchy.}
Each task of the Cauchy environment can be interpreted as a two-dimensional mixture of Cauchy distributions plus a function sampled from a Gaussian process prior with zero mean and SE kernel function $k(x, x') = \exp\left( \frac{||x - x'||_2^2}{2l} \right)$ with $l=0.2$. The (unnormalized) mixture of Cauchy densities is defined as:
\begin{equation}
m(x) = \frac{6}{\pi \cdot (1 + ||x-\mu_1||^2_2)}  + \frac{3}{\pi \cdot (1 + ||x-\mu_2||^2_2)} \;,
\end{equation}
with $\mu_1 = (-1, -1)^\top$ and $\mu_2 = (2, 2)^\top$. 
Functions from the environment are sampled as
\begin{equation} \label{eq:Cauchy_sampling}
f(x) = m(x) + g(x) ~, \qquad g \sim \mathcal{GP}(0, k(x,x')) \;.
\end{equation}

Figure~\ref{fig:cauchy_tasks} depicts a one-dimensional projection of functions sampled according to (\ref{eq:Cauchy_sampling}). To draw training samples from each task, we draw $x$ from a truncated normal distribution and add Gaussian noise with standard deviation $0.05$ to the function values $f(x)$:  \begin{equation}
x:= \min\{\max\{\tilde{x}, 2\}, -3\} ~, ~~ \tilde{x} \sim \calN(0, 2.5^2) ~, \qquad  y \sim \calN(f(x), 0.05^2) \;.
\end{equation}

\paragraph{SwissFEL.}

\looseness -1 Free-electron lasers (FELs) accelerate electrons to generate pulsed laser beams in the X-ray spectrum. They can be used to map nanometer-scale structures, e.g., in molecular biology and material science. The accelerator and the electron beam line of a FEL consist of multiple undulators whose parameters can be adjusted to maximize the pulse energy \citep{kirschner2019linebo}. Due to different operational modes, parameter drift, and changing (latent) conditions, the laser's pulse energy function changes across time. Hence, optimizing the laser's parameters is a recurrent task.

\looseness -1 Overall, our meta-learning environment consists of different parameter optimization runs (i.e., tasks) on the SwissFEL \citep{milne2017swissfel}. The input space, corresponding to the laser's parameters, has 12 dimensions. The scalar regression target is the pulse energy. For details on the individual parameters, we refer to \citet{kirschner2019swissfel}. For each run, we have around 2000 data points. Since these data-points are generated with online optimization methods, the data are non-i.i.d.\ and get successively less diverse throughout the optimization. Hence, we only take the first 400 data points per run and split them into training and test subsets of size 200. Overall, we have 9 runs (tasks) available. 5 of those runs are used for meta-training and the remaining 4 runs are used for meta-testing. 

\paragraph{PhysioNet.}

The 2012 Physionet competition \citep{silva2012predicting} published a dataset of patient stays on the intensive care unit (ICU).
Each patient stay consists of a time series over 48 hours, where up to 37 clinical variables are measured.
The original task in the competition was binary classification of patient mortality, but, due to the large number of missing values (ca. 80~\% of features), the dataset is also popular as a test bed for time series prediction.
We treat each patient as a separate task and the different clinical variables as different environments.
We use the Glasgow coma scale (GCS) and hematocrit value (HCT) as environments for our study, since they are among the most frequently measured variables in this dataset. From the dataset, we remove all patients where less than four measurements of CGS (and HCT respectively) are available. From the remaining patients we use 100 patients for meta-training and 500 patients each for meta-validation and meta-testing. Here, each patient corresponds to a task. Since the number of available measurements differs across patients, the number of training points $m_i$ ranges between 4 and 24.

\paragraph{Berkeley-Sensor.}

We use data from 46 sensors deployed in different locations at the Intel Research lab in  Berkeley \citep{intel_sensor_data}. The dataset contains 4 days of data, sampled at 10 minute intervals. Each task corresponds to one of the 46 sensors and requires auto-regressive prediction, in particular, predicting the next temperature measurement given the last 10 measurement values. 36 sensors (tasks) with data for the first two days are used for meta-training, whereas the remaining 10 sensors with data for the last two days are employed for meta-testing. Note that we separate meta-training and -testing data both temporally and spatially, since the data is non-i.i.d. For the meta-testing, we use the 3rd day as context data, i.e., for target training, and the remaining data for target testing.

\subsection{Experimental Methodology}
\label{appendix:exp_methodology}

In the following, we describe our experimental methodology and provide details on how the empirical results reported in Section~\ref{sec:experiments} were generated. 
Overall, evaluating a meta-learner consists of two phases, {\em meta-training} and {\em meta-testing}, outlined in Appendix \ref{appendix:pacoh_nn}. The latter can be further sub-divided into {\em target training} and {\em target testing}. Figure~\ref{fig:overview} illustrates these different stages for our PAC-Bayesian meta-learning framework.



\looseness -1 The outcome of the training procedure is an approximation of the generalized Bayesian posterior $Q^*(S,P)$ (see Appendix \ref{appendix:pacoh_nn}), pertaining to an unseen task $\tau = (\calD, m) \sim \calT$ from which we observe a dataset $\tilde{S} \sim \calD^m$. In {\em target-testing}, we evaluate its predictions on a held-out test dataset $\tilde{S}^* \sim \calD$ from the same task. For PACOH-NN, NPs, and MLAP, the respective predictor outputs a probability distribution $\hat{p}(y^*|x^*, \tilde{S})$ for the $x^*$ in $\tilde{S}^*$. The respective mean prediction corresponds to the expectation of $\hat{p}$, that is $\hat{y} = \hat{\E}(y^*|x^*, \tilde{S})$. In the case of MAML, only a mean prediction is available. Based on the mean predictions, we compute the {\em root mean-squared error (RMSE)}
and the {\em calibration error} (see Appendix \ref{appendix:calib_error}). Rather than reporting the test log-likelihood, this allows us to measure the quality of mean predictions and the quality of uncertainty estimates separately.
The meta-training and meta-testing procedure is repeated for five random seeds that influence both the initialization and gradient-estimates of the algorithms. The reported averages and standard deviations are based on the results obtained for different seeds.

\subsubsection{Calibration Error} \label{appendix:calib_error}
The concept of calibration applies to probabilistic predictors that, given a new target input $x_i$, produce a probability distribution $\hat{p}(y_i|x_i)$ over predicted target values $y_i$. 

\textbf{Calibration error for regression.}
Corresponding to the predictive density, we denote a predictor's cumulative density function (CDF) as $\hat{F}(y_j|x_j) = \int_{-\infty}^{y_j} \hat{p}(y|x_i) \, dy$. For confidence levels $0 \leq q_h < ... < q_H \leq 1$, we can compute the corresponding empirical frequency
\begin{equation}
\hat{q}_h = \frac{ | \{ y_j ~ | ~ \hat{F}(y_j|x_j) \leq q_h, j=1, ..., m \} | }{m} \;,
\end{equation} 
based on dataset $S=\{(x_i, y_i)\}_{i=1}^m$ of $m$ samples. If we have calibrated predictions we would expect that $\hat{q}_h \rightarrow q_h$ as $m \rightarrow \infty$. Similar to \citep{Kuleshov2018}, we can define the calibration error as a function of residuals $\hat{q}_h - q_h$, in particular,
\begin{equation} \label{eq:calib_err}
\text{calib-err} = \frac{1}{H} \sum_{h=1}^H |\hat{q}_h - q_h| \;.
\end{equation}
Note that while \citet{Kuleshov2018} report the average of squared residuals $|\hat{q}_h - q_h|^2$, we report the average of absolute residuals $|\hat{q}_h - q_h|$ in order to preserve the units and keep the calibration error easier to interpret. In our experiments, we compute (\ref{eq:calib_err}) with $M=20$ equally spaced confidence levels between 0 and 1.

\textbf{Calibration error for classification.} Our classifiers output a categorical probability distribution $\hat{p}(y=k|x)$ for $k=1,..., C$ where $\calY = \{1, ..., C\}$ with $C$ denoting the number of classes. The prediction of the classifier is the most probable class label, i.e., $\hat{y}_j = \argmax_{k} \hat{p}(y_j=k|x_j)$. Correspondingly, we denote the classifiers confidence in the prediction for the input $x_j$ as $\hat{p}_j := \hat{p}(y_j=\hat{y}_j|x_j)$. Following the calibration error definition of \citet{guo2017calibration}, we group the predictions into $H=20$ interval bins of size $1/H$ depending on their prediction confidence. In particular, let $B_h = \{j ~|~ p_j \in \left(\frac{h-1}{H}, \frac{h}{H} \right]\} $ be the set of indices of test points $\{(x_j,y_j)\}_{j=1}^m$ whose prediction fall into the interval $\left(\frac{h-1}{H}, \frac{h}{H} \right] \subseteq (0,1]$. Formally, we define the accuracy of within a bin $B_h$ as 
$
    \text{acc}(B_h) = \frac{1}{|B_h|} \sum_{j\in B_h} \mathbf{1}(\hat{y}_i = y_j)
$
and the average confidence within a bin as
$
    \text{conf}(B_h) = \frac{1}{|B_h|} \sum_{j\in B_h} \hat{p}_j \;.
$
If the classifier is calibrated, we expect that the confidence of the classifier reflects its accuracy on unseen test data, that is, $\text{acc}(B_h)=\text{conf}(B_h) ~ \forall h = 1,...., H$. As proposed by \citet{guo2017calibration}, we use the expected calibration error (ECE) to quantify how much the classifier deviates from this criterion: More precisely, in Table \ref{tab:classification}, we report the ECE with the following definition:
\begin{equation}
    \text{calib-err} = \text{ECE} = \sum_{h=1}^H \frac{|B_h|}{m} \big|\text{acc}(B_h) - \text{conf}(B_h) \big|
\end{equation}
with $m$ denoting the overall number of test points.
\subsection{Open Source Code and Hyper-Parameter Selection}
We provide open-source implementations of the PACOH method. In particular, the PACOH-GP variants are implemented in PyTorch and available in the PACOH-GP code repository\footnote{\href{https://github.com/jonasrothfuss/meta_learning_pacoh}{\texttt{https://github.com/jonasrothfuss/meta\_learning\_pacoh}}}. The PACOH-NN variants are implemented in Tensorflow and available here\footnote{\href{https://github.com/jonasrothfuss/pacoh_nn}{\texttt{https://github.com/jonasrothfuss/pacoh\_nn}}}.

\looseness -1 For each of the meta-environments and algorithms, we ran a separate hyper-parameter search to select the hyper-parameters. In particular, we used the \texttt{hyperopt}\footnote{http://hyperopt.github.io/hyperopt/} package \citep{bergstra13} which performs Bayesian optimization based on regression trees. As the optimization metric, we employed the average log-likelihood, evaluated on a separate validation set of tasks.

The scripts for reproducing the hyper-parameter search are included in the PACOH-GP code repository.
For the reported results, we provide the selected hyper-parameters and detailed evaluation results under 
\href{https://tinyurl.com/s48p76x}{\texttt{https://tinyurl.com/s48p76x}}.

\subsection{Meta-Learning for Bayesian Optimization: Vaccine Development}
\label{appx:vaccine_exp}

Here, we provide additional details on the experiment in Section \ref{sec:bandit}.

\paragraph{Data and BO task setup.}
We use data from \citet{widmer2010inferring}, which contains the binding affinities ($\text{IC}_{50}$ values) of many peptide candidates to seven different MHC-I alleles. 
Following \citet{krause2011contextual}, we convert the $\text{IC}_{50}$ values into negative log-scale and normalize them such that 500nM corresponds to zero, i.e., $r := -\log_{10}(\text{IC}_{50}) + \log_{10}(500)$, which is used as the reward signal\added{(i.e. objective function) for our Bayesian Optimization}.

\begin{table}[t]
\centering
\begin{tabular}{l|ccccc}
 Allele & A-0202 & A-0203 & A-0201 &  A-2301 &  A-2402 \\ \hline
 $m_i$ & 1446 & 1442 & 3088 & 103  & 196 \\ 
\end{tabular}
\caption{\looseness -1 MHC-I alleles used for meta-training and corresponding number of samples $m_i$. \vspace{-10pt}}
\label{tab:alleles}
\end{table}

We use 5 alleles\added{(A-0202, A-0203, A-0201, A-2301, A-2402)}to meta-learn a BNN prior. The alleles and the corresponding number of data points, available for meta-training, are listed in Table \ref{tab:alleles}. The most genetically dissimilar allele (A-6901) is used for our BO task. In each iteration, the experimenter (i.e., BO algorithm) chooses to test one peptide among the pool of 813 candidates and receives $r$ as the reward feedback. Hence, we are concerned with an 813-arm bandit wherein the action $a_t \ \in \{1, ..., 813\} = \calA$ in iteration $t$ corresponds to testing the $a_t$-th peptide candidate. In response, the algorithm receives the respective negative log-$\text{IC}_{50}$ as reward $r(a_t)$.

As metrics, we report the \emph{average regret}
$$
R^{avg.}_T :=  \max_{a \in \calA} r(a) - \frac{1}{T} \sum_{t=1}^T r(a_t)
$$
and the \emph{simple regret}
$$
R^{simple}_T :=  \max_{a \in \calA} r(a) - \max_{t=1,...,T} r(a_t) \;.
$$

\added{\paragraph{The PACOH-UCB/TS approach.}
First, we perform meta-learning with PACOH-NN-SVGD on the five datasets $S_1, ..., S_5$ which correspond to the five alleles A-0202, A-0203, A-0201, A-2301, A-2402. As a result, we obtain an approximate hyper-posterior, i.e., a set of priors $\{ P_{\phi_1}, ..., P_{\phi_K} \}$.}

\added{After the meta-learning stage, we perform BO on the test allele (A-6901). Initially, we query a random action $a^0 \in \calA$. From then on, actions $a_t$ ($t > 0$) are chosen either via UCB or Thompson Sampling (TS).
In every iteration $t=1, ..., T$, we have access to the previously queried data $\tilde{S}_{<t} = \{(s_{a_{t'}}, y_{a_{t'}})\}_{t'=0}^{t-1}$. As described in Appendix \ref{appx:meta_testing}, we use this data in combination with the meta-learned prior to obtain an approximate posterior which is represented as a set of NN parameters
$\bigcup_{k=1}^K \{\theta_1^k,\dots, \theta_L^k\}$ (cf. Algorithm \ref{algo:pacoh_nn_target_training}).
In the case of the Upper-Confidence Bound (UCB) acquisition algorithm, we choose the action that maximizes the UCB, i.e.,
\begin{equation}
    a_t = \argmax_{a \in \calA} \tilde{\mu}(y_a | x_a, \tilde{S}_{<t}) + \beta \tilde{\sigma}(y_a | x_a, \tilde{S}_{<t}) ~,
\end{equation}
where $\tilde{\mu}(y | x, \tilde{S}_{<t}) = \frac{1}{K \cdot L}\sum_{k=1}^K \sum_{l=1}^L h_{\theta^k_l}(x)$ is the predictive mean and $\tilde{\sigma}^2(y | x \tilde{S}_{<t}) = \frac{1}{K \cdot L}\sum_{k=1}^K \sum_{l=1}^L (h_{\theta^k_l}(x) - \tilde{\mu}(y | x, \tilde{S}_{<t}))^2$ the epistemic variance of the BNN trained with $\tilde{S}_{<t}$. We use $\beta=2$ for the BO experiments.
In the case of Thompson Sampling (TS), the next action is chosen by sampling a function from the posterior and picking its maximum. In our case, this is done by uniformly sampling one of the NN particles/parameters and taking the action $a$ that maximizes the corresponding NN function $h_{\theta^k_l}(x_a)$ 
\begin{equation}
    a_t = \argmax_{a \in \calA} h_{\theta^k_l}(x_a) ~~~ \text{with} ~~ k \sim \calU(1, ..., L), l \sim \calU(1, ..., L) ~.
\end{equation}}

\paragraph{Baselines.}
\looseness -1\added{The BNN-UCB and BNN-TS baselines work analogously except that instead of meta-learned priors, we use a zero-centered Gaussian prior over the NN parameters. In the case of GP-UCB \citep{srinivas2009gaussian}, a Gaussian Process (GP) instead of a BNN is used as a surrogate model of the objective function.} To ensure a fair comparison, the prior parameters of the GP are meta-learned by minimizing the GP's marginal log-likelihood on the five meta-training tasks. For the prior, we use a constant mean function and tried various kernel functions (linear, SE, Matern). Due to the 45-dimensional feature space, we found the linear kernel to work best. So, the constant mean and the variance parameter of the linear kernel are meta-learned.

%% file: main.bbl
\begin{thebibliography}{105}
\providecommand{\natexlab}[1]{#1}
\providecommand{\url}[1]{\texttt{#1}}
\expandafter\ifx\csname urlstyle\endcsname\relax
  \providecommand{\doi}[1]{doi: #1}\else
  \providecommand{\doi}{doi: \begingroup \urlstyle{rm}\Url}\fi

\bibitem[Alquier(2008)]{alquier2008pac}
Pierre Alquier.
\newblock {PAC-Bayesian bounds for randomized empirical risk minimizers}.
\newblock \emph{Mathematical Methods of Statistics}, 17:\penalty0 279--304,
  2008.

\bibitem[Alquier(2021)]{alquier2021user}
Pierre Alquier.
\newblock {User-friendly introduction to PAC-Bayes bounds}.
\newblock \emph{arXiv preprint arXiv:2110.11216}, 2021.

\bibitem[Alquier and Guedj(2018)]{alquier2018simpler}
Pierre Alquier and Benjamin Guedj.
\newblock {Simpler PAC-Bayesian bounds for hostile data}.
\newblock \emph{Machine Learning}, 107:\penalty0 887--902, 2018.

\bibitem[Alquier et~al.(2016)Alquier, Ridgway, and Chopin]{Alquier2016a}
Pierre Alquier, James Ridgway, and Nicolas Chopin.
\newblock {On the properties of variational approximations of Gibbs
  posteriors}.
\newblock \emph{Journal of Machine Learning Research}, 17:\penalty0 1--41,
  2016.

\bibitem[Amit and Meir(2018)]{amit2017meta}
Ron Amit and Ron Meir.
\newblock {Meta-learning by adjusting priors based on extended PAC-Bayes
  theory}.
\newblock In \emph{International Conference on Machine Learning}, 2018.

\bibitem[Andrychowicz et~al.(2016)Andrychowicz, Denil, Colmenarejo, Hoffman,
  Pfau, Schaul, Shillingford, and De~Freitas]{Andrychowicz2016}
Marcin Andrychowicz, Misha Denil, Sergio~Gómez Colmenarejo, Matthew~W.
  Hoffman, David Pfau, Tom Schaul, Brendan Shillingford, and Nando De~Freitas.
\newblock {Learning to learn by gradient descent by gradient descent}.
\newblock In \emph{Advances in Neural Information Processing Systems}, 2016.

\bibitem[Auer(2002)]{auer2002using}
Peter Auer.
\newblock Using confidence bounds for exploitation-exploration trade-offs.
\newblock \emph{Journal of Machine Learning Research}, 3:\penalty0 397--422,
  2002.

\bibitem[Baxter(2000)]{baxter2000model}
Jonathan Baxter.
\newblock A model of inductive bias learning.
\newblock \emph{Journal of Artificial Intelligence Research}, 2000.

\bibitem[Bengio et~al.(1991)Bengio, Bengio, and Cloutier]{bengio1991}
Y.~Bengio, S.~Bengio, and J.~Cloutier.
\newblock Learning a synaptic learning rule.
\newblock In \emph{International Joint Conference on Neural Networks}, 1991.

\bibitem[Bergstra et~al.(2013)Bergstra, Yamins, and Cox]{bergstra13}
James Bergstra, Daniel Yamins, and David Cox.
\newblock Making a science of model search: Hyperparameter optimization in
  hundreds of dimensions for vision architectures.
\newblock In \emph{International Conference on Machine Learning}, 2013.

\bibitem[Blei et~al.(2017)Blei, Kucukelbir, and McAuliffe]{Blei2016}
David~M. Blei, Alp Kucukelbir, and Jon~D. McAuliffe.
\newblock Variational inference: {A} review for statisticians.
\newblock \emph{Journal of the American Statistical Association}, 112:\penalty0
  859--877, 2017.

\bibitem[Bonilla et~al.(2008)Bonilla, Chai, and Williams]{bonilla2008multi}
Edwin~V. Bonilla, Kian~M. Chai, and Christopher Williams.
\newblock {Multi-task Gaussian Process prediction}.
\newblock In \emph{Advances in Neural Information Processing Systems}, 2008.

\bibitem[Boucheron et~al.(2013)Boucheron, Lugosi, and Massart]{Boucheron2013}
Stephane Boucheron, Gabor Lugosi, and Pascal Massart.
\newblock \emph{{Concentration inequalities: a nonasymptotic theory of
  independence}}.
\newblock {Oxford University Press}, 2013.

\bibitem[Catoni(2007)]{catoni2007pac}
Olivier Catoni.
\newblock {PAC-Bayesian supervised classification: the thermodynamics of
  statistical learning}.
\newblock \emph{IMS Lecture Notes Monograph Series}, 56, 2007.

\bibitem[Cella and Pontil(2021)]{cella2021multi}
Leonardo Cella and Massimiliano Pontil.
\newblock Multi-task and meta-learning with sparse linear bandits.
\newblock In \emph{Uncertainty in Artificial Intelligence}, 2021.

\bibitem[Cella et~al.(2022)Cella, Lounici, and Pontil]{cella2022meta}
Leonardo Cella, Karim Lounici, and Massimiliano Pontil.
\newblock Meta representation learning with contextual linear bandits.
\newblock \emph{arXiv preprint arXiv:2205.15100}, 2022.

\bibitem[Chen et~al.(2021)Chen, Shui, and Marchand]{chen2021generalization}
Qi~Chen, Changjian Shui, and Mario Marchand.
\newblock Generalization bounds for meta-learning: An information-theoretic
  analysis.
\newblock \emph{Advances in Neural Information Processing Systems}, 2021.

\bibitem[Chen et~al.(2017)Chen, Hoffman, Colmenarejo, Denil, Lillicrap,
  Botvinick, and De~Freitas]{Chen2017a}
Yutian Chen, Matthew~W. Hoffman, Sergio~Gómez Colmenarejo, Misha Denil,
  Timothy~P. Lillicrap, Matt Botvinick, and Nando De~Freitas.
\newblock {Learning to Learn without Gradient Descent by Gradient Descent}.
\newblock In \emph{International Conference on Machine Learning}, 2017.

\bibitem[Csisz{\'a}r(1975)]{csiszar1975divergence}
Imre Csisz{\'a}r.
\newblock I-divergence geometry of probability distributions and minimization
  problems.
\newblock \emph{{The Annals of Probability}}, 3:\penalty0 146--158, 1975.

\bibitem[Ding et~al.(2021)Ding, Chen, Levinboim, Goodman, and
  Soricut]{ding2021bridging}
Nan Ding, Xi~Chen, Tomer Levinboim, Sebastian Goodman, and Radu Soricut.
\newblock {Bridging the Gap Between Practice and PAC-Bayes Theory in Few-Shot
  Meta-Learning}.
\newblock In \emph{Advances in Neural Information Processing Systems}, 2021.

\bibitem[Donsker and Varadhan(1975)]{donsker1975}
M.D. Donsker and S.R.S. Varadhan.
\newblock {Large deviations for Markov processes and the asymptotic evaluation
  of certain Markov process expectations for large times}.
\newblock In \emph{Probabilistic Methods in Differential Equations}, 1975.

\bibitem[Dziugaite and Roy(2018)]{dziugaite2018data}
Gintare~Karolina Dziugaite and Daniel~M Roy.
\newblock {Data-dependent PAC-Bayes priors via differential privacy}.
\newblock \emph{Advances in Neural Information Processing Systems}, 2018.

\bibitem[Dziugaite et~al.(2021)Dziugaite, Hsu, Gharbieh, Arpino, and
  Roy]{dziugaite2021role}
Gintare~Karolina Dziugaite, Kyle Hsu, Waseem Gharbieh, Gabriel Arpino, and
  Daniel Roy.
\newblock {On the role of data in PAC-Bayes bounds}.
\newblock In \emph{International Conference on Artificial Intelligence and
  Statistics}, 2021.

\bibitem[Farid and Majumdar(2021)]{farid2021generalization}
Alec Farid and Anirudha Majumdar.
\newblock {Generalization bounds for meta-learning via PAC-Bayes and uniform
  stability}.
\newblock In \emph{Advances in Neural Information Processing Systems}, 2021.

\bibitem[Finn et~al.(2017)Finn, Abbeel, and Levine]{finn2017model}
Chelsea Finn, Pieter Abbeel, and Sergey Levine.
\newblock Model-agnostic meta-learning for fast adaptation of deep networks.
\newblock In \emph{International Conference on Machine Learning}, 2017.

\bibitem[Finn et~al.(2018)Finn, Xu, and Levine]{finn2018probabilistic}
Chelsea Finn, Kelvin Xu, and Sergey Levine.
\newblock Probabilistic model-agnostic meta-learning.
\newblock In \emph{Advances in Neural Information Processing Systems}, 2018.

\bibitem[Fortuin and R{\"a}tsch(2019)]{fortuin2019deep}
Vincent Fortuin and Gunnar R{\"a}tsch.
\newblock {Deep Mean Functions for Meta-Learning in Gaussian Processes}.
\newblock \emph{arXiv preprint arXiv:1901.08098}, 2019.

\bibitem[Fortuin et~al.(2022)Fortuin, Garriga-Alonso, Wenzel, R{\"a}tsch,
  Turner, van~der Wilk, and Aitchison]{fortuin2021bayesian}
Vincent Fortuin, Adri{\`a} Garriga-Alonso, Florian Wenzel, Gunnar R{\"a}tsch,
  Richard Turner, Mark van~der Wilk, and Laurence Aitchison.
\newblock {Bayesian neural network priors revisited}.
\newblock In \emph{International Conference on Learning Representations}, 2022.

\bibitem[Garnelo et~al.(2018)Garnelo, Schwarz, Rosenbaum, Viola, Rezende,
  Eslami, and Teh]{garnelo2018neural}
Marta Garnelo, Jonathan Schwarz, Dan Rosenbaum, Fabio Viola, Danilo~J Rezende,
  SM~Eslami, and Yee~Whye Teh.
\newblock Neural processes.
\newblock In \emph{ICML 2018 workshop on Theoretical Foundations and
  Applications of Deep Generative Models}, 2018.

\bibitem[Germain et~al.(2016)Germain, Bach, Lacoste, and
  Lacoste-Julien]{germain2016pac}
Pascal Germain, Francis Bach, Alexandre Lacoste, and Simon Lacoste-Julien.
\newblock {PAC-Bayesian theory meets Bayesian inference}.
\newblock In \emph{Advances in Neural Information Processing Systems}, 2016.

\bibitem[Goldblum et~al.(2020)Goldblum, Reich, Fowl, Ni, Cherepanova, and
  Goldstein]{goldblum2020unraveling}
Micah Goldblum, Steven Reich, Liam Fowl, Renkun Ni, Valeriia Cherepanova, and
  Tom Goldstein.
\newblock Unraveling meta-learning: Understanding feature representations for
  few-shot tasks.
\newblock In \emph{International Conference on Machine Learning}, 2020.

\bibitem[G{\"o}nen and Alpayd{\i}n(2011)]{gonen2011multiple}
Mehmet G{\"o}nen and Ethem Alpayd{\i}n.
\newblock Multiple kernel learning algorithms.
\newblock \emph{Journal of Machine Learning Research}, 12:\penalty0 2211--2268,
  2011.

\bibitem[Grant et~al.(2018)Grant, Finn, Levine, Darrell, and
  Griffiths]{grant2018recasting}
Erin Grant, Chelsea Finn, Sergey Levine, Trevor Darrell, and Thomas Griffiths.
\newblock {Recasting gradient-based meta-learning as hierarchical Bayes}.
\newblock In \emph{International Conference on Learning Representations}, 2018.

\bibitem[Guan and Lu(2022)]{guan2022fast}
Jiechao Guan and Zhiwu Lu.
\newblock {Fast-rate PAC-Bayesian generalization bounds for meta-learning}.
\newblock In \emph{International Conference on Machine Learning}, 2022.

\bibitem[Guedj(2019)]{guedj2019primer}
Benjamin Guedj.
\newblock {A primer on PAC-Bayesian learning}.
\newblock In \emph{2nd Congress of the French Mathematical Society}, 2019.

\bibitem[Guo et~al.(2017)Guo, Pleiss, Sun, and Weinberger]{guo2017calibration}
Chuan Guo, Geoff Pleiss, Yu~Sun, and Kilian~Q. Weinberger.
\newblock On calibration of modern neural networks.
\newblock In \emph{International Conference on Machine Learning}, 2017.

\bibitem[Hochreiter et~al.(2001)Hochreiter, Younger, and
  Conwell]{Hochreiter2001}
Sepp Hochreiter, A.~Steven Younger, and Peter~R. Conwell.
\newblock {Learning To Learn Using Gradient Descent}.
\newblock In \emph{International Conference on Artificial Neural Networks},
  2001.

\bibitem[Holland(2019)]{holland2019pac}
Matthew Holland.
\newblock {PAC-Bayes under potentially heavy tails}.
\newblock \emph{Advances in Neural Information Processing Systems}, 2019.

\bibitem[Jose and Simeone(2021)]{jose2021information2}
Sharu~Theresa Jose and Osvaldo Simeone.
\newblock Information-theoretic generalization bounds for meta-learning and
  applications.
\newblock \emph{Entropy}, 23:\penalty0 126, 2021.

\bibitem[Jose et~al.(2022)Jose, Park, and
  Simeone]{Jose2022InformationTheoreticAO}
Sharu~Theresa Jose, Sangwook Park, and Osvaldo Simeone.
\newblock {Information-Theoretic Analysis of Epistemic Uncertainty in Bayesian
  Meta-learning}.
\newblock In \emph{International Conference on Artificial Intelligence and
  Statistics}, 2022.

\bibitem[Kahn(1955)]{kahn1995}
Herman Kahn.
\newblock \emph{Use of Different Monte Carlo Sampling Techniques}.
\newblock RAND Corporation, 1955.

\bibitem[Kassraie et~al.(2022)Kassraie, Rothfuss, and Krause]{kassraie2022meta}
Parnian Kassraie, Jonas Rothfuss, and Andreas Krause.
\newblock Meta-learning hypothesis spaces for sequential decision-making.
\newblock In \emph{International Conference on Machine Learning}, 2022.

\bibitem[Kim et~al.(2019)Kim, Mnih, Schwarz, Garnelo, Eslami, Rosenbaum,
  Vinyals, and Teh]{kim2019attentive}
Hyunjik Kim, Andriy Mnih, Jonathan Schwarz, Marta Garnelo, Ali Eslami, Dan
  Rosenbaum, Oriol Vinyals, and Yee~Whye Teh.
\newblock Attentive neural processes.
\newblock In \emph{International Conference on Learning Representations}, 2019.

\bibitem[Kingma and Welling(2014)]{kingma2014auto}
Diederik~P. Kingma and Max Welling.
\newblock {Auto-Encoding Variational Bayes}.
\newblock In \emph{International Conference on Learning Representations}, 2014.

\bibitem[Kirschner et~al.(2019{\natexlab{a}})Kirschner, Mutn\'y, Hiller,
  Ischebeck, and Krause]{kirschner2019linebo}
Johannes Kirschner, Mojmir Mutn\'y, Nicole Hiller, Rasmus Ischebeck, and
  Andreas Krause.
\newblock {Adaptive and Safe Bayesian Optimization in High Dimensions via
  One-Dimensional Subspaces}.
\newblock In \emph{International Conference on Machine Learning},
  2019{\natexlab{a}}.

\bibitem[Kirschner et~al.(2019{\natexlab{b}})Kirschner, Nonnenmacher,
  Mutn{\`y}, Krause, Hiller, Ischebeck, and Adelmann]{kirschner2019swissfel}
Johannes Kirschner, Manuel Nonnenmacher, Mojm{\'\i}r Mutn{\`y}, Andreas Krause,
  Nicole Hiller, Rasmus Ischebeck, and Andreas Adelmann.
\newblock {Bayesian Optimization for Fast and Safe Parameter Tuning of
  SwissFEL}.
\newblock In \emph{International Free-Electron Laser Conference},
  2019{\natexlab{b}}.

\bibitem[Krause and Ong(2011)]{krause2011contextual}
Andreas Krause and Cheng~S. Ong.
\newblock {Contextual Gaussian Process bandit optimization}.
\newblock In \emph{Advances in Neural Information Processing Systems}, 2011.

\bibitem[Kuleshov et~al.(2018)Kuleshov, Fenner, and Ermon]{Kuleshov2018}
Volodymyr Kuleshov, Nathan Fenner, and Stefano Ermon.
\newblock Accurate uncertainties for deep learning using calibrated regression.
\newblock In \emph{International Conference on Machine Learning}, 2018.

\bibitem[Lake et~al.(2015)Lake, Salakhutdinov, and Tenenbaum]{lake2015human}
Brenden~M. Lake, Ruslan Salakhutdinov, and Joshua~B. Tenenbaum.
\newblock Human-level concept learning through probabilistic program induction.
\newblock \emph{Science}, 350:\penalty0 1332--1338, 2015.

\bibitem[Lever et~al.(2013)Lever, Laviolette, and Shawe-Taylor]{Lever2013}
Guy Lever, Fran{\c{c}}ois Laviolette, and John Shawe-Taylor.
\newblock {Tighter PAC-Bayes bounds through distribution-dependent priors}.
\newblock \emph{Theoretical Computer Science}, 473:\penalty0 4--28, 2013.

\bibitem[Li et~al.(2017)Li, Zhou, Chen, and Li]{li2017meta}
Zhenguo Li, Fengwei Zhou, Fei Chen, and Hang Li.
\newblock Meta-sgd: Learning to learn quickly for few-shot learning.
\newblock \emph{arXiv preprint arXiv:1707.09835}, 2017.

\bibitem[Liu and Wang(2016)]{Liu2016}
Qiang Liu and Dilin Wang.
\newblock {Stein Variational Gradient Descent: A General Purpose Bayesian
  Inference Algorithm}.
\newblock In \emph{Advances in Neural Information Processing Systems}, 2016.

\bibitem[Liu et~al.(2021)Liu, Lu, Yan, and Zhang]{liu2021statistical}
Tianyu Liu, Jie Lu, Zheng Yan, and Guangquan Zhang.
\newblock Statistical generalization performance guarantee for meta-learning
  with data-dependent prior.
\newblock \emph{Neurocomputing}, 465:\penalty0 391--405, 2021.

\bibitem[Luo et~al.(2020)Luo, Beatson, Norouzi, Zhu, Duvenaud, Adams, and
  Chen]{luo2020sumo}
Yucen Luo, Alex Beatson, Mohammad Norouzi, Jun Zhu, David Duvenaud, Ryan~P.
  Adams, and Ricky T.~Q. Chen.
\newblock {SUMO: Unbiased Estimation of Log Marginal Probability for Latent
  Variable Models}.
\newblock In \emph{International Conference on Learning Representations}, 2020.

\bibitem[Madden(2004)]{intel_sensor_data}
Samuel Madden.
\newblock Intel lab data.
\newblock \url{http://db.csail.mit.edu/labdata/labdata.html}, 2004.
\newblock Accessed: Sep 8, 2020.

\bibitem[Maurer(2004)]{maurer2004note}
Andreas Maurer.
\newblock {A note on the PAC Bayesian theorem}.
\newblock \emph{arXiv preprint arXiv:0411099}, 2004.

\bibitem[Maurer and Jaakkola(2005)]{maurer2005algorithmic}
Andreas Maurer and Tommi Jaakkola.
\newblock Algorithmic stability and meta-learning.
\newblock \emph{Journal of Machine Learning Research}, 6:\penalty0 967--994,
  2005.

\bibitem[McAllester(1999)]{mcallester1999some}
David~A McAllester.
\newblock {Some PAC-Bayesian theorems}.
\newblock \emph{Machine Learning}, 1999.

\bibitem[Micchelli and Pontil(2004)]{micchelli2004kernels}
Charles Micchelli and Massimiliano Pontil.
\newblock {Kernels for Multi-task Learning}.
\newblock \emph{Advances in Neural Information Processing Systems}, 2004.

\bibitem[Milne et~al.(2017)Milne, Schietinger, Aiba, et~al.]{milne2017swissfel}
Christopher~J Milne, Thomas Schietinger, Masamitsu Aiba, et~al.
\newblock {SwissFEL: the Swiss X-ray Free Electron Laser}.
\newblock \emph{Applied Sciences}, 2017.

\bibitem[Mishra et~al.(2018)Mishra, Rohaninejad, Chen, and Abbeel]{Mishra2018}
Nikhil Mishra, Mostafa Rohaninejad, Xi~Chen, and Pieter Abbeel.
\newblock {A Simple Neural Attentive Meta-Learner}.
\newblock In \emph{International Conference on Learning Representations}, 2018.

\bibitem[Nichol et~al.(2018)Nichol, Achiam, and Schulman]{nichol2018firstorder}
Alex Nichol, Joshua Achiam, and John Schulman.
\newblock {On First-Order Meta-Learning Algorithms}.
\newblock \emph{arXiv}, 2018.

\bibitem[Oneto et~al.(2016)Oneto, Anguita, and Ridella]{oneto2016pac}
Luca Oneto, Davide Anguita, and Sandro Ridella.
\newblock {PAC-Bayesian analysis of distribution dependent priors: Tighter risk
  bounds and stability analysis}.
\newblock \emph{Pattern Recognition Letters}, 80:\penalty0 200--207, 2016.

\bibitem[Ong et~al.(2005)Ong, Smola, and Williamson]{Ong2005}
Cheng~S. Ong, Alexander~J. Smola, and Robert~C. Williamson.
\newblock {Learning the Kernel with Hyperkernels}.
\newblock \emph{Journal of Machine Learning Research}, 3:\penalty0 1043--1071,
  2005.

\bibitem[Parameswaran and Weinberger(2010)]{parameswaran2010large}
Shibin Parameswaran and Kilian~Q Weinberger.
\newblock Large margin multi-task metric learning.
\newblock \emph{Advances in Neural Information Processing Systems}, 2010.

\bibitem[Parrado-Hernandez et~al.(2012)Parrado-Hernandez, Ambroladze,
  Shawe-Taylor, and Sun]{parrado12pac}
Emilio Parrado-Hernandez, Amiran Ambroladze, John Shawe-Taylor, and Shiliang
  Sun.
\newblock {PAC-Bayes Bounds with Data Dependent Priors}.
\newblock \emph{Journal of Machine Learning Research}, 13:\penalty0 3507--3531,
  2012.

\bibitem[Pentina and Lampert(2014)]{pentina2014pac}
Anastasia Pentina and Christoph Lampert.
\newblock {A PAC-Bayesian bound for lifelong learning}.
\newblock In \emph{International Conference on Machine Learning}, 2014.

\bibitem[P{\'e}rez-Ortiz et~al.(2021)P{\'e}rez-Ortiz, Rivasplata, Shawe-Taylor,
  and Szepesv{\'a}ri]{perez2021tighter}
Mar{\'\i}a P{\'e}rez-Ortiz, Omar Rivasplata, John Shawe-Taylor, and Csaba
  Szepesv{\'a}ri.
\newblock Tighter risk certificates for neural networks.
\newblock \emph{Journal of Machine Learning Research}, 22:\penalty0
  10326--10365, 2021.

\bibitem[Pharr et~al.(2016)Pharr, Jakob, and Humphreys]{matt2016}
Matt Pharr, Wenzel Jakob, and Greg Humphreys.
\newblock \emph{Physically Based Rendering: From Theory to Implementation},
  chapter 13.7.
\newblock Morgan Kaufmann, 3rd edition, 2016.

\bibitem[Qin et~al.(2018)Qin, Zhang, Zhao, Wang, Shi, Qi, Shi, and
  Lei]{qin2018rethink}
Yunxiao Qin, Weiguo Zhang, Chenxu Zhao, Zezheng Wang, Hailin Shi, Guojun Qi,
  Jingping Shi, and Zhen Lei.
\newblock Rethink and redesign meta learning.
\newblock \emph{arXiv preprint arXiv:1812.04955}, 2018.

\bibitem[Rasmussen and Williams(2006)]{rasmussen2003gaussian}
Carl~Edward Rasmussen and Christopher K.~I. Williams.
\newblock \emph{Gaussian Processes in machine learning}.
\newblock MIT Press, 2006.

\bibitem[Ravi and Beatson(2018)]{ravi2018amortized}
Sachin Ravi and Alex Beatson.
\newblock {Amortized Bayesian meta-learning}.
\newblock In \emph{International Conference on Learning Representations}, 2018.

\bibitem[Ravi and Larochelle(2017)]{Ravi2017}
Sachin Ravi and Hugo Larochelle.
\newblock {Optimization as a Model for Few-Shot Learning}.
\newblock In \emph{International Conference on Learning Representations}, 2017.

\bibitem[Reeb et~al.(2018)Reeb, Doerr, Gerwinn, and Rakitsch]{reeb2018learning}
David Reeb, Andreas Doerr, Sebastian Gerwinn, and Barbara Rakitsch.
\newblock {Learning Gaussian Processes by Minimizing PAC-Bayesian
  Generalization Bounds}.
\newblock In \emph{Advances in Neural Information Processing Systems}, 2018.

\bibitem[Rezazadeh et~al.(2021)Rezazadeh, Jose, Durisi, and
  Simeone]{rezazadeh2021conditional}
Arezou Rezazadeh, Sharu~Theresa Jose, Giuseppe Durisi, and Osvaldo Simeone.
\newblock Conditional mutual information-based generalization bound for meta
  learning.
\newblock In \emph{International Symposium on Information Theory}, 2021.

\bibitem[Rezende and Mohamed(2015)]{rezende2015variational}
Danilo~Jimenez Rezende and Shakir Mohamed.
\newblock Variational inference with normalizing flows.
\newblock \emph{International Conference on Machine Learning}, 2015.

\bibitem[Rivasplata et~al.(2018)Rivasplata, Parrado-Hern{\'a}ndez,
  Shawe-Taylor, Sun, and Szepesv{\'a}ri]{rivasplata2018pac}
Omar Rivasplata, Emilio Parrado-Hern{\'a}ndez, John~S. Shawe-Taylor, Shiliang
  Sun, and Csaba Szepesv{\'a}ri.
\newblock {PAC-Bayes bounds for stable algorithms with instance-dependent
  priors}.
\newblock \emph{Advances in Neural Information Processing Systems}, 2018.

\bibitem[Rothfuss et~al.(2019)Rothfuss, Lee, Clavera, Asfour, and
  Abbeel]{rothfuss2019promp}
Jonas Rothfuss, Dennis Lee, Ignasi Clavera, Tamim Asfour, and Pieter Abbeel.
\newblock {ProMP: Proximal Meta-Policy Search}.
\newblock In \emph{International Conference on Learning Representations}, 2019.

\bibitem[Rothfuss et~al.(2021{\natexlab{a}})Rothfuss, Fortuin, Josifoski, and
  Krause]{rothfuss2020pacoh}
Jonas Rothfuss, Vincent Fortuin, Martin Josifoski, and Andreas Krause.
\newblock {PACOH: Bayes-Optimal Meta-Learning with PAC-Guarantees}.
\newblock In \emph{International Conference on Machine Learning},
  2021{\natexlab{a}}.

\bibitem[Rothfuss et~al.(2021{\natexlab{b}})Rothfuss, Heyn, Chen, and
  Krause]{rothfuss2021meta}
Jonas Rothfuss, Dominique Heyn, Jinfan Chen, and Andreas Krause.
\newblock {Meta-learning Reliable Priors in the Function Space}.
\newblock In \emph{Advances in Neural Information Processing Systems},
  2021{\natexlab{b}}.

\bibitem[Rothfuss et~al.(2022)Rothfuss, Koenig, Rupenyan, and
  Krause]{rothfuss2022meta}
Jonas Rothfuss, Christopher Koenig, Alisa Rupenyan, and Andreas Krause.
\newblock {Meta-Learning Priors for Safe Bayesian Optimization}.
\newblock In \emph{Conference on Robot Learning}, 2022.

\bibitem[Salakhutdinov et~al.(2012)Salakhutdinov, Tenenbaum, and
  Torralba]{salakhutdinov2012one}
Ruslan Salakhutdinov, Joshua Tenenbaum, and Antonio Torralba.
\newblock {One-shot learning with a hierarchical nonparametric Bayesian model}.
\newblock In \emph{ICML Workshop on Unsupervised and Transfer Learning}, 2012.

\bibitem[Santoro et~al.(2016)Santoro, Bartunov, Botvinick, Wierstra, and
  Lillicrap]{santoro2016meta}
Adam Santoro, Sergey Bartunov, Matthew Botvinick, Daan Wierstra, and Timothy
  Lillicrap.
\newblock Meta-learning with memory-augmented neural networks.
\newblock In \emph{International Conference on Machine Learning}, 2016.

\bibitem[Schmidhuber(1987)]{Schmidhuber1987b}
Juergen Schmidhuber.
\newblock \emph{{Evolutionary principles in self-referential learning}}.
\newblock PhD thesis, Technische Universitaet Munchen, 1987.

\bibitem[Schur et~al.(2023)Schur, Kassraie, Rothfuss, and
  Krause]{schur2023lifelong}
Felix Schur, Parnian Kassraie, Jonas Rothfuss, and Andreas Krause.
\newblock Lifelong bandit optimization: no prior and no regret.
\newblock In \emph{Uncertainty in Artificial Intelligence}, 2023.

\bibitem[Seeger(2002)]{seeger2002pac}
Matthias Seeger.
\newblock {PAC-Bayesian generalisation error bounds for Gaussian process
  classification}.
\newblock \emph{Journal of Machine Learning Research}, 3:\penalty0 233--269,
  2002.

\bibitem[Sener and Koltun(2018)]{sener2018multi}
Ozan Sener and Vladlen Koltun.
\newblock Multi-task learning as multi-objective optimization.
\newblock In \emph{Advances in Neural Information Processing Systems}, 2018.

\bibitem[Shahriari et~al.(2015)Shahriari, Swersky, Wang, Adams, and
  De~Freitas]{shahriari2015taking}
Bobak Shahriari, Kevin Swersky, Ziyu Wang, Ryan~P. Adams, and Nando De~Freitas.
\newblock {Taking the human out of the loop: A review of Bayesian
  optimization}.
\newblock \emph{Proceedings of the IEEE}, 104\penalty0 (1):\penalty0 148--175,
  2015.

\bibitem[Shalaeva et~al.(2020)Shalaeva, Esfahani, Germain, and
  Petreczky]{shalaeva2020improved}
Vera Shalaeva, Alireza~Fakhrizadeh Esfahani, Pascal Germain, and Mihaly
  Petreczky.
\newblock {Improved PAC-Bayesian bounds for linear regression}.
\newblock In \emph{AAAI Conference on Artificial Intelligence}, 2020.

\bibitem[Shalev-Shwartz and Ben-David(2014)]{shalev2014understanding}
Shai Shalev-Shwartz and Shai Ben-David.
\newblock \emph{Understanding machine learning: From theory to algorithms}.
\newblock Cambridge University Press, 2014.

\bibitem[Silva et~al.(2012)Silva, Moody, Scott, Celi, and
  Mark]{silva2012predicting}
Ikaro Silva, George Moody, Daniel~J. Scott, Leo~A. Celi, and Roger~G. Mark.
\newblock Predicting in-hospital mortality of icu patients: The
  physionet/computing in cardiology challenge 2012.
\newblock In \emph{Computing in Cardiology}, 2012.

\bibitem[Snell et~al.(2017)Snell, Swersky, and Zemel]{snell2017prototypical}
Jake Snell, Kevin Swersky, and Richard Zemel.
\newblock Prototypical networks for few-shot learning.
\newblock In \emph{Advances in Neural Information Processing Systems}, 2017.

\bibitem[Srinivas et~al.(2009)Srinivas, Krause, Kakade, and
  Seeger]{srinivas2009gaussian}
Niranjan Srinivas, Andreas Krause, Sham~M. Kakade, and Matthias Seeger.
\newblock Gaussian process optimization in the bandit setting: No regret and
  experimental design.
\newblock In \emph{International Conference on Machine Learning}, 2009.

\bibitem[Thakur et~al.(2019)Thakur, Van~Hoof, Gupta, and
  Meger]{thakur2019unifying}
Sanjay Thakur, Herke Van~Hoof, Gunshi Gupta, and David Meger.
\newblock Unifying variational inference and pac-bayes for supervised learning
  that scales.
\newblock \emph{arXiv preprint arXiv:1910.10367}, 2019.

\bibitem[Thompson(1933)]{thompson1933likelihood}
William~R Thompson.
\newblock On the likelihood that one unknown probability exceeds another in
  view of the evidence of two samples.
\newblock \emph{Biometrika}, 25:\penalty0 285--–294, 1933.

\bibitem[Thrun and Pratt(1998)]{thrun1998}
Sebastian Thrun and Lorien Pratt.
\newblock \emph{Learning to Learn}.
\newblock Springer, 1998.

\bibitem[Vinyals et~al.(2016)Vinyals, Blundell, Lillicrap, Kavukcuoglu, and
  Wierstra]{vinyals2016matching}
Oriol Vinyals, Charles Blundell, Timothy Lillicrap, Koray Kavukcuoglu, and Daan
  Wierstra.
\newblock Matching networks for one shot learning.
\newblock In \emph{Advances in Neural Information Processing Systems}, 2016.

\bibitem[Widmer et~al.(2010)Widmer, Toussaint, Altun, and
  R{\"a}tsch]{widmer2010inferring}
Christian Widmer, Nora~C. Toussaint, Yasemin Altun, and Gunnar R{\"a}tsch.
\newblock Inferring latent task structure for multitask learning by multiple
  kernel learning.
\newblock \emph{Bioinformatics}, 2010.

\bibitem[Wilson et~al.(2016)Wilson, Hu, Salakhutdinov, and
  Xing]{wilson2016deep}
Andrew~Gordon Wilson, Zhiting Hu, Ruslan Salakhutdinov, and Eric~P Xing.
\newblock Deep kernel learning.
\newblock In \emph{International Conference on Artificial Intelligence and
  Statistics}, 2016.

\bibitem[Xu et~al.(2020)Xu, Ton, Kim, Kosiorek, and Teh]{xu2020metafun}
Jin Xu, Jean-Francois Ton, Hyunjik Kim, Adam Kosiorek, and Yee~Whye Teh.
\newblock Metafun: Meta-learning with iterative functional updates.
\newblock In \emph{International Conference on Machine Learning}, 2020.

\bibitem[Yang and Chu(2017)]{yang2017approximating}
Zhen-Hang Yang and Yu-Ming Chu.
\newblock On approximating the modified bessel function of the second kind.
\newblock \emph{{Journal of Inequalities and Applications}}, 41:\penalty0 1--8,
  2017.

\bibitem[Yin et~al.(2020)Yin, Tucker, Zhou, Levine, and Finn]{yin2020meta}
Mingzhang Yin, George Tucker, Mingyuan Zhou, Sergey Levine, and Chelsea Finn.
\newblock Meta-learning without memorization.
\newblock In \emph{International Conference on Learning Representations}, 2020.

\bibitem[Yoon et~al.(2018)Yoon, Kim, Dia, Kim, Bengio, and
  Ahn]{yoon2018bayesian}
Jaesik Yoon, Taesup Kim, Ousmane Dia, Sungwoong Kim, Yoshua Bengio, and Sungjin
  Ahn.
\newblock Bayesian model-agnostic meta-learning.
\newblock In \emph{Advances in Neural Information Processing Systems}, 2018.

\bibitem[Yu et~al.(2005)Yu, Tresp, and Schwaighofer]{yu2005learning}
Kai Yu, Volker Tresp, and Anton Schwaighofer.
\newblock {Learning Gaussian processes from multiple tasks}.
\newblock In \emph{International Conference on Machine Learning}, 2005.

\bibitem[Zien and Ong(2007)]{zien2007}
Alexander Zien and Cheng~Soon Ong.
\newblock {Multiclass Multiple Kernel Learning}.
\newblock In \emph{International Conference on Machine Learning}, 2007.

\end{thebibliography}
